\documentclass[12pt, letterpaper]{article}

\usepackage[sort&compress, numbers]{natbib}

\usepackage{graphicx}
\usepackage[top=0.75in,bottom=0.75in,right=0.75in,left=0.75in]{geometry}
\usepackage{acronym}
\usepackage{subcaption}
\usepackage{caption,booktabs}
\usepackage{hhline}
\usepackage{framed}

\usepackage{pgfsys}
\usepackage{url,hyperref,lineno,microtype,subcaption}
\hypersetup{colorlinks, citecolor={blue},urlcolor={blue}, linkcolor=[rgb]{0.66666667,0.26666667,0}}
\usepackage[onehalfspacing]{setspace}

\usepackage{siunitx}

\usepackage{eucal}                 
\usepackage{verbatim}              
\usepackage{amsmath,amsthm,amsfonts,amsopn,amssymb} 
\usepackage{ctable}                
\usepackage[overlay]{textpos}      
\usepackage{pstricks}              
\usepackage{afterpage}             
\usepackage{longtable} 

\usepackage{pdflscape} 

\setcitestyle{citesep={,}}

\usepackage{float}
\floatstyle{boxed}
\newfloat{code}{h}{ext}
\floatname{code}{Code}

\usepackage{fancyvrb}
\DefineVerbatimEnvironment{code}{Verbatim}{fontsize=\small}
\DefineVerbatimEnvironment{example}{Verbatim}{fontsize=\small}

\usepackage[utf8]{inputenc}
\usepackage[T1]{fontenc}

\usepackage[export]{adjustbox}
\usepackage{ragged2e}
\usepackage{rotating}
\usepackage{booktabs, tabularx, threeparttable}
\newcolumntype{L}{>{\small\linespread{0.84}\selectfont%
                    \RaggedRight\hspace{0pt}}X}
\newcolumntype{R}{>{\raggedleft\arraybackslash}p{6cm}}
\usepackage[referable]{threeparttablex}
\usepackage{booktabs, ltablex}

\usepackage{lscape}
\newcolumntype{C}{>{\centering\arraybackslash}X}
 \newcolumntype{b}{>{\hsize=1.35\hsize}C}
\newcolumntype{s}{>{\hsize=0.65\hsize}C}
\newcolumntype{m}{>{\hsize=1.0\hsize}C}

\usepackage{paralist}
\usepackage[nowatermark]{fixmetodonotes}

\usepackage{amsmath,amsfonts,bm}

\def\eqref#1{equation~\ref{#1}}

\def\1{\bm{1}}

\def\rvepsilon{{\bm{\epsilon}}}

\def\vzero{{\bm{0}}}
\def\vone{{\bm{1}}}

\def\ve{{\bm{e}}}

\def\vr{{\bm{r}}}

\def\vx{{\bm{x}}}
\def\vy{{\bm{y}}}
\def\vz{{\bm{z}}}

\def\evx{{x}}

\def\mP{{\bm{P}}}

\def\mR{{\bm{R}}}

\DeclareMathAlphabet{\mathsfit}{\encodingdefault}{\sfdefault}{m}{sl}
\SetMathAlphabet{\mathsfit}{bold}{\encodingdefault}{\sfdefault}{bx}{n}

\def\sR{{\mathbb{R}}}

\newcommand{\R}{\mathbb{R}}

\DeclareMathOperator*{\argmin}{arg\,min}

\DeclareMathOperator{\sign}{sign}

\usepackage{lineno}
\usepackage{url}
\usepackage{booktabs}       
\usepackage{multirow}
\usepackage{multicol}
\usepackage{soul}
\usepackage{nicefrac}
\usepackage{microtype}      
\usepackage{xcolor}         
\usepackage{makecell}
\newcolumntype{x}[1]{>{\centering\arraybackslash\hspace{0pt}}p{#1}}
\usepackage{enumitem}

\usepackage{stackengine}
\newsavebox\mybox

\graphicspath{
{Figures/}
}

\newcommand{\bs}[1]{\boldsymbol{#1}}

\def \relu{\mathsf{relu}}
\def \max{\mathsf{max}}

\def \d{{\hyperref[dDef]{{\rm d}}}}

\newcolumntype{C}{>{\centering\arraybackslash}m{6em}}

\DeclareUnicodeCharacter{FB01}{fi}
\DeclareUnicodeCharacter{FB03}{ffi}
\DeclareUnicodeCharacter{FB02}{fl}

\providecommand{\keywords}[1]
{
  \small	
  \textbf{\textit{Keywords---}} #1
}

\def \zero{{\hyperref[eDef]{\bs{{\rm 0}}}}}

\usepackage{authblk}

\title{Looking deeper into interpretable deep learning in neuroimaging: a comprehensive survey}
\author[1, 2]{Md Mahfuzur Rahman\thanks{mahfuz.gsu@gmail.com}}
\author[2]{Vince Calhoun\thanks{vcalhoun@gsu.edu}}
\author[1, 2]{Sergey Plis\thanks{s.m.plis@gmail.com}}

\affil[1]{Department of Computer Science, Georgia State University, Atlanta, GA, USA}
\affil[2]{Tri-institutional Center for Translational Research in Neuroimaging and Data Science (TReNDS)
Georgia State University, Georgia Institute of Technology, Emory University
Atlanta, GA, USA}
\date{}                     
\begin{document}

\newgeometry{top=0.5in,bottom=0.5in,right=0.75in,left=0.75in}

\maketitle

\vspace{-1.5cm}

\thispagestyle{plain}
\begin{center}       
    \vspace{0.5cm}
    \textbf{Abstract}
\end{center}

Deep learning (DL) models have been popular due to their ability to learn directly from the raw data in an end-to-end paradigm, alleviating the concern of a separate error-prone feature extraction phase. Recent DL-based neuroimaging studies have also witnessed a noticeable performance advancement over traditional machine learning algorithms. But the challenges of deep learning models still exist because of the lack of transparency in these models for their successful deployment in real-world applications. In recent years, Explainable AI (XAI)  has undergone a surge of developments mainly to get intuitions of how the models reached the decisions, which is essential for safety-critical domains such as healthcare, finance, and law enforcement agencies. While the interpretability domain is advancing noticeably, researchers are still unclear about what aspect of model learning a post hoc method reveals and how to validate its reliability. This paper comprehensively reviews interpretable deep learning models in the neuroimaging domain. 
Firstly, we summarize the current status of interpretability resources in general, focusing on the progression of methods, associated challenges, and opinions. Secondly, we discuss how multiple recent neuroimaging studies leveraged model interpretability to capture anatomical and functional brain alterations most relevant to model predictions. Finally, we discuss the limitations of the current practices and offer some valuable insights and guidance on how we can steer our future research directions to make deep learning models substantially interpretable and thus advance scientific understanding of brain disorders. 

 \keywords{deep learning, interpretability, neuroimaging, brain dynamics, psychiatric disorders} 
\afterpage{\aftergroup\restoregeometry}
\newpage

\tableofcontents
\newpage


\section{Introduction}
\label{intro}

Advancing our understanding of brain dynamics is the underpinning to uncovering the underlying neurological conditions~\cite{goldberg1992common, calhoun2014chronnectome, sui2020neuroimaging}. Thus, localization and interpretation of subject-specific spatial and temporal activity may help guide our understanding of the disorder. As such, a persistent goal of Artificial Intelligence (AI) in the neuroimaging domain is leveraging magnetic resonance imaging (MRI) data to enable machines to learn the functional dynamics or anatomical alterations associated with underlying neurological disorders. 

The current understanding of brain functions and structure reveals that the changes in different brain networks can best explain brain disorders~\cite{mulders2015resting, sheffield2016cognition}. Moreover, a brain network is not necessarily spatially localized. Traditional analytical approaches attempt to find group-level differences rather than deal with individual-level decision-making. However, to receive the full translational impact of neuroimaging studies on clinical practices, clinicians must deal with each individual as a separate case. These limitations naturally encouraged people to look for an AI-led understanding of mental disorders~\cite{zarogianni2013towards, dluhovs2017multi, schnack2019improving}. Instead of looking into the brain regions independently, machine learning (ML) models look for undiscovered holistic patterns from the data using the advanced knowledge of applied statistics and mathematical optimization techniques~\cite{hastie2009elements}.

Moreover, ML can generate individual-level diagnostic and prognostic decisions. Along these lines, standard machine learning (SML) models gained a varying degree of success, and the expert-led feature extraction and selection step is almost a prerequisite for its well-functioning~\cite{Khazaee2016}. However, these representations heavily rely on strong assumptions and can miss essential aspects of the underlying dynamics. Unfortunately, when trained on raw data, SML models cannot perform well~\cite{plis2014deep, bellman1961adaptive, rahman2022interpreting}. However, we need to go beyond existing knowledge, and learning from the raw data is essential for further advancement in mental health analysis. Specifically, direct learning from the data may reveal undiscovered and valuable patterns within the data and may bring translational value to clinical practices. It may also accelerate diagnostic and prognostic decision-making processes, eventually leading to personalized treatment plans. While SML models fail to learn from the raw data, deep learning (DL) has been very popular because it does not require prior feature selection or intermediate intervention~\cite{payan2015predicting, patel2016classification, islam2018brain, zhang2021explainable, lecun2015deep}. It can learn automatically from the raw data and find discriminative and potentially useful clinical features. 

While the interpretability of DL models is highly desirable and may faster uncover domain-specific knowledge~\cite{hicks2021explaining, ghorbani2020deep}, deep learning models are black-boxes~\cite{koh2017understanding} and the exact learning mechanism is still unknown. Introspecting DL models in post hoc manner can be unreliable because what the models actually learned depend on their architectures~\cite{fong2017interpretable} and their intializations~\cite{hinton2018deep} during training. Moreover, there is a disagreement problem among different interpretability methods~\cite{krishna2022disagreement,han2022explanation} because the methods are basically heuristically inspired and investigates various aspects of the model. Moreover, there is no agreed validation method for the post hoc explanations in neuroimaging studies, hindering the widespread use of automatic discovery. While interpreting models in a faithful and useful manner is a challenging task, it is an undeniable step before applying them to generate new reliable and actionable insights to combat the disorders. 

In this review article, we provide a comprehensive review of deep learning model interpretability for neuroimaging studies. We articulate the philosophical ground, dimensions, requirements for the interpretability problem and summarize the commonly used approaches and metrics to achieve reliable model interpretability. Then, we discuss the recent developments of DL approaches to neuroimaging and show some encouraging illustrations for how various interpretability concepts have been applied for new discoveries. We complement the discussion providing a set of guidance and caveats that we think will serve as a useful guide for the future practitioners. 


\section{Related Work}
There exist some reviews in the literature~\cite{vieira2017using, zhang2020survey, huff2021interpretation, thibeau2022interpretability} for interpretable deep learning in neuroimaging and medical domains~\cite{rasheed2022explainable, salahuddin2022transparency}. However, they either focused on machine learning models or general medical imaging, and very few focused on deep learning interpretability in connection to neuroimaging. The complete end-to-end guideline for interpretability practices in neuroimaging is clearly lacking. We anticipate, starting from the philosophical basis, that the complete guide should provide a broader notion of interpretability, a quick introduction to the commonly used methods, and validation metrics that future studies can use. Moreover, it needs to be clarified the usage trend of these methods and their utility in clinical practices and scientific discovery. That is, we need to clarify how frequently the most prevailing methods have been utilized and the major scientific progress made along the way. Moreover, very little research~\cite{kohoutova2020toward} discussed the desiderata of interpretability framework in neuroimaging. So, there still remains a scope to make a comprehensive accumulation of the prevailing concepts focusing on aspects of deep learning performance, novel findings in interpretability research, and possible implications and connections between them in the neuroimaging domain. This review aims to provide a field guide for interpretable deep learning for neuroimaging study, especially for new aspirants in this direction of research. 

\section{Organization of the Paper}
In Section \ref{philosophy}, we discuss the philosophical views of scientific explanations. We then introduce the problem of \emph{interpretability} for AI models from a holistic point of view in Section \ref{intro_interpretability}. In Section \ref{taxonomy}, we provide a useful taxonomy of interpretability methods and showed several illustrative neuroimaging studies using those methods. We also provide a brief introduction to all the major branches of interpretability approaches and the intuitions behind all the major methods in each branch. We discuss the desiderata of interpretability in AI and the axioms that need to be satisfied by the interpretability methods in Section \ref{secAxioms}. In Section \ref{eval}, we describe the common sanity tests to justify the initial validation of the post hoc explanations. We also accumulated the formal evaluation metrics proposed in interpretability literature to provide quantitative validation of the generated explanations for synthetic and real datasets. We also complement this section with the caveats the earlier studies talked about while considering using the post hoc approaches. In Section \ref{interpretable_neuro}, we discuss the deep learning approach for neuroimaging and the significance of \emph{interpretability} in these studies. We start our discussion with the traditional feature engineering approach in Section \ref{feature_eng}, and then in Section \ref{deep_learning}, we turn our discussion to the potential of deep learning approaches for neuroimaging research. For deep learning approaches, we emphasize the need for transfer learning (Section \ref{transfer}), and interpretability (Section \ref{interpretability}) to support the discoveries as clinically and neuroscientifically valuable. In Section \ref{review}, we provide a detailed review of the recent neuroimaging studies that used all the major interpretability methods as depicted in Figure \ref{fig:taxonomy}. We show some demonstrative examples of how recent neuroimaging studies are using the idea of interpretability for novel neuropsychiatric biomarker discoveries. In Section \ref{usage_trend}, we investigated the usage trend of interpretability approaches in more than 300 neuroimaging studies. We then, based on the overall findings of this review, propose useful suggestions and caveats for future interpretability practices in neuroimaging in Section \ref{suggestions}. We finally discuss our conclusive remarks in Section \ref{conclusion}.

\section{Philosophy of Scientific Explanations}
\label{philosophy}

Hempel and Oppenheim (1948)~\cite{hempel1948studies} believed that explanation and prediction have the same logical structure, and hence they referred to explanations as "deductive systematization." Bechtel and Abrahamsen (2005)~\cite{bechtel2005explanation} viewed explanations as a mechanistic alternative and may depart from widely accepted nomological explanations, which means a phenomenon if explained, must subsume under a law. The authors deemed explanations in life sciences as "identifying the mechanism responsible for a given phenomenon." Lewis (1986)~\cite{lewis1986philosophy} viewed it as "to explain an event is to provide some information about its causal history." However, Lewis did not provide any restricted notion of what information qualifies as part of the causal history. Still, there is no formal definition of "Explainability" or "Interpretability" in the field of \emph{Artificial Intelligence} ~\cite{lipton2018mythos, doshi2017towards, miller2019explanation}. As many researchers indicated, the ongoing interpretability practices use only researchers' intuition that is susceptible to cognitive biases~\cite{kahneman2011thinking} and social expectations~\cite{hilton1990conversational}. However, as de Graaf and Malle~\cite{de2017people} hypothesized, this is not unnatural because as long as people build intentional \emph{agents}, people will expect explanations from the models using the same conceptual framework people use to explain human behavior. In the current practices of "Explainable AI," the communication gap between the researchers and practitioners is evident, and Miller et al.~\cite{miller2017explainable} describes this phenomenon as "the inmates running the asylum." While we also admit that the current practices have some inherent human bias and social expectations, interpretability literature so far has been rich with different useful methods and valuable opinions that we will discuss in this paper.

\section{What and Why Is Model Interpretability?}
\label{intro_interpretability}

ML systems, generally optimized to exhibit task performance, outperform humans on different computer vision and language processing tasks. However, the deployment of these systems requires satisfying other auxiliary desiderata such as safety, nondiscrimination, justice, and providing the right to explanation~\cite{doshi2017towards}. The unique purpose of model interpretability is to satisfy these additional criteria. 

Traditionally, an ML system optimizes an objective function upon which it exhibits its predictive performance. However, a mere objective function does not include other desiderata of ML systems for its wide-ranging real-world scenarios. Thus, regardless of an ML system's performance, those systems are still incomplete. In other words, stakeholders might seek trust, causality, transferability, informativeness, and fairness as defined in~\cite{lipton2018mythos}. Hence, as argued in~\cite{doshi2017towards},\emph{interpretability} or, in other words, explanations can be one of many ways to make these gaps in problem visualization more evident to us. Some scenarios Doshi-Velez and Kim include:

\begin{itemize}[leftmargin=0.3in]
\item \emph{Scientific Understanding/Data Interpretation:} We may want to create knowledge from an ML system. Explanations may be one of the ways to create knowledge from the machine's learned behavior. 

\item \emph{Safety:} Incorporating all the accompanying scenarios in developing an artificial agent is not feasible. In that case, an explanation may flag undesirable model behavior. 

\item \emph{Ethics:} In problem formulation, one might not consider apriori to remove any potential bias, but the model may learn some unwanted discriminating pattern within the data.

\item \emph{Mismatched Objectives:} 
Often, for building an agent, one may optimize for a proxy function rather than the actual goal. In that case, the agent may discard all other factors that were very relevant to the ultimate goal. For example, a scientist may want to investigate different progressive stages of Alzheimer's but end up building a classifier for Alzheimer's patients from healthy controls. 

\item \emph{Multi-objective Trade-offs :} When an ML system has multiple competing objectives to be satisfied, it may only be possible to incorporate some of them due to the unknown dynamics of their trade-offs. 
\end{itemize}

\subsection{How to Achieve Interpretability?}
Interpretability in machine learning models can be achieved in different ways~\cite{ancona2019gradient}. The first and most preferable approach is to build an inherently interpretable model, e.g., a linear one. However, these models may compromise their predictive capacity for transparent Interpretability. The second approach is to build a model that can perform predictions and simultaneously generate explanations. However, it is a very challenging task because the accepted meaning of the term 'interpretability 'still needs to be settled in the research community. Moreover, it requires both the ground-truth explanations and the labeled samples to train simultaneously for prediction and explanation generation. The third approach is to use separate explanation methods to work on top of the existing models. That is, the existing models can be any black-box model (e.g., deep learning models), and the explanation methods are responsible for generating explanations for the models. Interpretability is especially important when deep learning models are used for knowledge extraction. Regardless of good predictive performance by a DL model,  it may still not be useful for discovery as the model may have only learned spurious correlations \cite{geirhos2020shortcut}. Most of the interpretability methods in the literature are designed around the third interpretability approach, frequently referred to as \emph{post hoc} methods.

\subsection{Global vs. Local Interpretability}
The scope of Interpretability in machine learning is another consideration. For example, {\em Global Interpretability} deals with the overall behavior of the model, such as discovering patterns and the interrelationships among them used for predictions. {\em Global Interpretability} is useful to debug a model, specifically to diagnose if the model has any inherent bias or has learned any artifact instead of the objects of interest. As global Interpretability is very hard to obtain because it requires building a relationship among all predictions made by the model, people traditionally end up with local Interpretability that deals with explaining model behavior case-by-case basis. For example, {\em Local Interpretability} tries to explain why the image has been classified as "cat"/"dog" or why a particular loan application has been "accepted"/"rejected."

While we recommend reading some other literature reviews~\cite{guidotti2018survey, gilpin2018explaining, carvalho2019machine, arrieta2020explainable, linardatos2020explainable, samek2021explaining, ras2022explainable} that cover comprehensive discussion of interpretability methods, we briefly describe the key concepts, axioms, methods and metrics used in interpretable machine learning.  

\subsection{Looking at the Model Interpretability Problem}
\begin{figure} [!htbp]
 \includegraphics[width=1\linewidth]{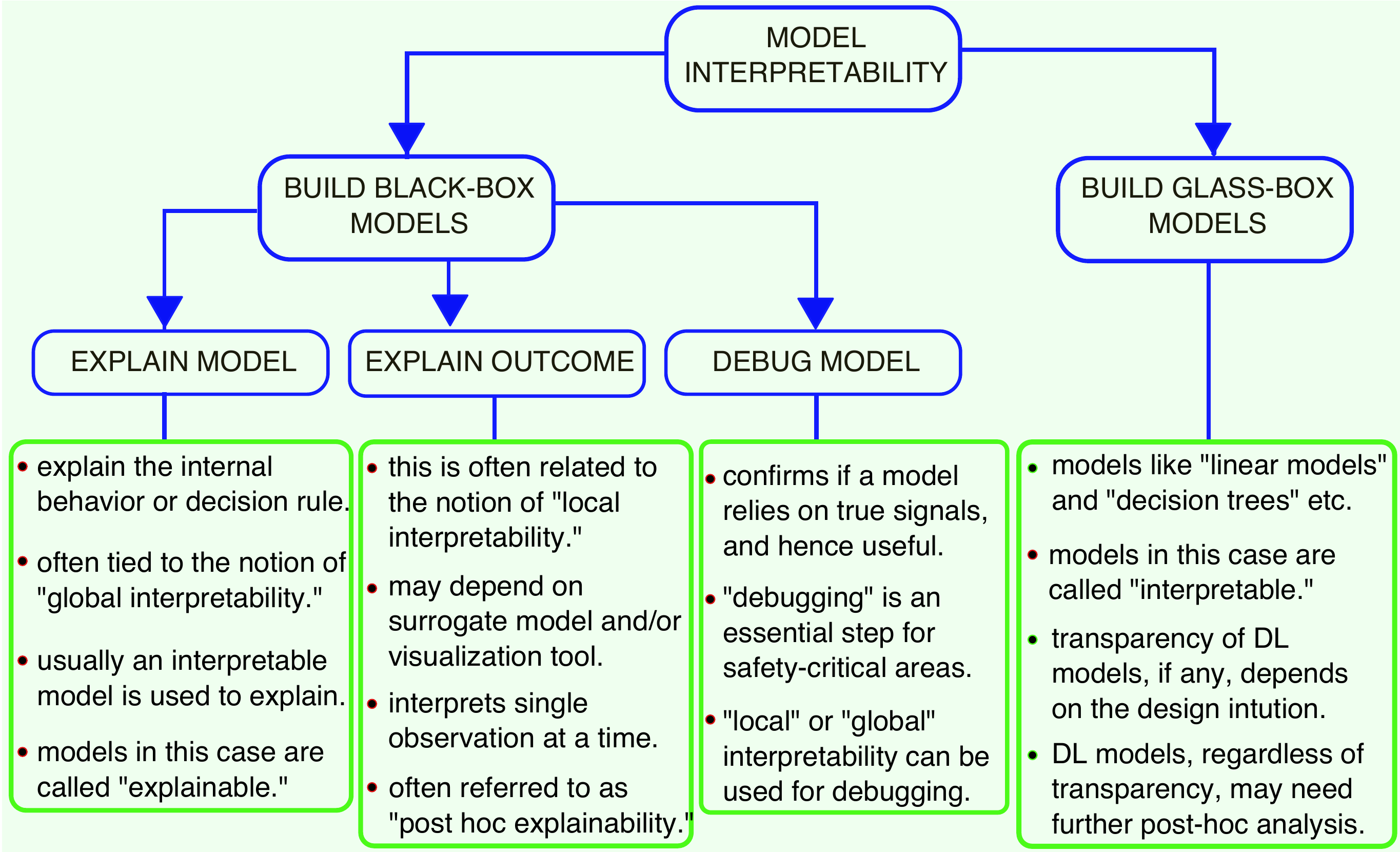}
 \centering
 \caption{The different aspects of Explainable AI from the holistic standpoint. We can address the transparency of the interpretability problem in many ways---by building transparent glass-box models or by building black-box models and explaining different inner (functional mechanism) and outer facets (predictions) of the models. We use the term ``explainability" as a level of ``interpretability," where the latter term refers to the inherent interpretability of the model that comes from its design perspective, and the earlier term is more focused on clarifying the internal functions of black-box models.}
 \label{fig:taxonomy_holistic}
 \end{figure}

Guidotti et al.~\cite{guidotti2018survey} divides the black-box explanation into three sub-categories: \emph{model explanation} means explaining the overall logic of the model; \emph{outcome explanation} means finding the correlation between individual input and corresponding decision; \emph{model inspection} means explaining the behavioral change with changes in input and other parameters or explaining what parts of the model take specific micro-decisions. We provide comprehensive insights into the different aspects of the interpretability problem in Figure~\ref{fig:taxonomy_holistic}.

\subsection{Important Terminology in Interpretability}

As the field of ``Explainable AI (XAI)" is growing rapidly, researchers have defined several important notions useful for the discussion. In this section, we discuss several terminologies often considered significant in interpretability literature. One important point is to note that the terms \emph{interpretability} and \emph{explainability} are elusive. Many studies used the terms interchangeably~\cite{gilpin2018explaining,miller2019explanation,arrieta2020explainable}. However, to define these important terminologies, we have attempted to come to a general agreement with most of the interpretability literature. We define the terms as follows: 

\begin{itemize}[leftmargin=0.25in]
\item \emph{Interpretability:} Doshi-Velez and Kim~\cite{doshi2017towards} defined \emph{interpretability} as  the ``ability to explain or to present in understandable terms to a human." \emph{Interpretability} is a passive characteristic of a model that indicates the level the model makes sense to humans~\cite{arrieta2020explainable}. \emph{Interpretability} is more about the design perspective of a model, and is often tied to the notion of \emph{transparency}. For a fully interpretable model, explanations about the decisions or the decision-making process are obvious, and hence no other separate explanation tools are required. 

\item \emph{Explainability:} The term \emph{explainability} is a broader general term compared to \emph{interpretability}. \emph{Explainability} is an active characteristic of a model~\cite{arrieta2020explainable}, referring to its ability to clarify the internal functions or the rationale the model is using to make decisions, usually in the case of black-box models. \emph{Explainability} is used to specify a level of \emph{interpretability} and is usually considered as a concession to the latter. In short, interpretable models are inherently explainable, but the reverse is not always true. That is, \emph{interpretability} refers to ``how" and ``why" aspects of a model's decision-making process, whereas \emph{explainability} only attempts to make a non-interpretable model to an explainable one, as a concession and attempts to answer only ``why" aspect of the model's decisions. 

\item \emph{Understandability:} \emph{understandability} is associated with the notion that if a model's behavior makes sense to humans without even understanding the mechanistic or algorithmic aspect of the model~\cite{montavon2018methods}. \emph{understandability} is also referred to as \emph{intelligibility}~\cite{arrieta2020explainable}.

\item \emph{Comphrenhensibility:} An interpretable model is \emph{comprehensible}, so they imply the same aspect of a model~\cite{guidotti2018survey}. It is the ability of how the learned knowledge of a model can be represented in human understandable form~\cite{arrieta2020explainable}.

\item \emph{Transparency:} The concept of \emph{transparency} is related to understanding the mechanism by which the model works~\cite{lipton2018mythos}. According to Lipton, transparency can be at different levels---at the level of the entire model, the level of the individual components such as input, parameters, and calculation, and the level of the training algorithm.

\item \emph{Fidelity:} \emph{fidelity} of an interpretable model is a comparative assessment of its accuracy with respect to the black-box model the interpretable model is trying to explain~\cite{guidotti2018survey}.

\end{itemize}

As a disclaimer for the remaining part of the article, we emphasize that \emph{model interpretability} is more about designing inherently interpretable models and \emph{model explainability} is a concession for \emph{model interpretability} with the intent to clarify the functions of black-boxes. However, we took the freedom of using the more commonly used term \emph{interpretability} hereafter even in the context of black-box models, where \emph{explainability} is more appropriate in its true sense.




\section{Taxonomy of Interpretability Methods}
\label{taxonomy}

In this section, we describe different interpretability methods in the literature. We provide a taxonomy of the interpretability methods in Figure~\ref{fig:taxonomy}. We note that this taxonomy is not perfect in the traditional sense, as the categorization of interpretability methods is still evolving. While we discard some infrequent or obsolete approaches and include some emerging methods, this taxonomy is inspired mainly by Ras et al.~\cite{ras2022explainable}.

\begin{enumerate}[leftmargin=0.25in]

\item {\bf Visualization:} Visualization methods focus on highlighting the discriminative regions of the input that mainly influenced the model's decision. This approach is prevalent for deep learning models, especially in computer vision.

\item {\bf Distillation:} Distillation methods focus on building a separate "transparent box" model, which is directly interpretable to extract the salient regions or crucial decision rules that guide the original model to reach its decisions. Methods under this category are usually model-agnostic. Moreover, the resulting explanations may be a set of rules or visualization of important regions, similar to visualization methods. 

\item {\bf Intrinsic:} Intrinsic methods consider model interpretability during model design or training. This approach usually leads toward joint training for predictions and explanations or provides a more transparent model where an explanation is somewhat intuitive. A separate post hoc analysis may be required for the latter ones. 

\item {\bf Counterfactual:} Counterfactual explanations~\cite{wachter2017counterfactual, mothilal2020explaining} usually do not explain the specific output. Instead, it explains in the form of hypothetical scenarios, potentially intending to provide algorithmic recourse. It provides a better understanding of how the decisions change over the input space and allows users more options to change the model's decision~\cite{dandl2020multi}. 

\item{\bf Influence Functions:} To generate an explanation for a prediction, influence functions~\cite{koh2017understanding} find the influence of the training points on the learning algorithm that leads toward this model prediction. 
\end{enumerate}

\fontfamily{phv}\selectfont
\renewcommand{\rmdefault}{phv}\rmfamily

\begin{figure} 
\centering
  \def\svgwidth{\columnwidth}
  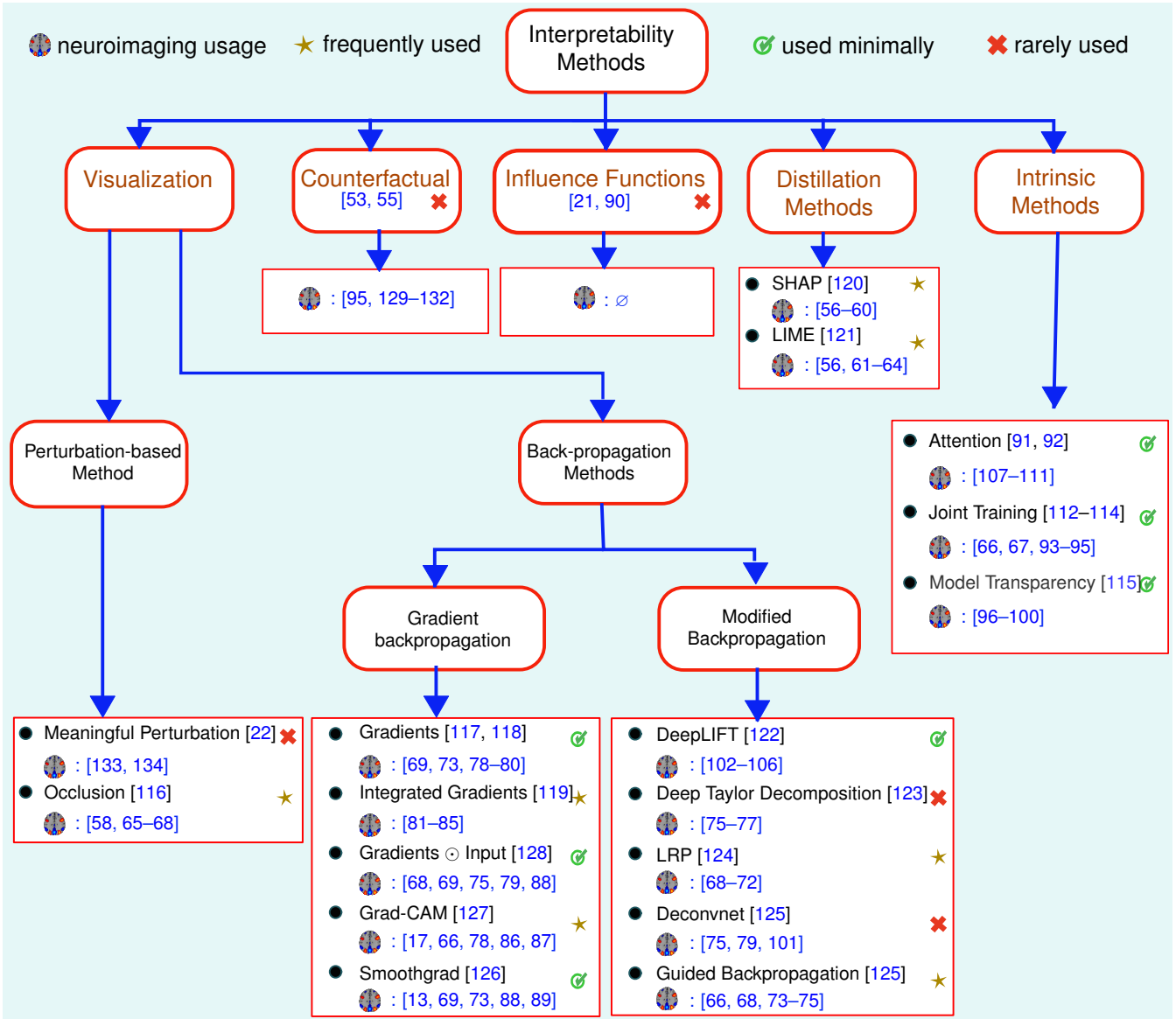
  
\fontfamily{cmr}\selectfont
\renewcommand{\rmdefault}{cmr}\rmfamily

  \caption{Taxonomy of Explainable AI. The figure depicts all the major branches of interpretability approaches and the methods within each branch with references to the studies that proposed them to facilitate diverse nature of explanations for AI models. We also complement the figure by showing representative neuroimaging studies that used the aforementioned methods.}
  \label{fig:taxonomy} 
\end{figure}

\fontfamily{cmr}\selectfont
\renewcommand{\rmdefault}{cmr}\rmfamily

To precisely define the interpretability methods, we define an \emph{input} as a vector $\vx \in \sR^{d}$. We also define the model as a function $F: \sR^{d} \rightarrow \sR^{C}$, where $C$ is the number of classes in the classification problem. Moreover, let us also assume that the mapping $F_{c} (\vx): \sR^{\d} \rightarrow \sR$ defines the class-specific logit, where $c$ is the predicted class. An explanation method generates an \emph{explanation map} $E : \sR^{d} \rightarrow \sR^{d}$ that maps $\vx$ to a saliency map of the same shape, highlighting the important regions influencing the prediction. 

\subsection{Visualization Methods}
\label{vis_methods}

As defined earlier, visualization methods highlight the most influencing regions of the input that drive the model's output. Generally, visualization methods for model interpretability fall under two main categories. The first category is {\em Backpropagation Methods}, also called {\em Sensitivity Methods}, and the latter category is {\em Perturbation-Based Methods}, also called {\em Salience Methods}~\cite{ancona2019gradient}. Though other methods (e.g., LIME and SHAP) may still use visualizations to communicate explanations, we omit them from the visualization category because they require a separate interpretable model to generate explanations.

Backpropagation methods are further classified into gradient backpropagation and modified backpropagation methods based on how backpropagation is performed during the computation of saliency maps. 

\subsubsection{Gradient Backpropagation}
In gradient backpropagation, also called {\em sensitivity methods}, we measure how the output score changes with the tiny change in each feature dimension. The sensitivity methods assume this change rate indicates the importance of the corresponding input dimension.

\begin{itemize}[leftmargin=0.25in]

\item {\bf Gradients (GRAD):} Gradient (GRAD)~\cite{baehrens2010explain, simonyan2013deep} is the gradient of the class-specific logit with respect to input features $\vx$. Mathematically, $\ve = \nabla_\vx  \: \mathcal{F}_i (\vx)$, where $\ve$ is the vector representing the feature importance estimate for each input variable in the sample. In fact, it determines the input features for which the least perturbation will end up with the most change in the target response. However, gradients are usually noisy indications of attribution~\cite{montavon2017explaining, samek2016evaluating, smilkov2017smoothgrad}. The major pitfall of using gradients is that the partial derivative $\nicefrac{\partial \mathcal{F}_i (\vx)}{\partial x_k}$ is not independently related with $x_k$ but also with other input dimensions. Furthermore, the concept of saliency does not apply to the linear classifier because saliency is independent of the input for linear models.

\item {\bf Gradient $\odot$ Input:}  Gradient $\odot$ Input~\cite{shrikumar2016not} was introduced to improve the sharpness of the attribution maps obtained through sensitivity analysis. However, Ancona et al. \cite{ancona2017towards} showed that {\bf Gradient $\odot$ input} becomes equivalent to DeepLIFT and $\epsilon$-LRP, if the network has only ReLU activation functions and no additive biases. This point-wise multiplication was initially justified to sharpen the gradient explanations. However, it is better justified when the measure of salience is a priority over mere sensitivity~\cite{ancona2019gradient}.

\item {\bf Integrated Gradients (IG):} Integrated Gradients~\cite{sundararajan2017axiomatic} is an attribution method that satisfies {\em implementation invariance} and gives one estimate per feature. IG uses the interpolation technique to integrate importance at different discrete intervals between uninformative baseline, say $\bar{\vx}$ and the input $\vx$, to give an integrated estimate of feature importance. The feature importance based on integrated gradients is computed as follows:
\begin{equation}
\ve = (\vx - \bar{\vx}) \times \sum_{i = 1}^{k} \frac{\partial \mathcal{F}_i(\bar{\vx}+\frac{i}{k} \times (\vx - \bar{\vx}))} {\partial \vx} \times \frac{1}{k}
\end{equation}
The ultimate estimate $\ve$ depends on the value of $k$ (number of intervals) and the choice of a suitable uninformative baseline $\bar{\vx}$. IG also satisfies {\em sensitivity-N} axiom since $\sum_{i=1}^n R^c(x_i) = \mathcal{F}_i(\vx) - \mathcal{F}_i(\bar{\vx})$

\item {\bf Smooth-Grad (SG):} Smoothgrad~\cite{seo2018noise, smilkov2017smoothgrad} expresses a feature as an averaging of $N$ noisy estimates obtained when input is perturbed with some Gaussian noise $\rvepsilon$, expressed as:
\begin{equation}
\ve = \frac{1}{N} \sum_{j=1}^{N} \nabla_{\vx+\rvepsilon} \, \mathcal{F}_i (\vx+\rvepsilon)\mbox{, where }\rvepsilon \sim \mathcal{N}(\vzero,\, \vone)
\end{equation}

Other variants~\cite{hooker2019benchmark} of smooth-grad, especially their squared and variance versions, exist in the literature. However, their usage is very limited in model interpretability. 

\item {\bf CAM and GRAD-CAM:} Zhou et al.~\cite{zhou2016learning} proposed \emph{Class Activation Map (CAM)} to visualize the focal regions using global average pooling on the last layer activations in convolutional neural networks. Subsequently,  Selvaraju et al.~~\cite{selvaraju2017grad} proposed a gradient-weighted class activation map called Grad-CAM and generalized the CAM computation to a broader set of networks by leveraging the gradients of the last layer activation maps. Indeed, Grad-CAM computes the gradients of the class score (logit) with respect to the last convolution layer. Let $A^k$ be the set of feature maps of size $m \times n$. Grad-CAM computes $\alpha_k^c = \frac{1}{m \cdot n} \sum_{i}^m \sum_{j}^n \frac{\partial y_c}{\partial A_{i, j}^k}$, the gradients of the output with respect to each feature map, and use average pooling of the gradients to assign a score to the feature map.  Finally, it takes the weighted combination of the feature maps followed by ReLU only, i.e., $\relu(\sum_k \alpha_k^c A^k)$, to consider the positive influence on the class of interest. As Grad-CAM visualization is in the feature map space, Grad-CAM explanation is first upsampled to the input resolution using bilinear interpolation and then overlaid on the input image. Grad-CAM is sometimes combined with Guided backpropagation for pixel-space visualization through an element-wise product called Guided Grad-CAM. Several variants of Grad-CAM, such as GRAD-CAM++~\cite{chattopadhay2018grad} and Score-CAM~\cite{wang2020score}, have been proposed to improve upon Grad-CAM. 


\end{itemize}

Kapishnikov et al. also proposed two approaches~\cite{kapishnikov2019xrai, kapishnikov2021guided}, called \emph{eXplanation with Ranked Area Integrals (XRAI)} and \emph{Guided IG}, that can refine the results of integrated gradients and can produce improved explanations. However, their usage in neuroimaging studies is still minimal. 

\subsubsection{Modified Backpropagation}

Modified backpropagation category refers to the methods that use different forms of backpropagation other than standard backpropagation. The modification can be based on how gradients should flow backward when the ReLU layer is encountered, such as in guided backpropagation and DeConvNet methods. Another trend is to use relevance backpropagation instead of gradients, such as in layer-wise relevance propagation and deep Taylor decomposition methods. 

\begin{itemize}[leftmargin=0.25in]

\item {\bf Guided Backpropagation (GBP):} Guided backpropagation~\cite{springenberg2014striving} modifies the gradients during backpropagation to make it consistent with ReLU activation functions. Let $\{f^{l}, f^{l-1}, \dots, f^{0}\}$ be the input and output features maps of the ReLU activations during the forward pass of a DNN. Also, let $\{R^{l}, R^{l-1}, \dots, R^{0}\}$ be the intermediate gradients during the backward propagation. Precisely, the forward ReLU function at the intersection of $l-1$ and $l$-th layers is defined as $f^{l} = \relu(f^{l-1}) = \max(f^{l-1}, 0)$ and Guided backpropagation overrides the gradients of ReLU functions. The unique purpose of this modification is to allow only non-negative gradients during backpropagation. Mathematically, 

\begin{equation}
\mR^{l} = 1_{\mR^{l+1}\, >\, 0} 1_{f_l \, > \, 0}\, \mR^{l+1}
\end{equation}
That is, GBP considers only positive activations with respect to ReLUs and positive gradients from the earlier step during backward propagation. Figure \ref{fig:all_cams} shows example saliency maps produced using variants of Grad-CAM and Guided Backpropagation methods.  

 \begin{figure} [!htbp]
 \includegraphics[width=1\linewidth]{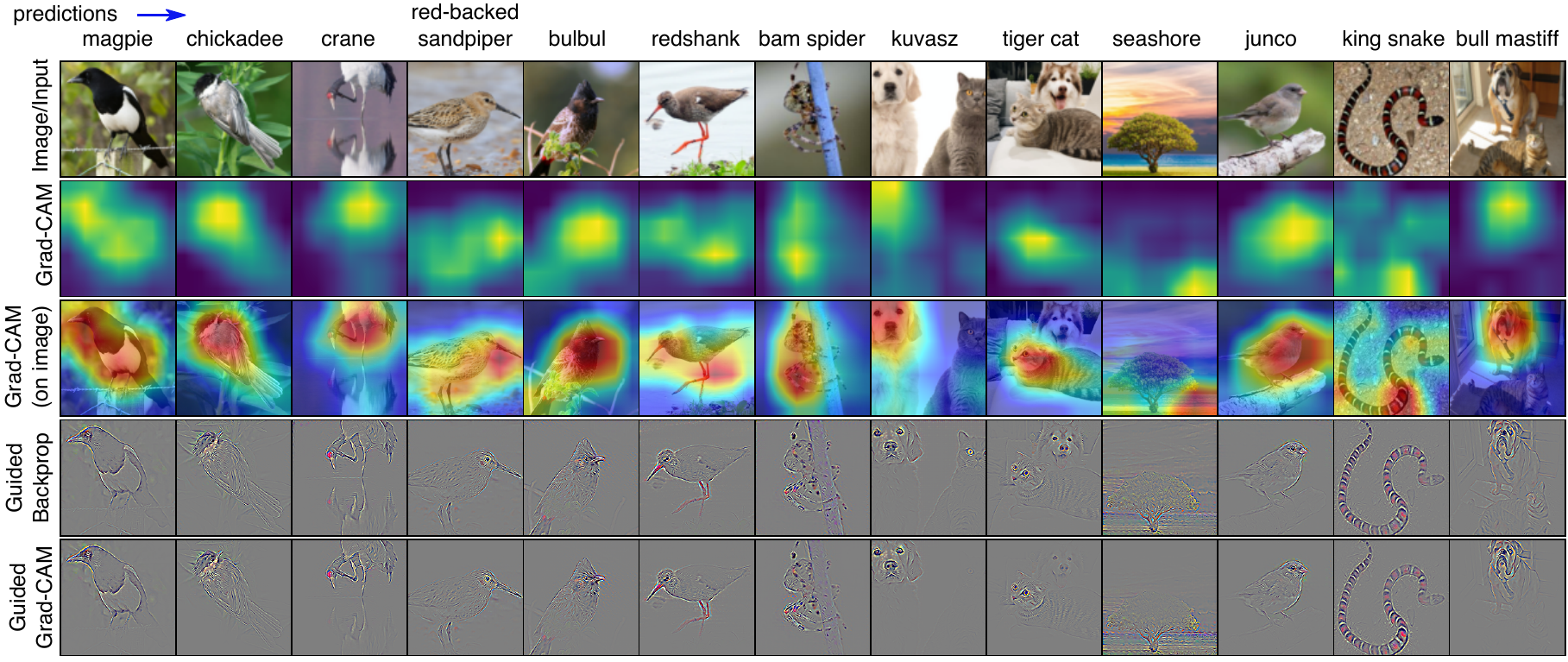}
 \centering
 \caption{Explanations generated using Grad-CAM, Guided Backpropagation, and Guided Grad-CAM methods for \emph{Resnet 50} (trained on ImageNet) model predictions. }
 \label{fig:all_cams}
 \end{figure}

\item {\bf DeConvNet: } 
DeConvNet~\cite{springenberg2014striving} is another "guided" method but slightly differs from the \emph{Guided Backpropagation} in that it only passes "positive" gradients from the upper to the lower layer when the ReLU layer is encountered. The use of DeConvNet to interpret models in neuroimaging domain is very limited. 

\item {\bf Layer Relevance Propagation ($\epsilon$-LRP):} Layer relevance propagation~\cite{bach2015pixel} uses the term "relevance" denoted as $r_i^{(l)}$ to refer to the relevance of the unit $i$ in layer $l$. It starts at target neuron $c$ in the last layer $L$ and treats the target neuron's activation as its relevance. The relevance of all other neurons in layer $L$ are set to $0$. Subsequently, during backward propagation, it computes attributions for neurons at other layers using a recursive $\epsilon$-rule as described in Eq. \ref{recRule}. Let $z_{ij} = x_i^{(l)} w_{ij}^{(l,  \, l+1)}$ be the weighted activation of unit $i$ in layer $l$ onto neuron $j$ in the next layer, $b_j$ be the additive bias for the unit $j$ and $\epsilon$ be the small numerical constant to ensure stability. The final attribution for the $i$-th input is defined as $R_i^c(\vx) = r_i^{(1)}$.
\begin{align}
 r_i^{(L)} =
 \begin{cases}
                                  \mathcal{F}_i(\vx) & \hspace{0.5cm} \text{if unit $i$ is the target neuron} \\
                                  0  &  \hspace{0.5cm} \text{otherwise}
 \end{cases}
\end{align}
Layer relevance scores are backpropagated and distributed according to the following rule:
\begin{align}
\label{recRule}
 r_i^{(l)} = \sum_{j} \frac{z_{ij}}{\sum_{i^{\prime}} z_{i'j} + b_j + \epsilon.\sign(\sum_{i'} z_{i'j} + b_j)} r_j^{(l+1)}
\end{align}

Ancona et al.~\cite{ancona2017towards} showed that $\epsilon$-LRP is equivalent to the feature-wise product of the input and the modified partial derivative. Readers may refer to the study~\cite{samek2016evaluating} for variants of LRP dealing with improved numerical stability. 

\item {\bf DeepLIFT Rescale:} DeepLIFT ({\em Deep Learning Important FeaTures}) assigns attributions to each unit $i$ based on activations using original input $\vx$ and baseline input $\bar{\vx}$~\cite{shrikumar2017learning}. Similar to LRP, DeepLIFT Rescale assigns attribution through backward propagation. Let $\bar{z}_{ij}$ be the weighted activation of neuron $i$ in layer $l$ into neuron $j$ in the next layer and defined as $\bar{z}_{ij} = \bar{\vx}_i^{(l)} w_{ij}^{(l, \, l+1)}$.  The rule for assigning attributions during the backward pass is described in Eq. \ref{deepLiftEq}. The intended attribution for the $i$-th input is defined as $R_i^c(\vx) = r_i^{(1)}$. Baseline reference values are created based on a forward pass with input $\bar{\vx}$.
\begin{equation}
 r_i^{(L)} =
 \begin{cases}
                                 \mathcal{F}_i(\vx) - \mathcal{F}_i(\bar{\vx}) & \hspace{0.5cm} \text{if unit $i$ is the target neuron} \\
                                  0  &  \hspace{0.5cm} \text{otherwise}
 \end{cases}
\end{equation}
The attributions are backpropagated according to the following rule:
\begin{equation}
\label{deepLiftEq}
 r_i^{(l)} = \sum_{j} \frac{z_{ij} - \bar{z}_{ij}}{\sum_{i^{\prime}} z_{i'j} - \sum_{i^{\prime}} \bar{z}_{i'j}} r_j^{(l+1)}
\end{equation}
DeepLIFT Rescale generalizes the concept of $\epsilon$-LRP with no assumption of the baseline or a particular choice of non-linearity. In other words, $\epsilon$-LRP becomes equivalent to DeepLIFT if the baseline is $\zero$ and only {\em ReLU} or {\em Tanh} is used in the network with no additive biases. DeepLIFT and $\epsilon$-LRP replace the gradient of the non-linearities with their average gradient. However, this replacement does not apply to discrete gradients. Hence the overall computed gradient of the function may not be the average gradient of the function as a whole. Due to this constraint, DeepLIFT and $\epsilon$-LRP do not satisfy {\em implementation invariance}. DeepLIFT was originally designed for feed-forward networks, and Ancona et al. \cite{ancona2017towards} showed that DeepLIFT is a good approximation of Integrated Gradients for feed-forward networks.

\item {\bf Deep Taylor Decomposition:} Montavon et al.~\cite{montavon2017explaining} proposed another relevance backpropagation approach to pass relevance from the output to the input space. This backpropagation of relevance is similar to LRP but uses a different formulation using first-order Taylor expansion.  

\end{itemize}

\subsubsection{Perturbation-Based Methods:}
\label{secPerturbBased}

In perturbation-based methods, also called {\em salience methods}, the marginal effect of a feature on the output score is computed relative to the same input where such a feature is absent.  

\begin{itemize}[leftmargin=0.25in]

\item {\bf Occlusion Sensitivity:} Zeiler and Fergus (2014)~\cite{zeiler2014visualizing} proposed a perturbation-based approach called \emph{Occlusion Sensitiveity} to measure the sensitivity of the output score when some regions in the input image are occluded. This approach is also known as \emph{Box Occlusion} because of using a grid or box structure during occlusion. Precisely, this method occludes different portions of the input with a grey square and expects a significant drop in classification score if the portion is strongly discriminative for the prediction the model has made. Figure \ref{fig:all_majors} shows example heatmaps generated using popular post hoc methods for Inception v3 model (trained on ImageNet samples) predictions.

 \begin{figure} [!htbp]
 \includegraphics[width=1\linewidth]{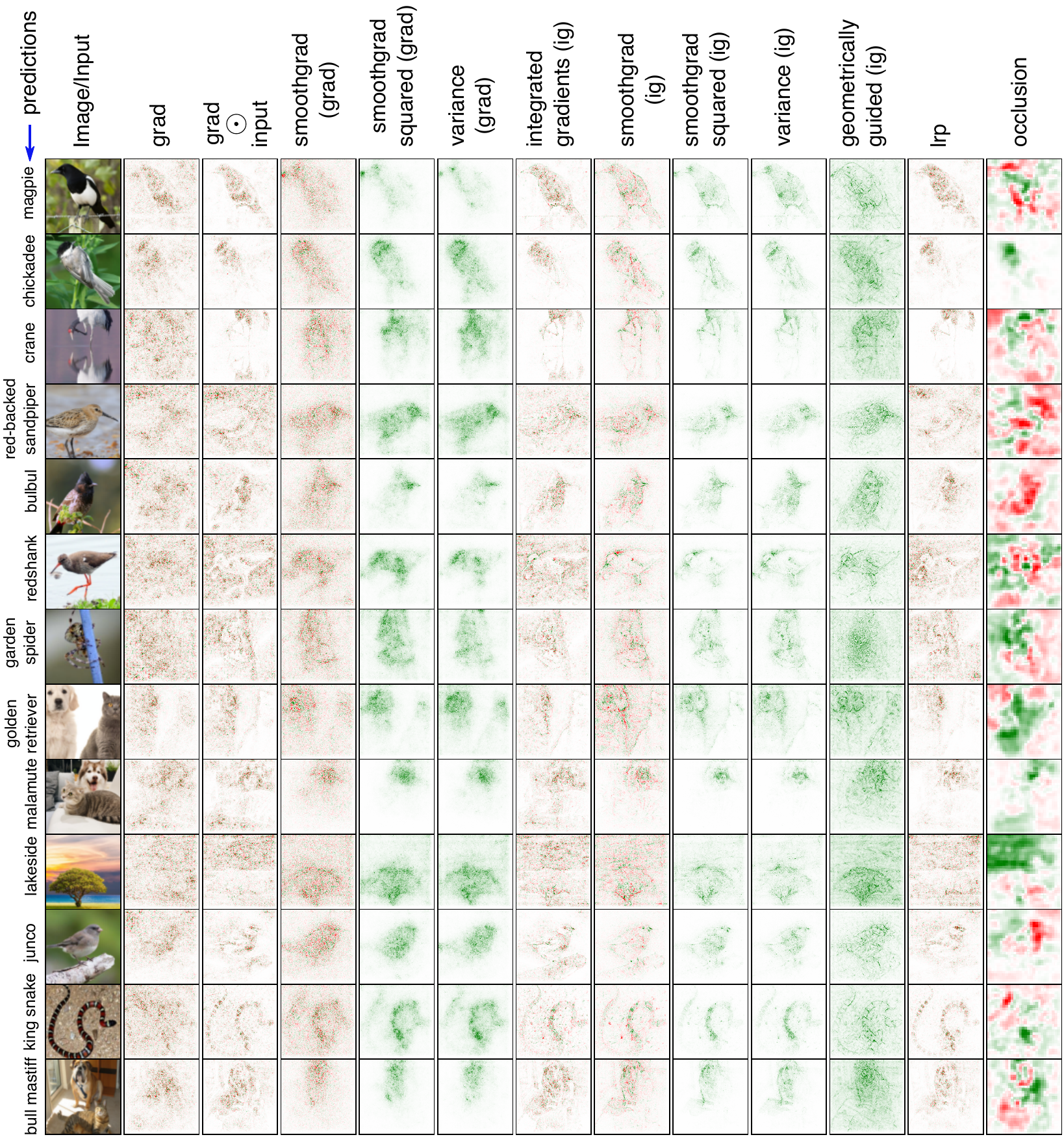}
 \centering
 \caption{Explanations generated using popular interpretability methods for \emph{Inception v3} model predictions. Most of the methods (except layer relevance propagation and occlusion sensitivity) are based on standard gradient backpropagation. LRP, however, backpropagates, relevance from top to the bottom layers using pre-defined rules. Occlusion, as mentioned, relies on the perturbation of the input using a moving small square grid.}
 \label{fig:all_majors}
 \end{figure}

\item {\bf Meaningful Perturbation:} Fong and Vedaldi (2017)~\cite{fong2017interpretable} proposed a model-agnostic generalization of gradient-based saliency that uses input perturbations and integrates information obtained through all backpropagation.  Suppose the input image be $\vx_0$ and $f(\vx) \in \sR^{C}$. The goal is to find the smallest deletion mask $m: \Lambda \to [\,0, 1] \,$ for which the classification score drops very significantly, i.e., $f_c(\Phi(\vx_0; m)) \ll f_c(\vx_0)$, where $\Phi(\vx_0; m)$ is perturbation operator. The problem of finding the minimum deletion mask is defined as the following optimization problem:
\begin{equation}
m^\ast = \argmin_{m \in [ \,0, 1] \,^\Lambda} \lambda \|\1- m\|_1 + f_c(\Phi(\vx_0; m))
\label{meanPerturb}
\end{equation}
$\lambda$ is a regularizing parameter that enforces small deletion to generate a highly informative region to explain the prediction. This optimization problem is solved using the gradient descent technique. 

\end{itemize}

Gradient-based methods are fast, easy to implement, and readily applicable \cite{sundararajan2017axiomatic} to existing models compared to perturbation-based methods. However, gradient-based methods are extremely noisy, usually affected by high-frequency variations, and may not represent the model's decision-making process. In contrast, perturbation-based methods are directly interpretable (because it computes the marginal effect), model-agnostic, and do not require accessing the internal operations of the models. 

While the major advantage of perturbation-based methods is the direct computation of the marginal effect of each feature or a small subset of features, the obvious limitations are that the perturbation methods are very slow compared to gradient-based methods. Moreover, they must choose the number of input features to perturb at each iteration and the perturbation technique because the explanations depend heavily on these hyperparameters. Ideally, for realistic reasons, it is not possible to test perturbations of all possible subsets. Moreover, there is no rigorous theoretical foundation to choose from the available perturbation techniques, thus making the explanations unreliable. 

\subsection{Distillation Methods}
\label{distil_methods}

In distillation methods, a separate \emph{explanation model}, also called \emph{interpretable model}, is required to explain the decision of the original model. This approach is model-agnostic, and the interpretable model does not need the internal behavior of the model. As a separate model is used to extract the essential aspects of the original model, this process is called distillation. However, similar to visualization methods, distillation methods may still produce visualization as explanations. 

\begin{figure} [!htbp]
 \includegraphics[width=1\linewidth]{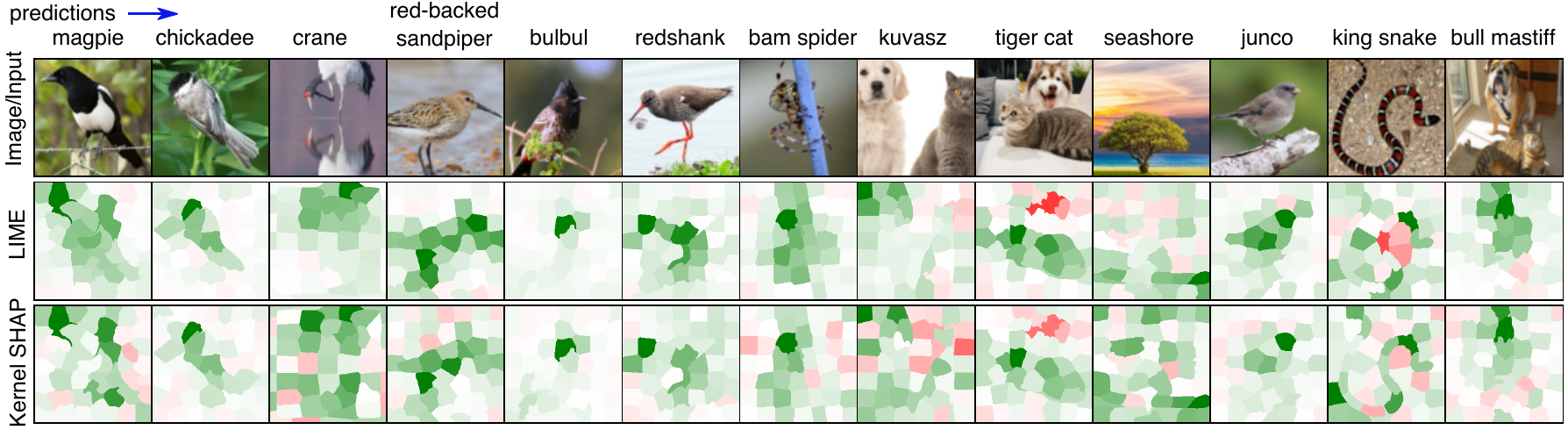}
 \centering
 \caption{Explanations generated for \emph{Resnet 50} model predictions using distillation methods---LIME and Kernel SHAP for ImageNet samples. Feature mask was generated using simple linear iterative clustering (SLIC) of scikit-image. We used 1000 iterations to build the interpretable model.}
 \label{fig:distillation}
 \end{figure}

\begin{itemize}[leftmargin=0.25in]

\item {\bf LIME:} LIME~\cite{ribeiro2016should}, also called {\em Local Interpretable Model-agnostic Explanations}, is based on a surrogate model. The surrogate model is usually a linear model constructed based on different samples of the main model. It does this by sampling points around an example and evaluating models at these points. LIME generally computes attribution per sample basis. It takes a sample, perturbs multiple times based on random binary vectors, and computes output scores in the original model. It then uses the binary features (binary vectors) to train an interpretable surrogate model to produce the same outputs. Each of the coefficients in the trained surrogate linear model serves as the input feature's attribution in the input sample. Let $\vx=h_\vx(\vx')$ be a mapping function between "interpretable inputs" ($\vx'$) and "original inputs" ($\vx$).  Also, let $\vx' \in \{ 0, \, 1\}^M$, $M$ be the number of simplified features, and $\phi_i \in \R$. The local interpretable explanation model is defined as: 
\begin{equation}
g(\vx') = \phi_0 + \sum_{i = 1}^{M} \phi_i x'_{i}
\end{equation}
The explanation model $g$ can be obtained by solving the following optimization problem: 
\begin{equation}
\label{lime}
\xi = \argmin_{g \, \in \, \mathcal{G}} L(f, g, \pi_{\vx'}) + \Omega(g)
\end{equation}
$g(\vx')$ and $f(h_{\vx}(\vx'))$ are enforced to be equal. That is, $L(f, g, \pi_{\vx'})$ determines how unfaithful $g$ is when it approximates $f$ in the vicinity defined by the similarity kernel $\pi_{\vx'}$. $\Omega$ penalizes the complexity of $g$ and the Equation \ref{lime} can be solved using penalized linear regression. One of the major issues with LIME is robustness. LIME explanations can disagree if computed multiple times. This disagreement occurs mainly because this interpretation method is estimated with data, causing uncertainty. Moreover, the explanations can be drastically different based on kernel width and feature grouping policies. 

\item {\bf SHAP:} 
Historically, Shapley values are computed in a cooperative game theory to calculate the marginal contributions of each player. The computation of this marginal effect relies on game outcomes of all possible sets of coalitions. Suppose $P$ be a set of $N$ players and a function $\hat{\mathsf{f}}$ that maps any subset $S \subseteq P$ of players to a game score $\hat{\mathsf{f}} (S)$. This score is obtained when the subset $S$ of players participated in the game. The Shapley value is a way to compute the marginal contribution of each player $i$ for the game outcome $\hat{\mathsf{f}} (P)$---the outcome obtained when all players $P$ participated in the game.
\begin{equation}
R_i = \sum_{S \subseteq P \setminus \{i\}} \frac{|S|! ( |P|-|S|-1)!}{|P|!} [\,\hat{\mathsf{f}} (S \cup \{i\}) - \hat{\mathsf{f}} (S)]\, 
\end{equation}
The problem with Shapley values is that this attribution technique is computationally intractable when the number of players is large. Lundberg and Lee~\cite{lundberg2017unified} proposed a regression-based, model-agnostic formulation of Shapley values called SHapley Additive exPlanations (SHAP). This approach is also known as Kernel SHAP and is widely used to compute SHAP explanations. As SHAP ranks the features based on their influence on the prediction function, the occurrence of overfitting is usually reflected in the provided explanation. In fact, Kernel SHAP removes the need to use heuristically chosen parameters as used in LIME to recover SHAP values. Refer to Figure \ref{fig:distillation} to see few examples of generated LIME and Kernel SHAP explanations for \emph{Resnet 50} model (trained on ImageNet) predictions. 

\end{itemize}

LIME and SHAP could also be treated as perturbation-based methods because they both perturb the original input locally to build separate interpretable models. However, as described here, the category of perturbation-based methods does not rely on a separate interpretable model. Hence, LIME and SHAP belong to a separate category for their model-agnosticism and the usage of a separate model.

\subsection{Intrinsic Methods}
\label{intrinsic_methods}

Intrinsic methods focus on interpretation as part of the model design or training rather than doing a separate post hoc analysis. These methods are model-specific and are usually implemented based on different design or training perspectives. While some shallow models, such as linear models and decision trees, are directly interpretable, deep learning models are considered black boxes, and their internal functions are quite inscrutable. However, there are many doors to obtain intrinsic interpretability in DL, such as \emph{attention mechanism}, \emph{joint training}, and \emph{modular transparency}. In this section, we briefly discuss some of the common practices used to obtain intrinsic interpretability in deep learning.

\subsubsection{Attention Mechanism}
An \emph{attention mechanism} is a technique generally used in deep learning models which computes the conditional distribution over inputs leading to a vector of weights that specify the importance of different regions in the input for the given context. There are several approaches~\cite{bahdanau2014neural, vaswani2017attention} to compute attention weights for single-modal or multi-modal tasks. The attention mechanism has been proven to improve the deep learning model's performance, and attention weights can be visualized as heatmaps to provide easy-to-understand explanations. 

\subsubsection{Joint Training}
\emph{Joint training} is the concept of training a model simultaneously for performance and explanations~\cite{hendricks2016generating, zellers2019recognition, liu2019towards}. Joint training requires a complex objective function to optimize for the additional explanation task. The additional task may provide a direct textual explanation, generate an explanation association between inputs or latent features and human-understandable concepts, or learn semantically meaningful model prototypes~\cite{xie2020explainable}. A very high-level view of joint training optimization can be as follows:

\begin{equation}
\argmin_{\theta} \frac{1}{N} \sum_{i=1}^{N} \alpha \, L(\vy_n, \vy') + L(\ve_n, \ve')
\end{equation} 

The arguments $\vy_n$ and $\vy'$ refer to model output and output label, respectively. $\ve_n$ and $\ve'$ refer to model explanation and explanation label, respectively. 

\subsubsection{Modular Transparency}
\emph{Modular transparency}~\cite{thibeau2022interpretability} refers to a network consisting of multiple modules. The modules have pre-specified design goals and are usually black-boxes. However, the interaction among the modules is transparent. The explanation can be obtained from understanding how the model functions globally. Ba et al.~\cite{ba2014multiple} demonstrated a modular deep learning model constructed with attention mechanism and reinforcement learning for multiple object recognition tasks. The model was inspired by how humans perform visual sequence recognition tasks by continually moving to the next relevant locations, recognizing individual objects, and changing the internal sequence presentation. 

\subsection{Counterfactual Explanations}
\label{counterfact_methods}
Counterfactual explanations, by definition, provide explanations for hypothetical scenarios. Specifically, counterfactual explanations simply ask for the smallest change required to change the model's outcome. This category of explanations is human-friendly~\cite{molnar2020interpretablebook} because they allow humans to choose from multiple options to change the scenarios. Wachter et al.~\cite{wachter2017counterfactual} proposed a single-objective optimization method to generate a counterfactual explanation. 

\begin{equation}
L(\vx, \vx', y',  \lambda) = \lambda \cdot (\hat{f}(\vx') - y')^2 + d(\vx, \vx')
\end{equation}

The inequality $|\hat{f}(\vx') - y'| \le \epsilon$ determines the tolerance between the current and the counterfactual predictions. The parameter $\lambda$  balances the distance in prediction and the distance between original and counterfactual instances. $d(\vx, \vx')$ is the distance between the original instance $\vx$ and the counterfactual $\vx'$ measured as weighted Manhattan distance as defined below:

\begin{equation}
d(\vx, \vx') = \sum_{j=1}^{p} \frac{|\evx_j - \evx'_j|}{\text{MAD}_j}
\end{equation}

where $\text{MAD}_j$ is the median absolute deviation of feature $j$. Dandl et al.~\cite{dandl2020multi} proposed a multi-objective formulation of counterfactual explanations. This multi-objective formulation satisfies multiple requirements of counterfactual explanations. Other implementations of counterfactual explanations can be found in~\cite{mothilal2020explaining, karimi2020model}.

\subsection{Influence Functions}
\label{influence_methods}

Studies also proposed a data modeling approach to explaining a model prediction in terms of influence functions~\cite{koh2017understanding, ilyas2022datamodels}. Precisely, these methods attempt to find the representative training samples that influenced the prediction of the test sample. While this area of investigation toward explainability is still at the rudimentary level, few studies~\cite{hampel1986robust, koh2017understanding, basu2020second, zha2020ifme, han2020explaining, feldman2020neural, ghorbani2019data, jia2019towards, ilyas2022datamodels} proposed approaches to determine the influencing training points for a particular test case. While determining influence function is yet to use in neuroimaging research as far as we know, this approach, if carefully leveraged, can lead toward many advantageous use cases, including generating counterfactual explanations~\cite{ilyas2022datamodels} for different neurological disorders. 

Apart from the mainstream interpretability methods, people also attempted visualizing {\bf feature maps} either directly in the convolutional layers or in the input space via optimization techniques~\cite{olah2017feature}.

\section{Axiomatic Properties of Attribution Methods}
\label{secAxioms}

Recent interpretability research spelled out some desirable properties of attribution methods as follows:

\begin{itemize}[leftmargin=0.25in]
\item {\bf Sensitivity(a):} An attribution method satisfies {\bf Sensitivity(a)}~\cite{sundararajan2017axiomatic} if for every input a and baseline that differ in one feature but have different predictions, then the differing feature should be given a non-zero attribution.

\item {\bf Sensitivity(b):} Suppose the function implemented by the deep network does not depend (mathematically) on some variable. In that case, the attribution method is said to be satisfying {\bf Sensitivity(b)}~\cite{sundararajan2017axiomatic} if the attribution to that variable is always zero.

\item {\bf Linearity:} Suppose two deep networks modeled by the functions $f_1$ and $f_2$ are linearly composed to form a third network that models the function $a \times f_1 + b \times f_2$, i.e., a linear combination of the two networks. Then we call an attribution method to be satisfying linearity if the attributions for $a \times f_1 + b \times f_2$ to be the weighted sum of the attributions for $f_1$ and $f_2$ with weights $a$ and $b$ respectively \cite{sundararajan2017axiomatic}.

\item {\bf Explanation Continuity:} Let $S_c(\vx)$ be a continuous prediction function for the input $\vx$ and class $c$. Also, let $\vx_1$ and $\vx_2$ be two nearly identical points in the input space, i.e., $\vx_1 \approx \vx_2$ for which model responses are identical. Attribution methods, to maintain explanation continuity~\cite{montavon2018methods}, should generate nearly identical attributions $R^c(\vx_1)$ and $R^c(\vx_2)$ i.e., $R^c(\vx_1) \approx R^c(\vx_2)$. 

\item {\bf Implementation Invariance:} Let $m_1$ and $m_2$ be two implementations (models) $S_{m_1}(\vx)$, $S_{m_2}(\vx)$ that generate same outputs for the same input $\vx$: $\forall \vx: S_{m_1}(\vx) = S_{m_2}(\vx)$.  An attribution method is called \emph{implementation invariant}~\cite{sundararajan2017axiomatic} if it generates identical attributions when functions $S_{m_1}(\vx)$, and $S_{m_2}(\vx)$ are in the equivalence class for the same input $\vx$. That is, $\forall (m_1, m_2, \vx, c,): R^{c, m_1}(\vx) = R^{c, m_2}(\vx)$ 

\item {\bf Sensitivity-n:} An attribution method satisfies \emph{sensitivity-n} axiom~\cite{ancona2017towards} if the replacement of any subset of features by their non-informative baseline causes the output score to drop by the sum of the attributions previously assigned to those features. Let $\vx_S = \{x_1, x_2, \dots, x_n\} \subseteq \vx$ be the subset of features. Then:
\begin{equation}
\sum_{i=1}^n R^c(x_i) = S(\vx) - S(\vx \setminus \vx_S)
\end{equation}

This \emph{sensitivity-n} property is only applicable to salience methods that measure the marginal effect of input on the output. Ancona et al.~\cite{ancona2017towards} proved that attribution methods (based on gradients), when applied to non-linear models, cannot satisfy \emph{sensitivity-n} property at least for some values of $n$, possibly for the reduced degrees of freedom to capture non-linear interactions. 

\item {\bf Completeness or Summation to Delta:} This is a variant of the {\emph sensitivity}-$n$, also called {\emph sensitivity}-$N$. It constraints the attribution methods to produce attribution that sums equal to the classification score with an assumption that non-informative baseline should produce $S(\bar{\vx}) \approx 0$. This property is denoted as: $\sum_{i=1}^N R^c(x_i) = S(\vx) - S(\bar{\vx})$

\item {\bf Perturbation - $\epsilon$:} This axiom proposed in~\cite{kapishnikov2019xrai} is a relaxed version of \emph{sensitivity-$1$} axiom. Suppose $\{x_1, x_2, \dots, x_n\}$ be the input features. For a given $0 < \epsilon \le 1$, if all the features except $x_i$ are fixed, and removal of $x_i$ causes the output to change by $\Delta y$, then the Perturbation - $\epsilon$ is satisfied if the attribution holds the inequality: $attr(x_i) \ge \epsilon \star \Delta y$. 
\end{itemize}

\section{Evaluation Approaches}
\label{eval}

\subsection{Sanity Checks for Interpretability Methods}
It is generally expected that model explanation methods should be reasonably sensitive to model parameters. Moreover, the people expect that model should map data and the associated label based on the data generation mechanism relevant to the target. So, to understand if the behavior of an explanation method is reasonable or not, Adebayo et al.~\cite{adebayo2018sanity} proposed the following sanity checks:

\noindent
{\bf Model Randomization Test:} 
As a model goes through an intensive training process and learns its parameters during the training process, explanations must be sensitive to the model parameters. For this kind of model randomization test, people use either full randomization or cascading randomization and expect to have varied explanations from the explanations generated using the original (non-randomized) model.

\noindent
{\bf Data Randomization Test:} In this test, training labels are permuted to break the relationship between data and associated labels. A model is trained on these shuffled data and forced to memorize the labels against each training sample. As the model memorizes rather than learning the inherent logical, structural, or causal relationship between data and labels, it performs no better than a random model during inference. However, for any plausible explanation method, the post hoc explanation of this model should be substantially different from the model trained on the original training data. However, this test is extremely time-consuming because a model trained on randomized data takes a long time and customized hyperparameters to achieve reasonable convergence. 

\subsection{Evaluation Metrics}
Human evaluation (qualitative) of explanation methods can be entirely wrong because it is possible to create adversarial samples~\cite{goodfellow2014explaining, szegedy2013intriguing} that can fool the human eye, totally changing the model predictions. For quantitative assessment, we need to define the domain-specific desired properties of the interpretability methods formally. Moreover, we need appropriate quantitative metrics to assess the behavior of an interpretability method. When the generated attributions do not become plausible, it is hard to identify if the problem is due to the model itself or to the interpretability method that generated the attributions. In this section, we present some evaluation metrics proposed in the interpretability literature. 

\subsection{Metrics for Ground-truth Datasets}
Arras et al.~\cite{arras2022clevr} proposed two evaluation metrics that can reliably quantify the explanation methods for the datasets that have ground truths.

\noindent
{\bf Relevance Mass Accuracy:} This metric calculates the proportions of total attributions that reside within the relevance area. 
\begin{equation}
\text{Relevance Mass Accuracy} \, = \frac{\mR_{\textit{within}}}{\mR_{\textit{total}}} \\
\text{with } \, \mR_{\text{within}} = \sum_{\substack{k=1 \\ \text{s.t.} \, p_k \, \in \, GT}}^{|GT|} r_{p_k} \, \text{ and } \mR_{\text{total}} = \sum_{k=1}^{N} r_{p_k}
\end{equation} 

where $r_{p_k}$ is the relevance score for the pixel $p_k$. $N$ is the total number of pixels. $GT$ is the set of all pixels within the relevance area (ground-truth area). 

\medskip
\noindent
{\bf Relevance Rank Accuracy:}  Let $K$ be the number of pixels within the ground truth masks. This metric measures how many high-ranked $K$ pixels are within the relevance area.
Let $\mP_{\text{top K}} = \{p_1, p_2, \dots, p_K \, | \, r_{p_1} >  r_{p_2} > r_{p_3} \dots > r_{p_K} \}$ be the top $K$ pixels sorted in descending order of their attribution values. \emph{Rank Accuracy} is defined as follows: 
\begin{equation}
\text{Relevance Rank Accuracy} \, = \frac{|\mP_{\text{top K}}  \, \cap \, GT|}{|GT|} 
\end{equation} 
The argument $GT$ refers to the set of pixels within the ground-truth region. 

\subsection{Metrics for Real Datasets}
Several studies proposed different measures, such as \emph{Remove And Retrain (ROAR)} ~\cite{hooker2019benchmark}, RemOve And Debias (ROAD), \emph{Accuracy Information Curves, Softmax Information Curves} ~\cite{kapishnikov2019xrai}, \emph{Infidelity, Sensitivity} ~\cite{yeh2019fidelity}, to assess the quality of explanations.

\medskip
\noindent
{\bf Remove and Retrain (ROAR):} Hooker et al.~\cite{hooker2019benchmark} proposed another approach to evaluate the performance of an interpretability method. In this approach, samples are modified based on the post hoc explanations. In particular, the features that receive significant attributions during explanation are removed. The model is trained over the modified training data, and people expect a sharp drop in model performance because important discriminative features are absent from the training data. The method is time-consuming as it requires full retraining of the model. Another pitfall of this evaluation process is that the ROAR metric may produce erroneous evaluations when correlations among features exist and capturing only the subset of correlated features is sufficient for correct prediction \cite{sturmfels2020visualizing}. However, ROAR fails to evaluate the feature relevance correctly in that scenario.

\medskip
\noindent
{\bf log-odds score:} Shrikumar et al.~\cite{shrikumar2017learning} proposed a metric to evaluate the quality of explanations. This method greedily identifies the main contributing pixels to convert the original prediction $c_0$ to some target prediction $c_t$. That is, it removes pixels (20\% of the image) based on descending ranking of  $S_{c_0} - S_{c_t}$. Finally, it measures the change in the log-odds score between $c_0$ and $c_t$ for the original image and the image with pixels removed to get the prediction $c_t$. The greater change in log-odds score implies the greater significance of the removed pixels for the original class and thus better capture the true importance. This metric is not useful for natural images and possibly meaningful for images with a strong structural association as in MNIST.  

\medskip
\noindent
{\bf Area Over MoRF Precision Curve: } Samek et al.~\cite{samek2016evaluating} proposed an evaluation technique for the heatmaps based on the idea of how quickly the function value $f(\vx)$ (probability score) drops if the most relevant regions are perturbed. To achieve this agenda, it creates an ordered set $\mathcal{O}=(\vr_1, \vr_2, \dots, \vr_L)$ based on the importance scores of pixels as assigned by the interpretability method. This procedure follows a region perturbation (most relevant first (MoRF)) process, where gradually, a small rectangular region $m \times m$ surrounding each important pixel location $\vr_p$ is removed by the uniform distribution. The quantity of interest here is termed as Area Over MoRF Perturbation Curve (AOPC). 

\[\text{AOPC} = \frac{1}{L+1} \left\langle \sum_{k=0}^{L} f (\vx_{\text{MoRF}}^{(0)}) - f(\vx_{\text{MoRF}}^{(k)} )\right\rangle_{p(\vx)} \]

Here $\langle . \rangle_{p(\vx)}$ indicates average over all samples in the dataset. The intuition is that if the ranking strongly associates with the class label, the removal will cause a steeper drop in the functional value, causing a larger AOPC. 

Though localization and saliency have different connotations, the quality of a saliency map is often measured as its localization accuracy because they overlap. For example, for a dog image, the localization box usually encapsulates the entire dog without focusing on salient details of the dog. The usage of localization in saliency evaluation is often referred to as weakly supervised localization because neither model training nor post hoc interpretability use localization information. 

\medskip
\noindent
{\bf Smallest Sufficient Regions (SSR):} Dabkowski et al.~\cite{dabkowski2017real} proposed a metric based on the notion of the smallest sufficient region capable of correct prediction. This metric requires maintaining the same classification and finding the smallest possible area of the image. This metric is formally defined as follows: 

\begin{equation}
s(a, p) = \log(\tilde{a}) - \log(p)
\end{equation}

$\tilde{a} = \max(a, 0.05)$, where $a$ is the proportion of the cropped image to the original image. $p$ is the probability of the corresponding object class when the classifier classifies based on the cropped but resized image. The lower value of $s(a, p)$ indicates a better saliency detector because it directly translates the idea of SSR ---less area, greater probability score. However, this metric is not suitable if the model is susceptible to the scale and aspect ratio of the object. Moreover, as this metric depends on rectangular cropping and reports results as a function of the cropped area, this approach highly penalizes if the saliency map is coherently sparse~\cite{kapishnikov2019xrai}. Because, in that case, it may span a larger area of the image than the map, which is locally dense, even with the same number of pixels. However, this is counterintuitive from the human vantage point. Humans tend to have sparse and coherent explanations. Moreover, this imposes a severe challenge because masking creates a sharp boundary between the masked and salient region, causing an out-of-distribution problem for the model. 

\medskip
\noindent
{\bf RemOve And Debias (ROAD):} Rong et al.~\cite{rong2022consistent} proposed an evaluation strategy that overcomes the 99\% computational cost of retraining to evaluate attribution methods. The authors made a useful experimental observation that existing ROAR evaluations based on MoRF (most relevant first) or LeRF (least relevant first) removal strategies are inconsistent in ranking the attribution methods. The authors attributed this inconsistency to the class information leakage through the shape of the removed pixels. To mitigate these unwanted influences, the authors proposed a \emph{Noisy Linear Imputation} operator that debiases the masking effect and removes the need for additional retraining.  

\medskip
\noindent
{\bf Performance Information Curve (PIC):}  Kapishnikov et al. proposed another perturbation-based evaluation metric~\cite{kapishnikov2019xrai}, called \emph{Performance Information Curve} to evaluate the appropriateness of an attribution method. The PIC evaluation builds a saliency-focused image. It starts with a blurred image and combines with a saliency mask thresholded, for example, at x\%, to produce the saliency-focused image. The saliency-focused image is then fed into the model to assess the performance of the attribution. The accuracy/softmax score of the model is then mapped as a function of \emph{Information Level}, i.e., calculated entropy. The entropy is a proxy measure of the information content re-introduced for evaluation. The compressed image size is an approximate proxy for the information content of an image. It normalizes the entropy of the re-introduced image by considering the proportion of the entropy from the original image. The aggregate performance measurement over all the information levels for all samples in the dataset finally generates the PIC. The PIC has two variants: 

\medskip
\noindent
{\bf Accuracy Information Curve (AIC):} For AIC, the x-axis uses normalized entropy values and divides them into several bins. The y-axis reports the accuracy calculated over all the saliency-focused images for each bin of image information level (entropy). 

\medskip
\noindent
{\bf Softmax Information Curve (SIC):} The x-axis uses the same normalized entropy values for SIC. The y-axis reports median scores for the proportion of the original label’s softmax score for the saliency-focused image versus the softmax for the original image.

\subsection{Criticisms of Post hoc Interpretability}

While evaluating interpretability methods, we also need to consider the criticisms or concerns people raised in the interpretability literature. The concept of interpretability is simultaneously considered essential and evasive~\cite{lipton2018mythos, tjoa2020survey}.  One of the obvious characteristics of interpretability is that it focuses on generating some understandable intuitions behind why a particular prediction has been made. Interpretability, however, does not care about how the model arrived at this decision. Particularly, the existing interpretability methods focus on the revelation of different aspects of the model's learned behavior~\cite{krishna2022disagreement, han2022explanation}, not necessarily the way a model functions. A vast amount of studies~\cite{rudin2019stop,leslie2019understanding, loyola2019black, rudin2019we, lakkaraju2020fool, laugel2019dangers, laugel2019issues, slack2020fooling, john2022some, chan2021explainable, laugel2019unjustified, loi2021transparency, john2021critical, hatherleyvirtues, paez2019pragmatic} talked about different pitfalls of post hoc interpretability methods.  People raised questions about the transparency of the DL models and the incapacity of the popular interpretability methods for reliable real-world deployments. 
Rudin (2019)~\cite{rudin2019stop}, for example, criticized attempts to explain black-box models. Instead, she suggested building inherently interpretable models. Rudin also thinks black-box models are not required in AI~\cite{rudin2019we}. Moreover, 
many methods are blamed to be computationally expensive~\cite{lundberg2017unified, zeiler2014visualizing}, unstable~\cite{ribeiro2016should}, model insenstive~\cite{selvaraju2017grad, springenberg2014striving}, noisy~\cite{baehrens2010explain, simonyan2013deep, sundararajan2017axiomatic}. Furthermore, some methods~\cite{bach2015pixel, shrikumar2017learning} are criticized for not satisfying the desirable \emph{implementation invariance}~\cite{sundararajan2017axiomatic} property. In medical imaging contexts, multiple studies~\cite{arun2021assessing, kindermans2019reliability, rudin2019stop} reported the unreliability of saliency maps for localizing abnormalities in medical images. Moreover, we should address some ethical dilemmas as indicated by~\cite{herman2017promise} because inherent human bias may violate the transparency of interpretable systems.

While post hoc interpretability methods have been widely used in different applications and neuroimaging studies, we should always be aware of their usage when safety and trust are our significant concerns. For example, we must accept the explanations wisely when we want to use interpretable DL models to understand how the brain functions or what dynamics are responsible for a particular mental disorder. Generally, people use post hoc interpretability methods without any pre-condition applied to the model's design. Paez~\cite{paez2019pragmatic} argued that model transparency or model approximation is useful for objectively understanding the model. Moreover, it is also a necessary condition to achieve post hoc interpretability. We discuss more on this issue and provide a set of detailed insights and suggestions in Section \ref{suggestions} for future practitioners.

\section{Interpretable Neuroimaging}
\label{interpretable_neuro}

Psychiatric disorders have strong correspondence with underlying complex brain dynamics. These ever-changing dynamics supposedly reflect the progression of these disorders. Identifying the essential, interpretable, non-invasive imaging biomarkers from the dynamics can be a significant breakthrough for early diagnosis, potentially preventing its future progression with the help of new insights the model can gain from the data. As discussed earlier, DL is a powerful data-adaptive technology that can automatically learn from data~\cite{lecun2015deep}. As such, DL can bring breakthroughs in healthcare~\cite{hinton2018deep} via uncovering unforeseen and scientifically valid information about disorders. However, a strong caveat is that it is not an easy task because DL models may find different sets of hidden factors contributing to the same input-output relationship~\cite{hinton2018deep}. While the field of interpretability has advanced rapidly in recent years~\cite{guidotti2018survey,adadi2018peeking,ras2022explainable}, we still need rigorous methods and validation techniques to deploy these models effectively in clinical practices and in advancing scientific understanding of the neurological disorders. 

In the following sections, we discuss the significance of interpretability in neuroimaging studies adopting deep learning approaches. We reviewed more than 300 neuroimaging studies that considered model interpretability as their essential component. We refer to Figure \ref{fig:taxonomy} for a quick reference to some neuroimaging studies utilizing all the prevailing interpretability methods. We reckon that these analyses will be helpful for future neuroimaging practitioners looking for a general guideline. Additionally, we analyzed the recent usage trend of the most prevailing post hoc interpretability methods, which clearly shows their continued acceptance in the neuroimaging community. Finally, we discuss different caveats of interpretability practices and provide insights on how this specialized sub-field of AI can be used wisely and meaningfully.

\subsection{Feature Engineering Approach to Neuroimaging}
\label{feature_eng}

In this section, we discuss the traditional feature engineering and comparatively newer feature learning practices in neuroimaging studies.

One of the crucial challenges of Neuroimaging research is understanding the association between cognitive state and the underlying brain activity~\cite{rahman2022interpreting, thomas2019analyzing}. Traditionally, people use the feature engineering approach with shallow linear interpretable models to tackle these challenges. Feature engineering or feature selection step intends to reduce the dimension of the signals while preserving useful discriminative information. Global feature-based (voxel-based) or regional feature-based approaches are commonly used in neuroimaging for feature selection~\cite{zhang2021explainable}. Ashburner and Friston~\cite{ashburner2000voxel} summarized the advances of voxel-based morphometry (VBM), where voxel-wise parametric statistical tests are conducted to compare the smoothed gray-matter images from the two groups. Kloppel et al.~\cite{kloppel2008automatic} used normalized grey matter segment to classify AD patients from normal cohorts. Saima et al.~\cite{farhan2014ensemble} used the volume of gray matter (GM), the volume of white matter (WM), the volume of cerebrospinal fluid (CSF), the area of the left hippocampus, and the area of the right hippocampus to classify AD from sMRI images based on an ensemble of classifiers. Schnack et al.~\cite{schnack2014can} used gray matter densities (GMD) to model SVM for schizophrenia and bipolar classification using sMRI images. Patel et al. \cite{patel2016classification} proposed a stacked autoencoder for schizophrenia classification. The autoencoder was trained in an unsupervised fashion on 116 active gray matter regions to extract region-specific features. Subsequently, the extracted features were used to train an SVM model. Dluhovs et al.~\cite{dluhovs2017multi} used three imaging features (gray matter, white matter, and modulated GM and WM tissue segments of sMRI scans to feed into SVM classifiers in a distributed setting. Xiao et al.~\cite{xiao2019support} used the cortical thickness and surface area features of 68 cortical regions from sMRI images for the SVM-based classification of schizophrenia. Steele et al. \cite{steele2017machine} used mean grey matter volume and density across 13 paralimbic regions of sMRI scans in SVM based classifier to predict psychopathic traits in adolescent offenders. The regional feature-based approaches intend to summarize the whole brain signal by extracting features from some predetermined regions of interest (ROIs). For example, several studies~\cite{magnin2009support, zhang2011multimodal} divided the whole brain into multiple regions and extracted features from those regions to train machine learning models. The ROIs are predetermined based on prior neurobiological knowledge relevant to the disorders. 

Rashid et al.~\cite{rashid2016classification} used dynamic brain connectivity from resting state fMRI for schizophrenia and bipolar patients classification and showed that dynamic FNC outperforms static FNC. Iddi et al.~\cite{iddi2019predicting} proposed a two-stage approach for predicting AD progression. In the first stage, the authors used the joint mixed-effect model for multiple modalities such as cognitive and functional assessments, brain imaging, and biofluid assays with fixed effects for covariates like age, sex, and genetic risk. In the second stage of prediction, a random forest algorithm is used to categorize the panel of predicted continuous markers into a diagnosis of controls and stages of progression. Many other studies~\cite{shen2010discriminative, venkataraman2012whole, zeng2018multi} used functional network connectivity measured as Pearson's correlation coefficients as features for a range of classifiers. Shen et al. \cite{shen2010discriminative} also used locally linear embedding (LLE) to reduce the dimensionality of the feature space to demonstrate that PCA in place of LLE hardly provides separable data points. For a detailed review of feature reduction techniques, refer to~\cite{mwangi2014review}.

\subsection{Deep Learning Approach to Neuroimaging}
\label{deep_learning}

Feature engineering and shallow models suffer from several limitations: 1) the inherent interpretability of shallow models compromises the capacity to deal with high-dimensional neuroimaging data 2) it prevents the natural understanding of brain dynamics. While standard machine learning models can perform reasonably well on handcrafted features, their performance dramatically drops when trained on raw data because of their inability to learn adaptive features from the raw data~\cite{oh2019classification}.

In contrast, Deep Learning (DL) has gained significant progress in different application areas, especially for computer vision and natural language processing tasks. The primary benefit of DL is that it can independently learn from the data through varying levels of abstraction using a series of nonlinear functions. Importantly, it relieves the need to use error-prone feature engineering phase~\cite{mwangi2014review}, which predominantly relies on some preoccupations with the data that may prevent the natural emergence of significant features. To leverage the capacity of DL in neuroimaging research, researchers have started using DL to reach a new level of understanding of the association between psychiatric disorders and brain dynamics~\cite{plis2014deep, mensch2017learning, sarraf2016classification, nie20163d, milc2020, abrol2021deep, thomas2019analyzing}. 

However, the improved performance of DL comes at the cost of intelligibility—its decision-making process is quite incomprehensible to human beings. While deep learning methods can simultaneously achieve unprecedented predictive performance and potentially lead to identifying idiosyncratic brain regions associated with the disorders, the model may overfit and not generalize well to unseen subjects. Moreover, it may learn unexpected artefactual associations for its predictions. The need for explanations arises from inadequate knowledge of the data and associated data generation mechanism and poor understanding of the model's behavior during training. This lack of intelligibility prevents the widespread deployment of DL models in safety-critical domains such as healthcare, medicine, neuroscience, and self-driving cars, to name a few. 

Evidence from many recent studies reinforces the potential of deep learning toward new knowledge discovery in different domains. For example, several studies~\cite{hicks2021explaining, ghorbani2020deep} have demonstrated that a convolutional deep learning model, when introspected with gradients, smoothgrad, and GradCAM, might reveal crucial medical information from ECG signals. Often, interpretability may assist in identifying if the model has inherited any inherent bias from the data. For example,  Young, Booth, Simpson, Dutton, and Shrapnel~\cite{young2019deep} used GradCAM and Kernel SHAP to show that produced saliency maps pass some sanity checks and can be helpful at least to diagnose potential biases in the models trained for melanoma detection. In another study, Vellido~\cite{vellido2019importance} pointed out the significance of interpretability and visualization in medicine and healthcare. Lucieri et al.~\cite{lucieri2020interpretability} used a concept activation vector (CAV) to show that the deep learning model can encode understandable human concepts and apply the disease-relevant concepts for its predictions in a cancer classification task. 

From the perspective of neuroimaging applications, we must meet the two most crucial challenges to gain a broader level of acceptance of DL as a research and clinically supportive tool: 1) Neuroimaging data is inherently high-dimensional. Studies usually have a small sample size posing $m \ge n$ problem, which is very susceptible to cause overfitting in deep models. 2) DL models are considered as \emph{black box models} because of their multi-level non-linearity and lack of established theory behind their learning mechanism. Consequently, it is hard to establish an association between the predictive cognitive state and the underlying dynamics. In other words, the accuracy may not be representative of the quality of the features used by a model. For example, Lapuschkin et al.~\cite{lapuschkin2016analyzing} demonstrated how a \emph{Fisher Vector} model can learn to choose unintended artifacts for generating predictions. In this specific example, the model used \emph{copyright tag} to predict \emph{"horse"} as all the horse images contain the copyright tag, which turned out to be a characteristic of horses. This kind of phenomenon is entirely unexpected and must be avoided while leveraging deep learning models in medical domains. 

\subsection{Transfer Learning in Neuroimaging}
\label{transfer}

One of the major concerns in neuroimaging studies is the lack of sufficient training samples~\cite{bzdok2015semi, oh2019classification}, which is hostile to the efficient training of DL models~\cite{bansal2022systematic}. This constraint is due to the expensive data collection process in neuroimaging studies~\cite{landis2016coins}. In such a scenario, transfer learning can be a convenient approach to deal with this problem, as reported in several studies \cite{mensch2017learning, thomas2019deep, milc2020, mahmood2019transfer, newell2020useful}. While adapting transfer learning in neuroimaging domain is a harder problem due to the unavailability of transferable tasks and lack of ground truth, formulating a suitable task that supports transferrable representation learning from unrelated neuroimaging datasets is essential to support studies dealing with limited training data. 

Leonardsen et al.~\cite{leonardsen2022deep} proposed a CNN model for brain age prediction and subsequently showed evidence of how a model trained to predict age can learn abstractions of the brain and hence can be useful for a series of downstream tasks. The model was selected from some architectural variants and performed well for brain age prediction. The representations as learned by the model were noticeably predictive compared to a baseline model for different unseen datasets for multiple case-control studies. The authors further studied the deviation of the predicted age from the chronicle age by correlating the brain age delta and different standard measures of MRI images. Eitel et al.~\cite{eitel2019uncovering} emphasized the significance of transfer learning by showing how learned knowledge can be transferred across diseases (AD to MS) and MRI sequences (MPRAGE to FLAIR). However, we argue that transferring knowledge across diseases can be misleading. That is, transferring knowledge from a model trained on Alzheimer's patients to a study to classify MS patients may confuse the downstream model. Instead, we should define a pretext task and apply unsupervised or self-supervised pretraining of the model on a more neutral group (e.g., healthy controls). This knowledge transfer approach, as we think, may result in more interpretable knowledge transfer~\cite{rahman2022interpreting}. Rahman et al.~\cite{rahman2022interpreting} proposed a transfer learning mechanism that uses contrastive learning to pretrain a deep learning model on publicly available healthy subjects of the Human Connectome Project (HCP). The authors showed that the self-supervised pretraining improved performance of three downstream models separately trained to classify (schizophrenia, Alzheimer's disease, and autism spectrum disorder) patients of three disorders with the diverse demographic background. In addition, the improved representations improved the post hoc interpretability of the models.  Oh, et al.~\cite{oh2019classification} argued in favor of deep learning-based approaches compared to the traditional way of building classical machine learning models based only on feature extraction approaches. They incorporated a transfer learning mechanism to transfer knowledge (weights) learned during AD vs. NC classification for the pMCI (progressive mild cognitive impairment) vs. sMCI (stable mild cognitive impairment) classification task. For a more detailed review of how transfer learning has been used in magnetic resonance imaging, we refer to the paper~\cite{valverde2021transfer}.

\subsection{Interpretability in Neuroimaging}
\label{interpretability}

In this section, we discuss the significance of \emph{explainability} for AI models. This \emph{explainability} requirement is even more pronounced for deep learning models because of their black-box nature. In particular, we first provide evidence of how deep learning approaches have been recently used for neuroimaging. We also show how interpretability has been a pivotal area of research to make these models clinically valuable tools. Next, we provide a detailed review of the contexts to which interpretability was applied in neuroimaging and discuss the findings therein. 

\begin{figure}[H]
\centering
\begin{tabular}{ll}
\includegraphics[width=0.48\linewidth]{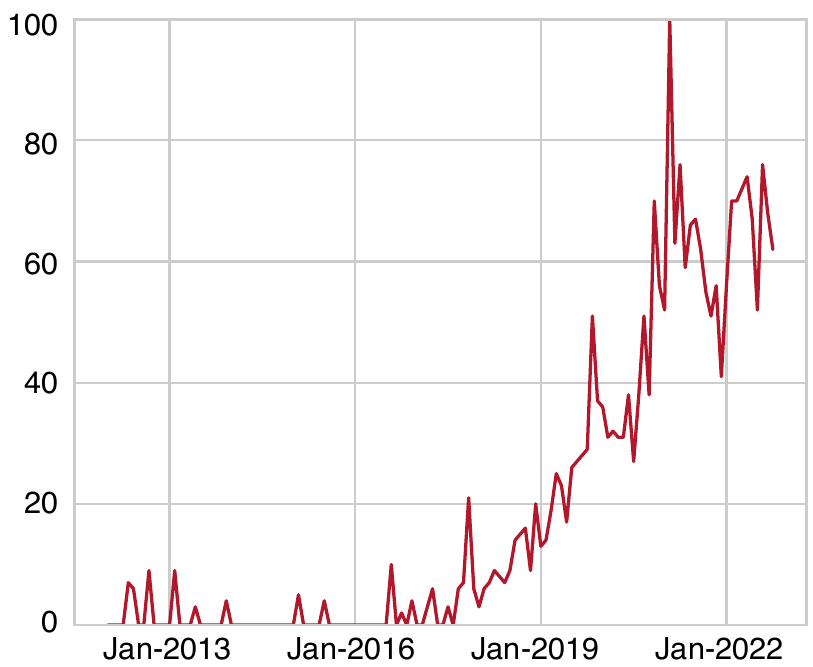}
&
\includegraphics[width=0.48\linewidth]{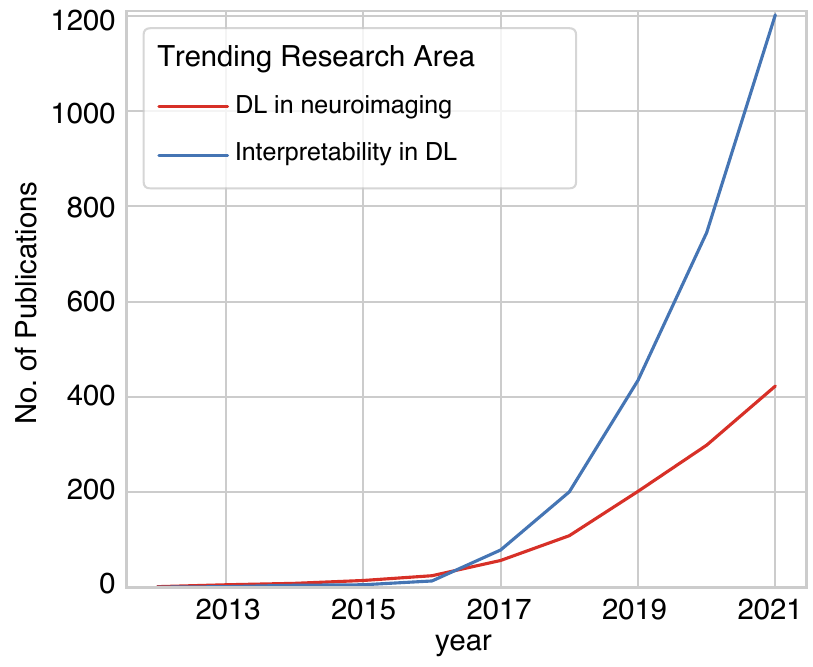}
\end{tabular}
\caption{{\bf Left: } "Explainable AI'' is getting popular or becoming an area of concern over the years (2012 - 2022) as reflected in the Google Trends Popularity Index (Max. value is 100). {\bf Right:} To get relevant statistics, we searched with the keywords "deep learning in neuroimaging" and "interpretability in deep learning" at this website \url{https://app.dimensions.ai/discover/publication} (Accessed on October 13, 2022). Neuroimaging studies increasingly used deep learning models during the last decade (2012 - 2021) to understand the dynamics of brain functions and anatomical structures. The need to interpret black-box models is growing accordingly. }
\label{trendFig}
\end{figure}

"Explainable AI"---a subfield of AI has been very popular because of the recent surge in AI models and algorithms as reflected in the left panel of Figure~\ref{trendFig}. Moreover, deep learning shares a larger part of most recent AI practices. Neuroimaging community has also witnessed a similar surge in deep learning practices in recent years. As DL models are black boxes, the need to interpret the DL models has become essential to validate the models or to advance our understanding of the problem domain, as we can see in the right panel of Figure~\ref{trendFig}. For a quick reference to some neuroimaging studies using popular interpretability methods, readers are advised to refer to the Figure~\ref{fig:taxonomy}.

\section{Review of Interpretability Methods in Neuroimaging}
\label{review}

For the comprehensive review, we group the papers based on the interpretability methods used in those studies. As some studies used several methods in a single study, we mention them at all relevant places. The summary of the review can be accessed from Table~\ref{review_table}.

\begin{landscape}
%
\begin{ThreePartTable}

\renewcommand\TPTminimum{\textwidth} 
\begin{TableNotes} 
\scriptsize
\setlength{\columnsep}{0.8cm}
        \setlength{\multicolsep}{0cm}
        \begin{multicols}{2}\raggedright

\item [{\bf I:}] {\bf Interpretability Methods}
\item[$i^0$] \label{tn:ucm} SA-3DUCM: sensitivity analysis by 3D ultrametric contour map
\item[$i^1$] \label{tn:gap} 3D-ResNet-GAP: 3D-ResNet with global average pooling layer

\item [{\bf V:}] {\bf Validation Methods}
\item [$v^0$]  \label{tn:v0} meta analysis with NeuroSynth~\cite{yarkoni2011large}
\item [$v^1$] \label{tn:v1} CPDB: ConsensusPathDB-human

\item [{\bf S:}] {\bf Studies}
\item [$s^0$] \label{tn:ad} AD: Alzheimer's Disease vs. NC: Normal Controls
 \item [$s^1$] \label{tn:cognitive} seeing images of body parts, faces, places or tools 
 \item [$s^2$] \label{tn:mul} Multiple Sclerosis vs. NC
\item [$s^3$] \label{tn:sz} SZ: Schizophrenia vs. NC
 \item [$s^4$] \label{tn:asd} ASD: Autism Spectrum Disorder vs. NC/TD (typically developing)
  \item [$s^5$] \label{tn:wrat} WRAT: wide range achievement test (LOW/HIGH)
 \item [$s^6$] \label{tn:pd} PD: Parkinson's Disease vs. NC
 \item [$s^7$] \label{tn:hcm} HCM: Hypertrophic cardiomyopathy  vs. NC
 \item [$s^8$] \label{tn:other} AD and other (clinical) variables
\item [$s^9$] \label{tn:ad_all} AD Variants (early/late/stable/progressive/amnestic) vs. NC
 \item [$s^{10}$] \label{tn:path_type} IRF: intraretinal fluid/SRF: subretinal fluid/PED: pigment epithelium detachments

\item [{\bf M:}] {\bf Modality}
\item [$m^0$] \label{tn:smri_all} T1-w/FLAIR/SWI sMRI
\item [$m^1$] \label{tn:complex} complex-valued resting-state fMRI
\item [$m^2$] \label{tn:spect} SPECT DaTSCAN 
\item [$m^3$] \label{tn:whole} WSI: whole slide imaging
\item [$m^4$] \label{tn:genome} Genomic Data

\item [{\bf D:}] {\bf Datasets}
\item [$d^0$] \label{tn:adni} ADNI: Alzheimer's Disease Neuroimaging Initiative
\item [$d^1$] \label{tn:hcp} HCP: Human Connectome Project (HCP S1200 release)
\item [$d^2$] \label{tn:vims} VIMS study
\item [$d^3$] \label{tn:fbirn} FBIRN: Function Biomedical Informatics Research Network 
\item [$d^4$] \label{tn:oasis} Open Access Series of Imaging Studies
\item [$d^5$] \label{tn:abide} ABIDE: Autism Brain Imaging Data Exchange
\item [$d^6$] \label{tn:abide2} ABIDE: Autism Brain Imaging Data Exchange II
\item [$d^7$] \label{tn:pnc} PNC: Philadelphia Neurodevelopmental Cohort
\item [$d^8$] \label{tn:illumina}Illumina HumanHap 610 array, 500 array, Illumina Human Omni Express array
\item [$d^9$] \label{tn:life} LIFE Adult Study~\cite{loeffler2015life}
\item [$d^{10}$] \label{tn:unm} UNM IRB: University of New Mexico Institutional Review Board
\item [$d^{11}$] \label{tn:ppmi} PPMI: Parkinson's Progression Markers Initiative Database
\item [$d^{12}$] \label{tn:1000fc}1000 Functional Connectomes Project
\item [$d^{13}$] \label{tn:abcd} Adolescent Brain Cognitive Development (ABCD) Study
\item [$d^{14}$] \label{tn:uk} UKB: UK Biobank
\item [$d^{15}$] \label{tn:icbm} International Consortium for Brain Mapping database (ICBM)
\item [$d^{16}$] \label{tn:ndar} National Database for Autism Research
\item [$d^{17}$] \label{tn:open} OpenfMRI
\item [$d^{18}$] \label{tn:ucd-adc} UCD-ADC: University of California, Davis Alzheimer’s Disease Center Brain Bank
\item [$d^{19}$] \label{tn:ping} PING: Pediatric Imaging, Neurocognition and Genetics
\item [$d^{20}$] \label{tn:aibl} AIBL: Australian Imaging, Biomarker and Lifestyle Flagship Study of Ageing 
\item [$d^{21}$] \label{tn:fhs} FHS: Framingham Heart Study
\item [$d^{22}$] \label{tn:nacc} NACC: National Alzheimer's Coordinating Center
\item [$d^{23}$] \label{tn:duke} DUKE dataset~\cite{farsiu2014quantitative}
\item [$d^{24}$] \label{tn:retouch} RETOUCH dataset \url{https://retouch.grand-challenge.org/}
\item [$d^{25}$] \label{tn:lpb} LPBA40 dataset~\cite{shattuck2008construction,maier2017isles}
\item [$d^{26}$] \label{tn:mcsa} MCSA: Mayo Clinic Study of Aging participants

\end{multicols}
\end{TableNotes}

\small
\begin{longtable}{@{}p{4.5cm}@{}p{4.75cm}@{}p{3.25cm}@{}p{2.5cm}@{}p{5.5cm}@{}p{4cm}}

\caption{Literature Review of Interpretable Deep Learning Research in Neuroimaging.} \label{table:major_papers} \\

    \toprule
         {\makecell[b]{\bf Authors,\\ \bf Year}}  & {\makecell[b]{\bf Study\\ \bf Objective}}  &  {\makecell[b]{\bf Dataset}}  &  \makecell[b]{\bf Modality}   & {\makecell[b]{\bf Interpretability}} &   {\makecell[b]{\bf Explanation\\ \bf Validation}} \endhead
    \midrule
    
     Yang et al., 2018~\cite{yang2018visual} & {\makecell[l]{AD Classification}} & \makecell[c]{ADNI} & \makecell[c]{sMRI}  & \makecell[c]{Occlusion, SA-3DUCM\\3D-CAM, 3D-GRAD-CAM} & {\makecell[c]{previous reports}} \\  \hline
    
      Rieke et al., 2018~\cite{rieke2018visualizing} & {\makecell[l]{AD Classification}} & \makecell[c]{ADNI} & \makecell[c]{sMRI}  & \makecell[c]{Gradients, Guided Backprop\\Occlusion, Brain Area Occlusion} & {\makecell[c]{AAL atlas\\Euclidean distance}} \\ \hline
    
   Thomas et al., 2019~\cite{thomas2019analyzing} & {\makecell[l]{Cognitive State Prediction}} & \makecell[c]{HCP} & \makecell[c]{rsfMRI} & \makecell[c]{$\epsilon$-LRP} & {\makecell[c]{meta analysis$^{\tnotex{tn:v0}}$}} \\ \hline
   
    Eitel et al., 2019~\cite{eitel2019uncovering} & {\makecell[l]{MS Classification$^{\tnotex{tn:mul}}$}} & \makecell[c]{ADNI and VIMS$^{\tnotex{tn:vims}}$} & \makecell[c]{sMRI} & \makecell[c]{$\epsilon$-LRP} & {\makecell[c]{previous reports}} \\  \hline
    
    Bohle et al., 2019~\cite{bohle2019layer} & {\makecell[l]{AD Classification}} & \makecell[c]{ADNI} & \makecell[c]{sMRI} & \makecell[c]{LRP-$\beta$} & {\makecell[c]{scalable atlas~\cite{bakker2015scalable}}} \\  \hline
    
   Rahman et al., 2022~\cite{rahman2022interpreting} & {\makecell[l]{SZ, AD, and ASD\\Classification}} & {\makecell[c]{\tnotex{tn:fbirn}\hspace{0.55cm}\tnotex{tn:oasis}\hspace{0.55cm}\tnotex{tn:abide}}} & \makecell[c]{rsfMRI} & \makecell[c]{Integrated Gradients (IG)\\Smoothgrad on IG} & {\makecell[c]{RAR framework\\previous reports}} \\ \hline
   
   Zhao et al., 2022~\cite{zhao2022attention} & {\makecell[l]{SZ, ASD Classification}} & {\makecell[c]{In-house, ABIDE}} & \makecell[c]{rsfMRI} & \makecell[c]{LRP} & {\makecell[c]{previous reports}} \\ \hline
   
   Hu et al., 2021~\cite{hu2021interpretable} & {\makecell[l]{WRAT Classification$^{\tnotex{tn:wrat}}$}} & \makecell[c]{PNC and \tnotex{tn:illumina}} & \makecell[c]{nback-fMRI\\genomic data}  & {\makecell[c]{Grad-CAM\\Guided Backprop}}  & {\makecell[c]{previous reports,\\CPDB database$^{\tnotex{tn:v1}}$}} \\ \hline
   
   Chen et al., 2022~\cite{chen2022interpretable} & {\makecell[l]{ASD Classification}} & \makecell[c]{ABIDE} & \makecell[c]{sMRI}  & \makecell[c]{Grad-CAM} & {\makecell[c]{previous reports}} \\ \hline
   
    Yan et al., 2017~\cite{yan2017discriminating} & {\makecell[l]{SZ Classification}} & \makecell[c]{Chinese} & \makecell[c]{rsFMRI}  & \makecell[c]{LRP} & {\makecell[c]{previous reports}} \\  \hline
    
     Levakov et al., 2020~\cite{levakov2020deep} & {\makecell[l]{Brain Age Prediction}} & \makecell[c]{15 open\\databases} & \makecell[c]{T1-w sMRI}  & \makecell[c]{Smoothgrad} & {\makecell[c]{Replicability/Similarity\\/Specificity tests}} \\  \hline
    
     Hofmann et al., 2022~\cite{hofmann2022towards}  & {\makecell[l]{Brain Age Estimation}} & \makecell[c]{LIFE Adult Study} & \makecell[c]{sMRI$^{\tnotex{tn:smri_all}}$}  & \makecell[c]{LRP} & {\makecell[c]{via simulation/atlases\\/signiﬁcance tests}} \\  \hline
    
   Lin et al., 2022~\cite{lin2022sspnet} & {\makecell[l]{SZ Classification}} & \makecell[c]{UNM IRB} & \makecell[c]{cv rsfMRI$^{\tnotex{tn:complex}}$}  & \makecell[c]{Gradients, Grad-CAM} & {\makecell[c]{previous reports}} \\  \hline
   
   Magesh et al., 2020~\cite{magesh2020explainable} & {\makecell[l]{PD Classification}} & \makecell[c]{PPMI} & \makecell[c]{SPECT$^{\tnotex{tn:spect}}$} & \makecell[c]{LIME} & {\makecell[c]{previous reports}} \\  \hline
   
    Zhang et al., 2021~\cite{zhang2021explainable} & {\makecell[l]{AD Classification$^{\tnotex{tn:ad_all}}$}} & \makecell[c]{ADNI-1,2,3} & \makecell[c]{sMRI} & \makecell[c]{Grad-CAM} & {\makecell[c]{previous reports}} \\  \hline
    
   Oh et al., 2019~\cite{oh2019classification} & {\makecell[l]{AD Classification$^{\tnotex{tn:ad_all}}$}} & \makecell[c]{ADNI} & \makecell[c]{sMRI} & \makecell[c]{Gradients} & {\makecell[c]{previous reports}} \\  \hline
   
    Abrol et al., 2020~\cite{abrol2020deep} & {\makecell[l]{AD Classification$^{\tnotex{tn:ad_all}}$}} & \makecell[c]{ADNI} & \makecell[c]{sMRI}  & \makecell[c]{FM visualization, Occlusion} & {\makecell[c]{previous reports}} \\  \hline
    
      Biffi et al., 2020~\cite{biffi2020explainable} & {\makecell[l]{HCM, AD Classification}} & \makecell[c]{multi-site, ADNI} & \makecell[c]{sMRI}  & \makecell[c]{FM visualization} & {\makecell[c]{previous reports}} \\  \hline

    Martinez et al., 2019~\cite{martinez2019studying} & {\makecell[l]{AD Classification$^{\tnotex{tn:other}}$}} & \makecell[c]{ADNI} & \makecell[c]{sMRI}  & \makecell[c]{FM visualization} & {\makecell[c]{previous reports}} \\  \hline

Leming et al., 2020~\cite{leming2020ensemble} & {\makecell[l]{ASD, Gender, Task v. Rest\\Classification}} & \makecell[c]{\tnotex{tn:adni}\hspace{0.55cm}~\tnotex{tn:abide}\hspace{0.55cm}~\tnotex{tn:abide2}\hspace{0.55cm}~\tnotex{tn:1000fc}\hspace{0.65cm}~\tnotex{tn:abcd}\\~\tnotex{tn:uk}\hspace{0.55cm}~\tnotex{tn:icbm}\hspace{0.65cm}~\tnotex{tn:ndar}\hspace{0.65cm}~\tnotex{tn:open}} & \makecell[c]{fMRI} & \makecell[c]{AM, Grad-CAM} & {\makecell[c]{previous reports}} \\ \hline

Tang et al., 2020~\cite{tang2019interpretable} & {\makecell[l]{A$\beta$ pathologies\\Classification}} & \makecell[c]{UCD-ADC} & \makecell[c]{WSI$^{\tnotex{tn:whole}}$}  & \makecell[c]{Guided Grad-CAM\\Occlusion Sensitivity} & {\makecell[c]{previous reports}} \\  \hline

Ball et al., 2021~\cite{ball2021individual} & {\makecell[l]{Brain Age Prediction}} & \makecell[c]{PING} & \makecell[c]{T1-w sMRI} & \makecell[c]{SHAP} & {\makecell[c]{previous reports}} \\ \hline

Jin et al., 2020~\cite{jin2020generalizable}  & {\makecell[l]{AD Classification$^{\tnotex{tn:ad_all}}$}} & \makecell[c]{In-house, ADNI} & \makecell[c]{sMRI} & \makecell[c]{Attention} & {\makecell[c]{Correlation Analysis}} \\  \hline

Zhu et al., 2022~\cite{zhu2022interpretable} & {\makecell[l]{AD Classification$^{\tnotex{tn:ad_all}}$}} & \makecell[c]{ADNI} & \makecell[c]{sMRI}  & \makecell[c]{Joint Training} & {\makecell[c]{predictive performance,\\previous reports}} \\  \hline

Qiu et al., 2020~\cite{qiu2020development} & {\makecell[l]{AD Classification}} & \makecell[c]{\tnotex{tn:adni}\hspace{0.45cm}~\tnotex{tn:aibl}\hspace{0.65cm}~\tnotex{tn:fhs}\hspace{0.65cm}~\tnotex{tn:nacc}} & \makecell[c]{sMRI}  & \makecell[c]{Model Transparency} & {\makecell[c]{neuropathological and\\neurologist-level}} \\ \hline

Uzunova et al., 2019~\cite{uzunova2019interpretable}  & {\makecell[l]{Pathology Images, Types$^{\tnotex{tn:path_type}}$,\\and Lesion Classification}} & \makecell[c]{\tnotex{tn:duke}\hspace{0.65cm}~\tnotex{tn:retouch}\hspace{0.65cm}~\tnotex{tn:lpb}} & \makecell[c]{retinal OCT\\Brain MRI} & \makecell[c]{VAE-perturbation\\Grad-CAM, Guided Backprop} & {\makecell[c]{visual inspection\\quantitative validation}} \\ \hline

Dyrba et al., 2020~\cite{dyrba2020comparison} & {\makecell[l]{AD Classification$^{\tnotex{tn:ad_all}}$}} & \makecell[c]{ADNI} & \makecell[c]{T1-w}  & \makecell[c]{Deconvnet, Gradient $\odot$ Input,\\Deep Taylor Decomposition,\\LRP, Grad-CAM, Guided BP} & {\makecell[c]{previous reports}} \\ \hline

Gupta et al., 2019~\cite{gupta2019decoding} & {\makecell[l]{AD Classification$^{\tnotex{tn:ad_all}}$}} & \makecell[c]{ADNI} & \makecell[c]{rs-fMRI}  & \makecell[c]{DeepLIFT} & {\makecell[c]{quantitative evaluation,\\previous reports}} \\  \hline

Oh et al., 2022~\cite{oh2022learn} & {\makecell[l]{AD Classification$^{\tnotex{tn:ad_all}}$}} & \makecell[c]{ADNI} & \makecell[c]{sMRI}  & \makecell[c]{(counterfactual)\\conditional GAN} & {\makecell[c]{qualitative and\\quantitative analysis}} \\  \hline

Parmar et al., 2020~\cite{parmar2020spatiotemporal} & {\makecell[l]{AD Classification$^{\tnotex{tn:ad_all}}$}} & \makecell[c]{ADNI} & \makecell[c]{rs-fMRI}  & \makecell[c]{FM visualization} & {\makecell[c]{visual inspection}} \\ \hline

Lombardi et al., 2021~\cite{lombardi2021explainable} & {\makecell[l]{Brain Age Prediction}} & \makecell[c]{ABIDE I} & \makecell[c]{T1-w sMRI}  & \makecell[c]{SHAP, LIME} & {\makecell[c]{intra-consistency,\\inter-similarity,\\correlation analysis}} \\  \hline

Lian et al., 2020~\cite{lian2020attention} & {\makecell[l]{AD Classification$^{\tnotex{tn:ad_all}}$}} & \makecell[c]{ADNI-1, 2/AIBL} & \makecell[c]{T1w sMRI} & \makecell[c]{Attention via CAM} & {\makecell[c]{previous reports}} \\  \hline

Bass et al., 2020~\cite{bass2020icam} & {\makecell[l]{Lesion Detection\\AD, Age Classification}} & \makecell[c]{\tnotex{tn:adni}\hspace{0.5cm}~\tnotex{tn:hcp}\hspace{0.5cm}~\tnotex{tn:uk}} & \makecell[c]{T1, T2 sMRI} & \makecell[c]{joint training} & {\makecell[c]{norm. cross correlation\\previous reports}} \\  \hline
   
 Ravi et al., 2022~\cite{ravi2022degenerative}  & {\makecell[l]{CN/MCI/AD\\ 4D MRI Reconstruction}} & \makecell[c]{ADNI} & \makecell[c]{T1 sMRI} & \makecell[c]{modular transparency} & {\makecell[c]{qualitative and\\quantitative assessment}} \\  \hline
 
 Gaur et al., 2022~\cite{gaur2022explanation} & {\makecell[l]{Brain Tumor Classification}} & \makecell[c]{MRI~\cite{bhuvaji_dataset}} & \makecell[c]{MRI}  &  \makecell[c]{SHAP, LIME} & {\makecell[c]{Not provided}} \\  \hline

Saboo et al., 2022~\cite{saboo2022deep} & {\makecell[l]{Cognition Prediction}} & \makecell[c]{MCSA} & \makecell[c]{sMRI,\\Diffusion MRI}  & \makecell[c]{LIME} & {\makecell[c]{prior reports,\\exploratory analysis}} \\ 
    \bottomrule
    \insertTableNotes
    \label{review_table}
  \end{longtable}
  \end{ThreePartTable}
  
\end{landscape}

\subsection{Backpropagation Methods}

\subsubsection{Gradient Backpropagation}

{\bf CAM/Grad-CAM/Guided Grad-CAM}\\
Yang et al.~\cite{yang2018visual} proposed three approaches for generating explanations. One of them, SA-3DUCM (sensitivity analysis by 3D ultrametric contour map), deals with sensitivity analysis of 3D-CNN via a hierarchical image segmentation approach, and the other two methods (3D-CAM, 3D-GRAD-CAM) generate explanations via visualization of network activations on a spatial map. The methods have their own constraints and complement each other. As a baseline method, the authors used occlusion using a cubic neighborhood of $7 \times 7 \times 7$. However, these occlusion methods are not semantically meaningful. The neighborhood size is a hyperparameter and can drastically change the results. Moreover, this method is computationally very expensive. To address these issues, the authors used 3DUCM to produce semantically meaningful, hierarchical, and compact brain segments. Subsequently, they used the occlusion technique based on these segments rather than individual voxels. However, this addition to the baseline occlusion does not consider correlations and interaction among segments. To resolve this, they used 3D Class Activation Mapping (3D-CAM) and 3D-Grad-CAM, which still suffer from the low-resolution problem and may miss the fine details of importance score in the input space. Further analysis of heatmaps reveals that \emph{occlusion} generated heatmaps fail to identify discriminative regions. SA-3DUCM and 3D-CAM are able to identify some regions that match with human expert evaluation.

Hu et al.~\cite{hu2021interpretable} proposed an interpretable DL framework to classify subjects' cognitive ability (low/high WRAT groups) from n-back fMRI data from the PNC cohort. The proposed model can learn from multimodal fusion data and preserve the association across modalities. The authors leveraged Grad-CAM to guide convolutional collaborative learning. This study takes advantage of multimodal fusion from brain FC data and single nucleotide polymorphism (SNP) data. This study intends to extract potentially useful brain mechanisms within and between brain FC and genetics. 264 ROIs were used for brain FC data. The genetic SNP data were collected from the Illumina HumanHap 610 array, the Illumina HumanHap 500 array, and the Illumina Human Omni Express array. The results show that the classifier based on convolutional collaborative learning outperforms the traditional ML classifiers. While it has been evident that all classifiers used some hand-engineered features, the low performance of traditional classifiers might arise from the dimensionality reduction of the original hand-engineered Brain FCs and SNPs. The model identified a large number of significant FCs for the low WRAT (Wide Range Assessment Test) group. In contrast, for the high WRAT group, the model identified a smaller number of significant FCs. The authors used a hypothetical validation technique, which has little empirical significance. This study used ConsensusPathDB-human (CPDB) database as a reference to validate the identified SNPs. The authors provided probable explanations for the identified SNPs, clarifying the model's discriminative behavior.  

\begin{figure} [!htbp]
 \includegraphics[width=1\linewidth]{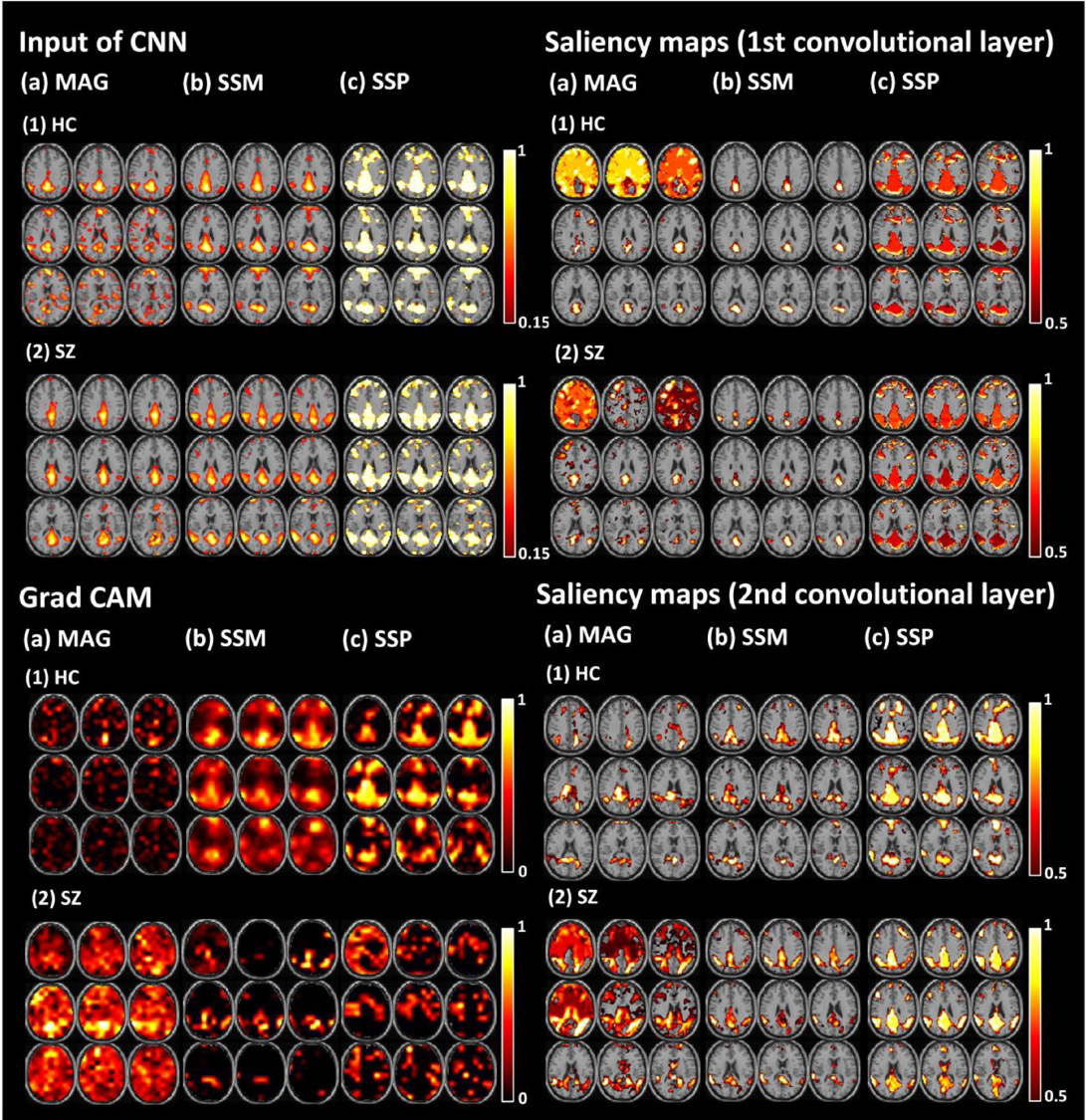}
 \centering
 \caption{Results of DMN saliency maps at the two convolutional layers and Grad-CAM heatmaps for (a) MAG (spatial maps derived from magnitude-only fMRI data), (b) SSM (spatial source magnitude derived from complex-valued fMRI data), and (c) SSP (spatial source phase maps derived from complex-valued fMRI data) extracted from (1) an HC individual and (2) an SZ individual. The input to the network is shown at the upper left panel. SSP, as described by the authors, produced intact but complementary saliency maps at the two convolutional layers. The ﬁrst-layer captured the DMN region edges, whereas the activations inside the DMN were more pronounced within the second layer. Moreover, SSP localized DMN regions with opposite strengths for HC and SZ. On the other hand, SSM and MAG provided maps were inconsistent and had undesirable activation continuity and the noise effects. (Image Courtesy: \cite{lin2022sspnet})}
 \label{fig:grad_methods_usage2}
 \end{figure}

Lin et al.~\cite{lin2022sspnet} proposed a 3D-CNN model to classify schizophrenia patients from normal controls using spatial source phase (SSP) maps derived from complex-valued fMRI data. This study showed the superior performance of SSP maps compared to magnitude maps (MAG) extracted from magnitude-only fMRI data, and spatial source magnitude maps (SSM) separated from complex-valued fMRI data. The authors used two interpretability methods, saliency maps and Grad-CAM, to separately understand the prominent and predictive regions associated with the model predictions. A snapshot of the generated explanations at the subject-level is shown in Figure~\ref{fig:grad_methods_usage2}. While CNN can be a powerful tool for feature extraction and classification, the underlying caveat was the susceptibility of model performance and associate heatmaps because they varied widely according to the number of convolutional layers used. 

Zhang et al.~\cite{zhang2021explainable} proposed a learning framework combining the residual network and self-attention to perform two classification tasks using sMRI images: classifying AD from NC and pMCI from sMCI. This study, in particular, showed that residual networks could learn from sMRI images compared to other variants of convolutional networks (e.g., 3D-VGGNet) and self-attention helps to upgrade the classification performance. The authors applied 3D Grad-CAM to explain individual predictions. One problem with Grad-CAM in understanding the characteristic patterns responsible for predictions is that it cannot capture the fine details in the brain space because of required upsampling. Often, people use convolution layers close to the input layer to increase the resolutions of heatmaps. However, different convolution layers learn different levels of abstraction from the data. So, in that case, explanation maps may not reflect the global behavior of the model. 

Leming et al.~\cite{leming2020ensemble} used a diverse collection of fMRI datasets and leveraged a deep convolutional neural network for three different classification tasks---ASD, gender, and resting/tasks---using functional connectivity (FC). The authors showed that the deep learning model is capable of good classification when datasets are a mixture of multi-site collections. The authors used the 116-area automated anatomical labeling (AAL) parcellation template~\cite{tzourio2002automated} and computed functional connectivity of $4 \times 116 \times 116$ (4 wavelet frequency scales and 116 nodes wavelet coeﬃcient correlation). This study showed that CAM could identify the brain's prominent spatial elements (connectome) that the models used for predictions. In contrast, activation maximization, though initially used to gain intuitions of neural network internals~\cite{erhan2009visualizing}, was able to provide insights into the critical predictive features suitable for classification. However, as the variation of the accuracies of the ensemble was very large, the identified areas may not fully characterize ASD. 

\bigskip
\noindent
{\bf Gradients and Guided Backpropagation}\\
Rieke et al.~\cite{rieke2018visualizing} proposed a 3D-CNN to classify AD patients from healthy controls. The authors used four visualization methods---gradients, guided backpropagation, occlusion, and brain area occlusion---to generate explanations. Relevance scores from gradient-based visualization methods were more distributed across the brain, as opposed to occlusion and brain area occlusion, where relevance scores are more focused on specific regions. Distributive relevance is not feasible for occlusion-based methods because of the limited size of the patch. Hence, the authors recommend using gradient-based approaches for scenarios where distributed relevance is expected. While all four methods focused on some regions considered relevant for AD, such as the inferior and middle temporal gyrus, the distribution of relevance scores varied widely across the patients. In particular, the relevance maps for some patients focused on the temporal lobe, whereas relevance maps for others focused on larger cortical areas. Unlike LRP, as claimed in~\cite{bohle2019layer}, the authors think that similar heatmaps as obtained for both AD and NC are reasonable because a given network should look into similar regions to detect the absence or presence of the disease. To quantify the difference between visualization methods, the authors used Euclidean distance between average heatmaps of the groups (AD or HC) obtained from two visualization methods. Gradient-based methods showed a very small distance.

Oh, et al.~\cite{oh2019classification} proposed a CNN-based end-to-end learning model to perform four different classification tasks classifying various stages of AD (Alzheimer's disease) from NC (normal control) and pMCI from sMCI. The study used a convolutional autoencoder to pretrain the model in an unsupervised fashion. The authors, after prediction, used the saliency method (gradients) to visualize predictive features that the models used for each classification. Analysis of the heatmaps revealed that the temporal and parietal lobes were most discriminative between AD patients and controls.

\bigskip
\noindent
{\bf Integrated Gradients and Smoothgrad} \\
In a recent study, Rahman et al.~\cite{rahman2022interpreting} proposed an interpretable deep learning framework. The framework includes a pre-trainable model suitable for multiple downstream studies with limited data size. The authors also proposed how we can investigate spatio-temporal dynamics associated with mental disorders using post hoc interpretability methods (integrated gradients (IG) and smoothgrad on integrated gradients). Apart from qualitative evaluation, the framework suggested a quantitative evaluation technique, called RAR, to objectively show that identified salient regions are indeed meaningful and highly predictive. This study demonstrates the utility of IG and smoothgrad for neuroimaging interpretability.

Levakov et al.~\cite{levakov2020deep} proposed an ensemble of CNNs and aggregate "explanation maps" to arrive at some conclusive remarks associated with brain age. The authors used smoothgrad as a post hoc interpretability method and were particularly interested in population-wise explanation rather than subject-specific identification of anatomical brain regions. This study also used ensembles of CNN to analyze the model uncertainty behavior. Population-based map for each ensemble was produced by averaging all the volumes in the test set. To generate the global population-based map, they aggregate population-based maps generated for each CNN by taking the median value for each voxel across the ensembles. While this approach highlights important areas in the brain space,  this approach is not able to comment on the direction of influence. It is impossible to determine if the regions contribute positively or negatively to brain age. 

Wang et al.~\cite{wang2022deep} applied Integrated Gradients (IG), LRP, and Guided Grad CAM to visualize CNN models designed for Alzheimer's classification. The authors observed that IG is the best, as revealed in the meta-analysis performed on top of all visualizations. IG heatmaps were particularly more focused on the hippocampus than Guided Grad-CAM and LRP heatmaps, consistent with well-supported biomarkers for Alzheimer's disease. 

\begin{figure} [!htbp]
 \includegraphics[width=1\linewidth]{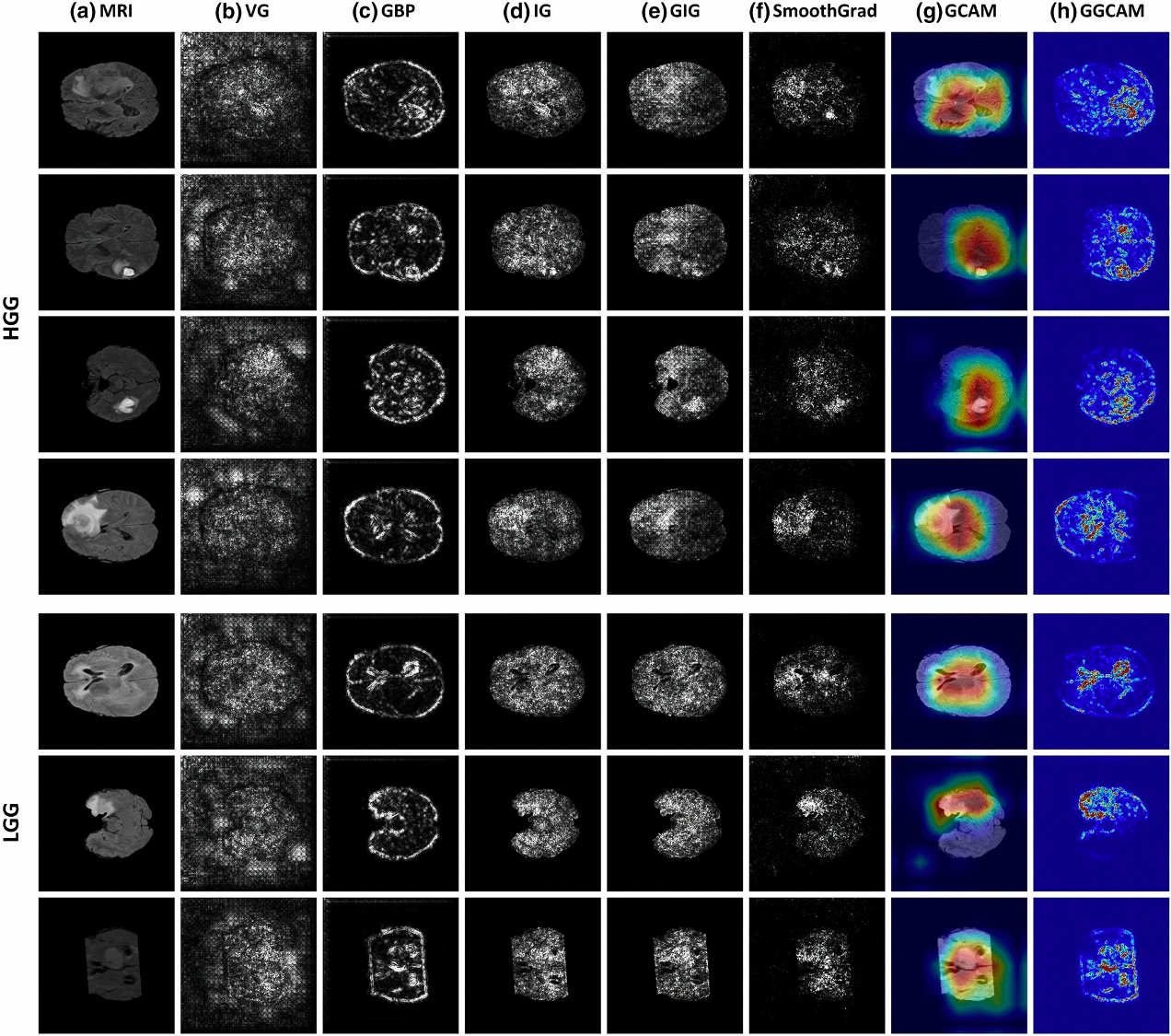}
 \centering
 \caption{Visual explanations generated using popular gradient-based interpretability methods for automatic brain glioma classification. The explanation maps (b - f) highlight contributing salient features, whereas maps in (g, h) highlight salient regions in the input space that drove the predictions. (Image Courtesy: \cite{zeineldin2022explainability})}
 \label{fig:grad_methods_usage}
 \end{figure}

Zeineldin et al.~\cite{zeineldin2022explainability} compared seven popular gradient-based explanation methods: gradients, Smoothgrad, integrated gradients, guided backpropagation (GBP), gradient-weighted class activation map (Grad-CAM), Guided Grad-CAM, and Guided Integrated Gradients for MRI image classification and segmentation tasks. For the brain glioma classification task, Guided Grad-CAM (i.e., combining GBP with GCAM) produced better localization, while Smoothgrad provided the best discriminative regions of the input. For the segmentation task, Smoothgrad was found to be the best choice because of its robustness to noise, while GCAM did the best visualization as it identified the most discriminative regions. Refer to Figure~\ref{fig:grad_methods_usage} for the heatmaps generated using the popular gradient-based methods to explain the predictions made by the brain glioma classification model.

\subsubsection{Modified/Relevance Backpropagation}

\begin{figure} [!htbp]
 \includegraphics[width=1\linewidth]{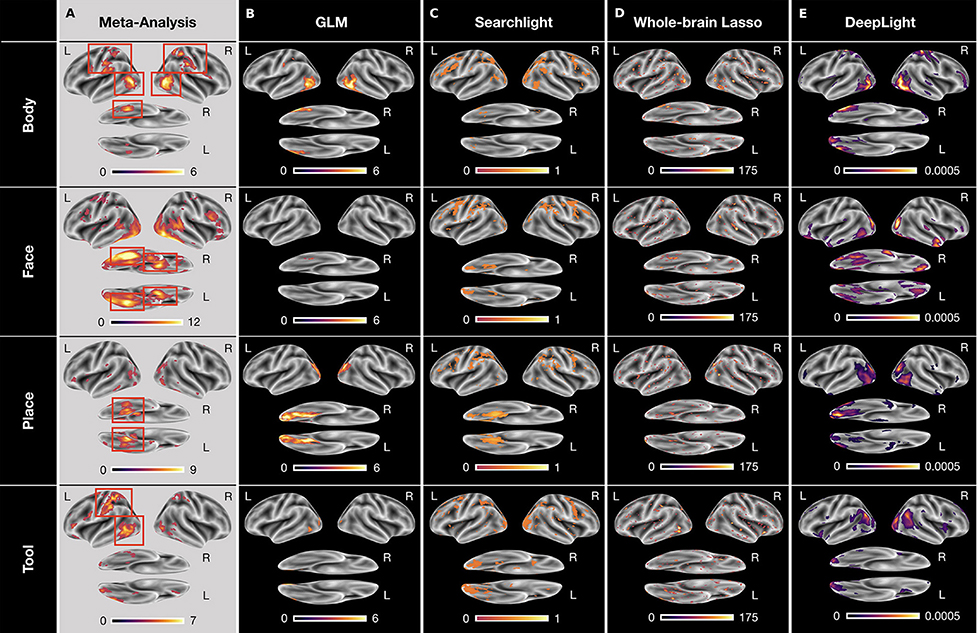}
 \centering
 \caption{Group-level brain maps for DeepLight and other baseline approaches, corresponding to each cognitive state. Column (A) shows the ROIs obtained from a meta-analysis using the NeuroSynth database for the cognitive state terms: "body," "face," "place," and "tools." Columns (B - D) show the results of group-level brain maps using other baseline approaches. Column (E) shows the group-level brain maps from DeepLight. (Image Courtesy: \cite{thomas2019analyzing})}
 \label{fig:lrp_fmri_usage}
 \end{figure}

{\bf Layer-wise Relevance Propagation}\\
DeepLight~\cite{thomas2019analyzing} proposed a DL model consisting of recurrent (LSTM) and convolutional elements to analyze the whole-brain activity associated with cognitive states. Each whole-brain volume is sliced into a set of axial images to feed into the convolutional and recurrent units. To generate post hoc explanations, DeepLight uses LRP (Layer-wise Relevance Propagation)~\cite{bach2015pixel}. The model was trained to predict four different cognitive states corresponding to four stimulus classes (seeing body parts, faces, places, or tools). The baselines used to assess the effectiveness were General Linear Model, Searchlight Analysis, and Whole-Brain Least Absolute Shrinkage Logistic Regression. The model takes each brain volume and then passes through a combination of convolutional and recurrent DL elements to predict the volume corresponding cognitive state. Along the time dimension, it produces a sequence of predictions, one for each sample time point. LSTM here is indeed used for learning spatial dependency within and across the brain slices. After each prediction for each brain volume, the LRP  method is used to generate a post hoc explanation for that prediction attributing relevance to the voxel levels. LRP was used only for the correct predictions. The overall accuracy was around 68.3\% on the held-out dataset. The validation or evaluation of the quality of the maps was achieved through a meta-analysis of the four cognitive states using an established cognitive state-brain association database called NeuroSynth. For relevance analysis, relevance volumes corresponding to a cognitive state were smoothened and averaged to produce a subject-level relevance map. Group-level map for each cognitive state was then produced by averaging subject-level maps relative to a cognitive state and generated for all the subjects in the test set. Figure \ref{fig:lrp_fmri_usage} shows a comparison of group-level maps generated using the DeepLight approach and other baseline approaches. The authors used the meta-analysis with the NeuroSynth database and identified several ROIs associated with each cognitive state. For example, the upper parts of the middle and inferior temporal gyrus, the postcentral gyrus, and the right fusiform gyrus were associated with the body state, the fusiform gyrus and amygdala were associated with the face state, the parahippocampal gyrus was associated with the place state and the upper left middle and inferior temporal gyrus and the left postcentral gyrus with the tool state. Only the top 10\% relevance values were considered for this comparison. While all other baseline approaches were able to identify the association of brain activity to the stimulus classes, the DeepLight model along with LRP was more accurate in finding the association of brain activity to the cognitive states, maintaining greater consistency with the meta-analysis results. 

\begin{figure} [!htbp]
 \includegraphics[width=1\linewidth]{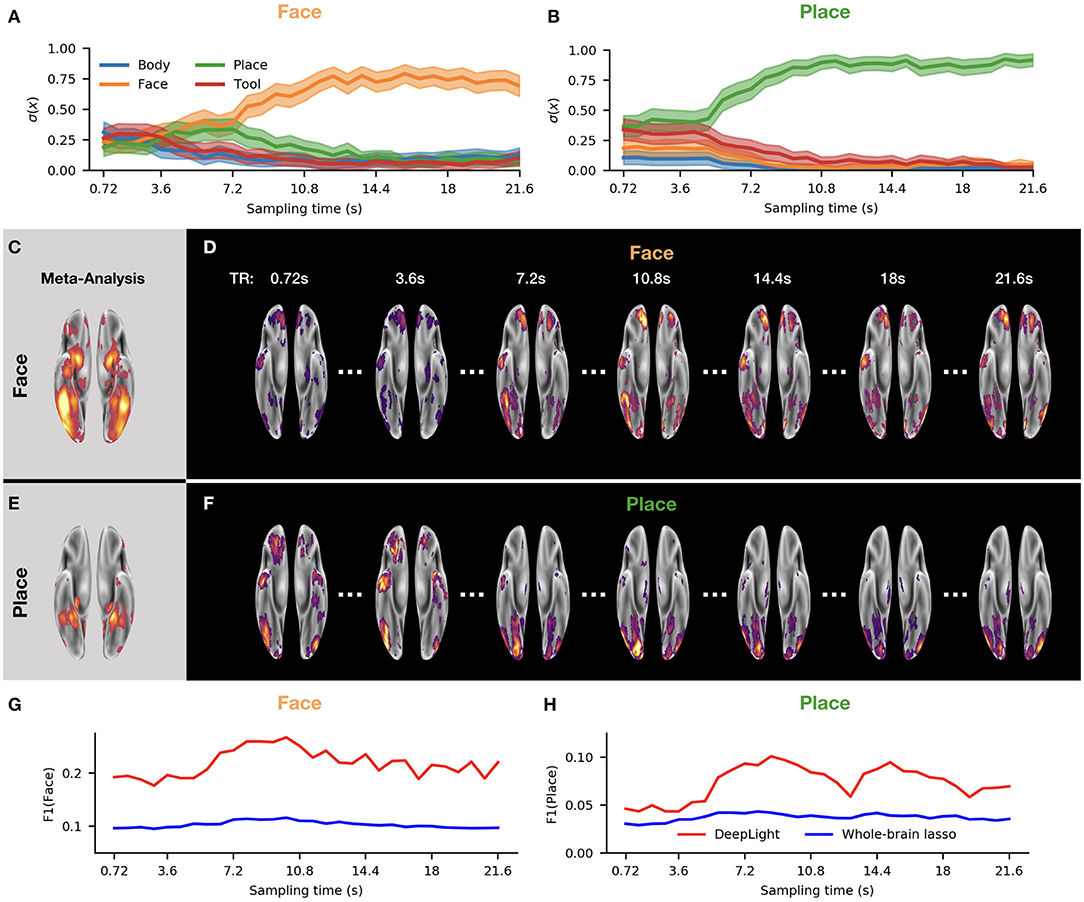}
 \centering
 \caption{Distribution of group-level relevance values over fMRI sampling at different time points of the first experiment block for the two stimuli classes---"place" and "tool." A and B show the average predicted probability of the classifier that the fMRI volume at that specific time point belongs to each of the cognitive states. C and E show the results of the meta-analysis to establish the target maps for these two stimuli classes. D and F show the group-level relevance distribution for different fMRI sampling time points. G and H show the similarity of the brain maps, quantified as F1 score, with the maps obtained from the meta-analysis. (Image Courtesy: \cite{thomas2019analyzing})}
 \label{fig:lrp_fmri_usage_2}
 \end{figure}

Figure \ref{fig:lrp_fmri_usage_2} shows the spatio-temporal distribution of brain activity within the first experiment block corresponding to two cognitive states: place and tool. Particularly, it demonstrates the distribution of group-level relevance values as a function of fMRI sampling time.  As we can observe, while DeepLight was initially uncertain about the cognitive state, its confidence improves very quickly and the relevance maps gradually become steadily similar to the target maps from the NeuroSynth meta-analysis. However, the brain maps of the whole-brain lasso analysis showed very low similarity (F1 score) and did not vary noticeably over time, and hence they have a less meaningful association.

Eitel et al.~\cite{eitel2019uncovering} investigated the possibility of layer-wise relevance propagation (LRP) to uncover the rationale behind decisions made by 3D convolutional neural networks (CNNs) trained to diagnose multiple sclerosis (MS). The identified features revealed that CNN, in conjunction with LRP, has the potential to identify relevant imaging biomarkers, for example, individual lesions, lesion location, non-lesional white matter, or gray matter areas. These biomarkers are considered established MRI markers in MS literature. 

Bohle et al.~\cite{bohle2019layer} used LRP to explain the decisions of a CNN model. They used a scalable brain atlas~\cite{bakker2015scalable} and defined two metrics, "relevance density" and "relevance gain," for objective assessment of the heatmaps. The key reason behind using LRP rather than gradient-based methods is that LRP decomposes the output in terms of contributions in the input space. As the authors mentioned, LRP has the potential to answer this question --- "what speaks for AD in this particular patient?" where explanations using gradient-based approaches apparently address the following question: "which change in voxels would change the outcome most?" We argue that these two questions are not mutually exclusive. For a comparison of LRP with gradient-based methods, the authors used "guided-backpropagation." While both LRP and GB were successful in localizing important regions, GB, compared to LRP, showed less contrast in importance scores between group-wise (AD vs. HCs) heatmaps. Fortunately, there are other gradient-based methods (e.g., integrated gradients \cite{sundararajan2017axiomatic} and smoothgrad~\cite{smilkov2017smoothgrad} on integrated gradients) with desirable properties that future studies may consider for further investigation.

Several studies have attempted to learn from different modalities. For example, Zhao et al.~\cite{zhao2022attention} proposed a hybrid deep learning architecture to combine sequential temporal dynamics (TCs) and functional dependency (FNCs). The authors used an attention module on top of C-RNN to extract temporal dynamic dependencies from TCs and used LRP to identify the most group-discriminative FNC patterns. Please note that LRP was used in a post hoc manner for the analysis of FNC patterns, not as part of the learning process. 

\begin{figure} [!htbp]
 \includegraphics[width=1\linewidth]{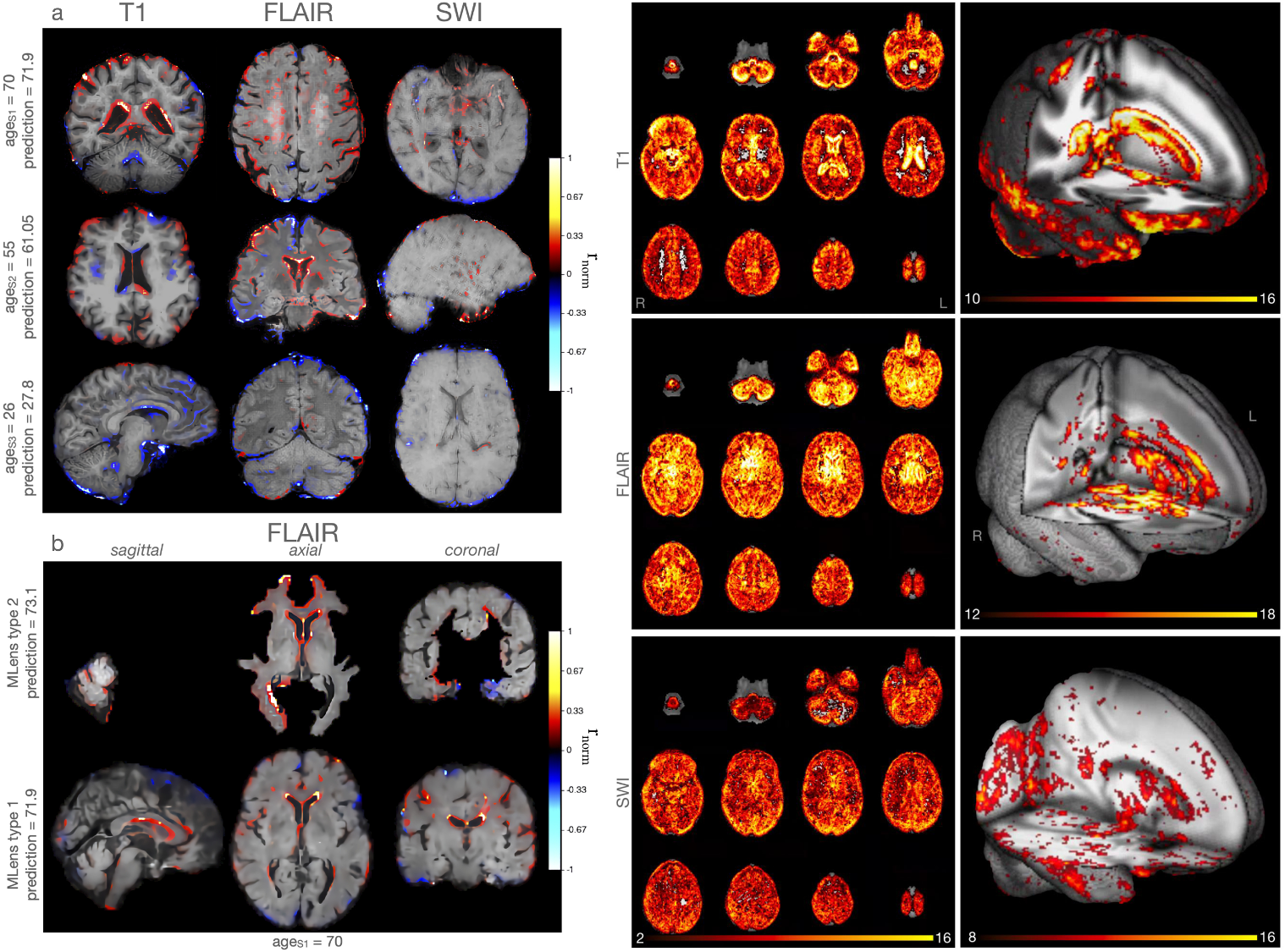}
 \centering
 \caption{{\bf Left:} a) Example individual LRP heatmaps using multi-level ensemble (modality level) trained on whole brain T1, FLAIR, and SWI data. The maps highlight all the crucial brain regions contributing to the subject's age. Relevance maps were produced by aggregating over the base models of each sub-ensemble. b) LRP heatmaps produced using region-based ensembles of FLAIR data (top row) and whole-brain FLAIR data (bottom row). The most pronounced areas across the experiments as found important for the predictions were areas around the ventricles, and subject-specific sulci. {\bf Right:} \emph {Left sub-panel:} T-maps of one-sample t-test over aggregated absolute LRP relevance maps shown as brain slices. The wider t-values (2-16) reveal that the model used information from the entire brain for the predictions. \emph{Right sub-panel:} 3D projection of the t-maps based on thresholded higher t-values for T1, FLAIR, and SWI MRI sequence, respectively. (Image Courtesy: \cite{hofmann2022towards})}
 \label{fig:lrp_usage}
 \end{figure}

Hofmann et al.~\cite{hofmann2022towards} proposed ensembles of convolutional neural networks with LRP to identify which neural features contribute most to brain age. The models were acceptably accurate and could capture aging at both small and large-scale changes.  Refer to Figure \ref{fig:lrp_usage} for the visual explanations. The models were also able to identify associated risk factors in case of diverging brain age. The study detected three major brain components (gray matter, white matter, and cortical spinal fluids) whose relevance scores were linearly correlated to the function of age. The authors argued in favor of ensemble models because the variability of predictions between different models, even when they have the same architecture and are trained on the same data, may arise because of the high variance and bias of individual models. Multiple studies have recommended aggregation of saliency maps generated from single base models~\cite{hofmann2022towards, levakov2020deep}. LRP, similar to other prevailing explanation methods, cannot inform us anything about the underlying biological mechanisms justifiable for the generated explanations. 

\bigskip
\noindent
{\bf DeepLIFT, Deconvnet and Deep Taylor Decomposition}\\
Gupta et al.~\cite{gupta2019decoding} proposed a 5-layer feed-forward neural network to perform three different classification tasks based on functional connectivity features computed for 264 anatomically and functionally diverse ROIs selected from resting-state fMRI: classifying cognitively control subjects from AD and MCI patients. Also, this study performed a harder task of classifying MCI patients from AD patients. 
For identifying salient regions associated with the predictions, DeepLIFT was used to generate explanations. The resulting explanations computed via DeepLIFT were evaluated via a recursive feature elimination process (10\% every time) and retraining the model (5-layer feedforward network) using only the relevant subset of features. For each of the classification tasks, the retrained models achieved higher accuracy compared to the original performance, even with the reduced salient features. 

Dyrba et al.~\cite{dyrba2020comparison} performed a comparison of explanations generated using popular interpretability methods for a 3D CNN model. Precisely, the study compared six interpretability methods: Deconvnet, gradient $\odot$ input, Deep Taylor Decomposition, LRP, Grad-CAM, and Guided Backpropagation. The key observations from this study reveal that the modified backpropagation methods like Deep Taylor Decomposition and LRP could produce clinically useful explanations as they were focused, and aligned with the previous domain reports of the disorders. However, some of the standard backpropagation-based methods, such as Grad-CAM and Guided Backpropagation, are more scattered and did not support the domain knowledge expectedly. The obvious limitations of the study are that the model was not evaluated on an independent dataset. Also, the study does not include a quantitative validation for the generated explanations.

\subsection{Perturbation-based Methods}

{\bf Occlusion Sensitivity}\\
Abrol et al.~\cite{abrol2020deep} experimented with a modified deep ResNet to predict the progression to AD. While the main focus was to predict the progression from MCI class to AD class, the study also experimented with eight combinations of binary, mixed-class (based on transfer learning), and multi-class diagnostic and prognostic tasks. The authors also leveraged network \emph{occlusion sensitivity} to identify the anatomical regions that were most predictive for the progression of MCI to AD. In the analysis, thirteen brain regions, including the middle temporal gyrus, cerebellum crus 1, precuneus, lingual gyrus, and calcarine, consistently emerged in the top 20 most relevant regions. As the \emph{occlusion sensitivity} method considers only the output score drop due to occlusion of a defined region and does not consider connectivity among regions, the method does suffer from several limitations as pointed out by~\cite{yang2018visual, oh2019classification}. The authors also projected the features from the first fully-connected layer onto a 2-dimensional space using t-SNE~\cite{van2008visualizing} to demonstrate the separability of the learned representations. 

\begin{figure} [!htbp]
 \includegraphics[width=0.495\linewidth]{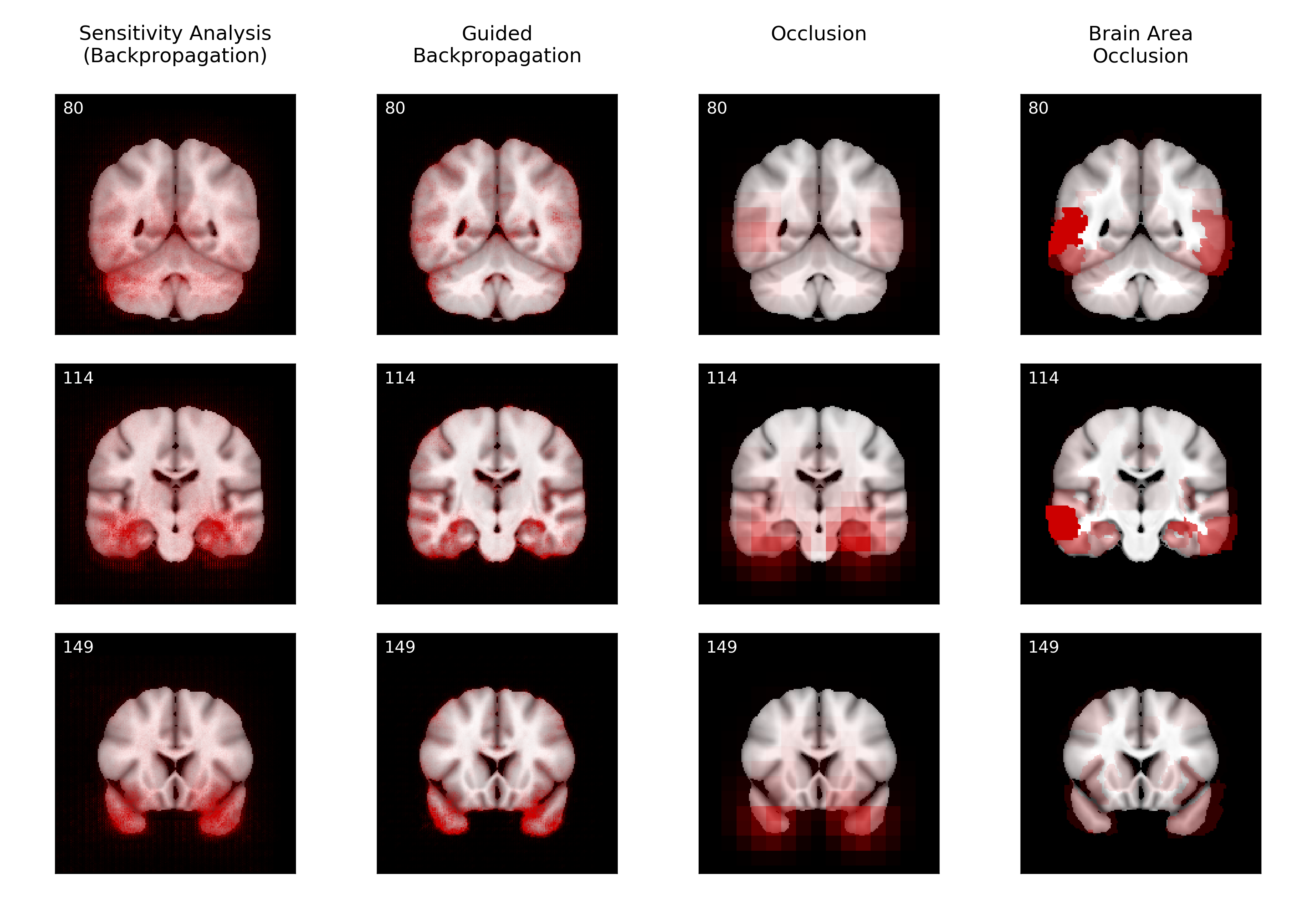}
  \includegraphics[width=0.495\linewidth]{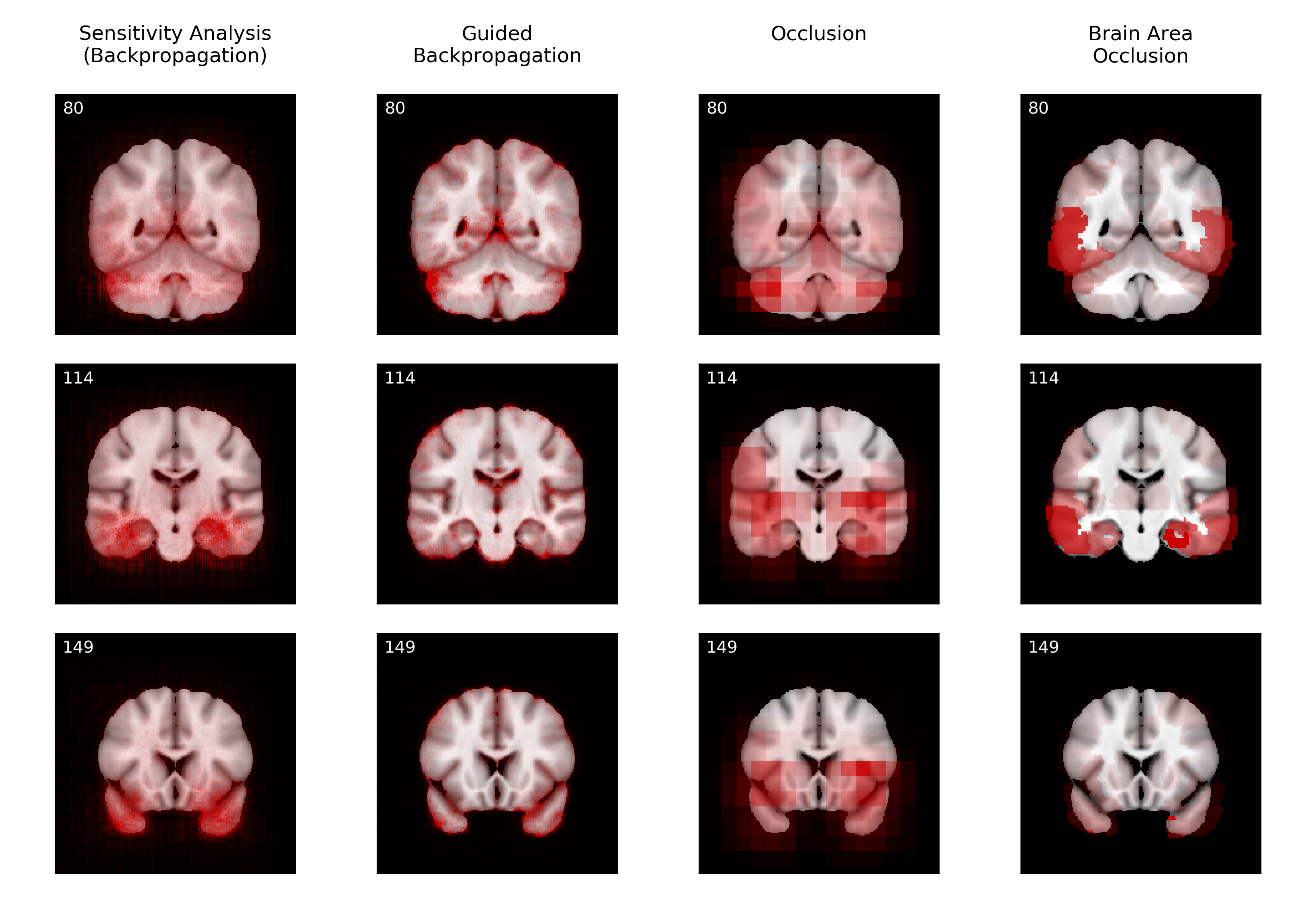}
 \centering
 \caption{Relevance heatmaps generated by four visualization methods: saliency (gradients), guided backpropagation, occlusion sensitivity, and brain area occlusion. The heatmaps shown here are averaged at the group level (AD or NC). The numbers in the image represent the slice indices (out of 229 total coronal slices). The red color indicates the magnitude of importance of that region to drive the model predictions. (Image Courtesy: \cite{rieke2018visualizing})}
 \label{fig:occlusion_usage}
 \end{figure}

It is hypothesized that plaque morphologies can serve as a guide to understanding AD progression and associated pathophysiology. To this end, Tang et al.~\cite{tang2019interpretable} proposed a six-layer convolutional architecture with two dense layers model trained based on whole slide images (WSIs) for the classification of amyloid-beta (A$\beta$) plaques. The authors also provided interpretations of the model decisions using deep learning model introspection techniques. As claimed in the report, the generated explanations aligned with the prior results of A$\beta$ pathology. Apart from different predictive performance estimates, the authors also investigated the interpretability of the model. This study also used two complementary model introspection methods---Guided Grad-CAM and occlusion sensitivity---to demonstrate that the models focused on relevant neuropathological features. As indicated, while Guided Grad-CAM is useful in identifying salient regions responsible for predictions, feature occlusion can reveal the interdependence of class-specific features.  Figure \ref{fig:occlusion_usage} shows examples of how ``occlusion sensitivity'' and other competing gradient-based approaches were used to explain CNN model predictions classifying Alzheimer's disease (AD) patients from normal controls (NC).

\bigskip
\noindent
{\bf Meaningful Perturbation}\\
Meaningful perturbation intuitively attempts to find the important brain regions responsible for the prediction. To do this, it removes certain brain regions and expects a change in model prediction if that region was responsible for the prediction in the first place. However, determining the perturbation nature is not straightforward in the medical domain and can drastically change the input distribution and hence may not be meaningful. Uzunova et al.~\cite{uzunova2019interpretable} proposed a method that generates the meaningful perturbation of the original images. As such, the method produces their closest healthy equivalent using variational autoencoders (VAE)~\cite{kingma2013auto}. The VAE indeed learns healthy variability of the disorder under consideration. Precisely, as shown in Eq \ref{meanPerturb}, for the deletion mask $m: \Lambda \to [\,0, 1] \,$ and each pixel $u \in \Lambda$ with a value $m(u)$, the deletion can be defined using VAE as follows:

\begin{equation}
[ \, \Phi(\vx_0; m) ] \, (u) = m(u)\vx_0(u) + (1-m(u))f_{\textit{VAE}}(\vx_0)
\end{equation}

\begin{figure} [!htbp]
 \includegraphics[width=1.0\linewidth]{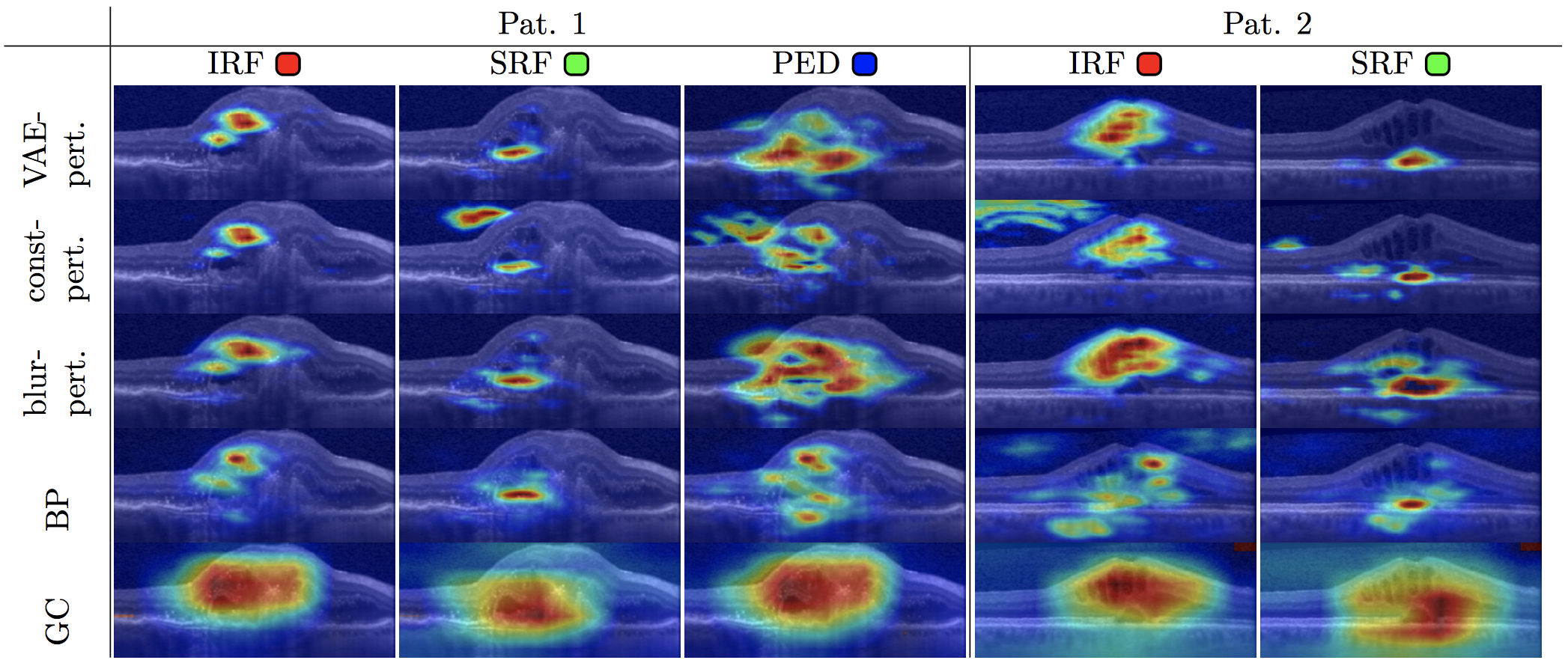}
 
 \centering
\caption{Explanations for two patients by different approaches for multi-label OCT classifier designed to distinguish disease pathologies (intraretinal ﬂuid (IRF), subretinal ﬂuid (SRF), and pigment epithelium detachments (PED)). For the first patient (Pat.1) explanations were generated for three labels, whereas for the second patient, explanations for the two labels were generated. VAE-based perturbation closely aligns with the disease pathologies. Blur perturbation performs fairly well but is only suitable for small structures. Constant perturbation does the worst performance for explanations. Grad-CAM identifies the infected regions but suffers from poor resolution, whereas Guided Backpropagation is noisy and was not found to be class-discriminative. (Image Courtesy: \cite{uzunova2019interpretable})}
 \label{fig:meaning_usage}
 \end{figure}

As VAE was trained based on only healthy subjects, the assumption here is that the VAE knows only the distribution of the healthy subjects in the latent $\vz$ space. So, the reconstruction from pathological images during test time will produce the nearest healthy equivalent. As shown in the patient vs. control classification of retinal OCT images, neural networks do not always learn pathological or expected features. In this case, the classifier captured the difference between two different datasets—one for patients, and one for controls. Similar to previously reported results, the study reported Grad-CAM to have low resolution and Guided Backpropagation to be noisy. For the quantitative evaluation, the study used ground truth segmentation information. While no explanation method produced satisfactory explanations, VAE-based perturbation outperformed constant and blur perturbation in meaningful perturbation setups. Figure \ref{fig:meaning_usage} shows the explanations and the use cases for this explantation technique for a classifier designed for retinal OCT images.

\subsection{Counterfactual}
\begin{figure} [!htbp]
 \includegraphics[width=1.0\linewidth]{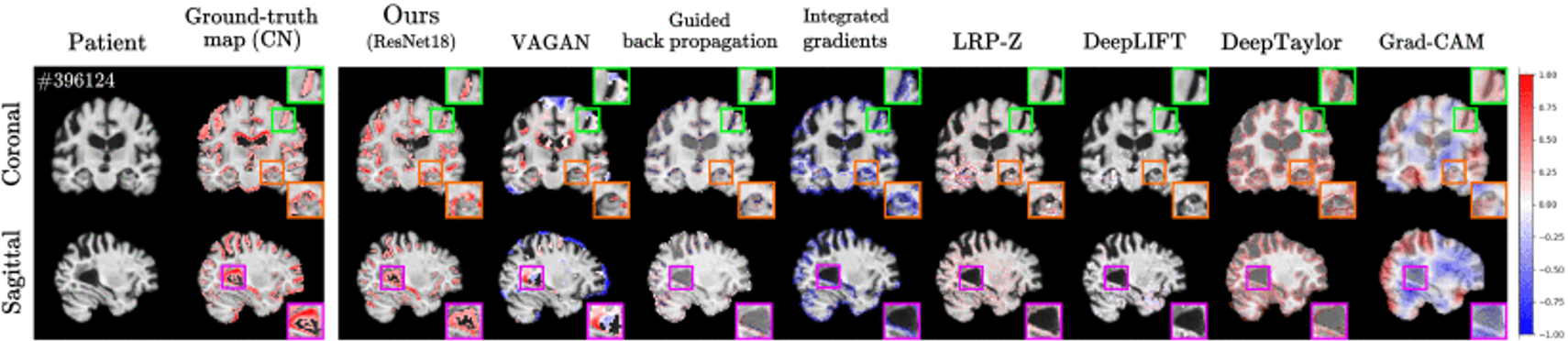}
 
 \centering
 \caption{Explanations generated by popular gradient-based approaches and the counterfactual maps as proposed in \cite{oh2022learn} for an ADNI dataset sample (subject ID shown on the top left corner). The purple, green, and orange boxes enclose ventricular, cortex, and hippocampus regions, respectively.  As the proposed counterfactual map captured all of the nuances (increased cortical thickness, reduced ventricular, and hypertrophy in the hippocampus) associated with the diagnosis, the method is superior in quality compared to visual attribution GAN (VAGAN) maps. While some gradient-based approaches exhibit some counterfactual reasoning ability, the traditional post hoc methods usually capture unnecessary regions not related to the disorder. (Image Courtesy: \cite{oh2022learn})}
 \label{fig:counterfactual_usage}
 \end{figure}

Oh et al.~\cite{oh2022learn} proposed an approach that combines model training, model counterfactual explanation, and model reinforcement as a unified learn-explain-reinforcement (LEAR) framework. After model training, a visual counterfactual explanation generates hypothetical abnormalities in the normal input image to be identified as a patient. The authors hypothesized that this counterfactual map generator when assisted by a diagnostic model can be a source of important generalized information about the disorder and the model can benefit from this. These counterfactual maps guide an attention-based module to refine the features useful for more efficient model training and representative for the disorder in consideration. For quantitative validation, ground-truth maps were created based on the clinical diagnosis of the longitudinal samples, and normalized correlation coefficients were calculated. The counterfactual maps for the targeted labels obtained using the proposed method were compared against the counterfactual reasoning ability of popular interpretability methods, such as LRP-Z, DeepLIFT, Deep Taylor Decomposition, Integrated Gradients, Guided Backpropagation, and Grad-CAM. While only guided backpropagation, integrated gradients, and deep Taylor show some counterfactual reasoning ability, all the post hoc methods were not designed to generate counterfactual explanations. Rather, they were intended to generate explanations for the natural predictions of the model. The most interesting thing about this study is that the counterfactual maps from the control class to the MCI class and from the MCI class to the AD class add up to the counterfactual map from the control class to the AD class, indicating the disorder-specific consistency of the method.  Moreover, the method is worth-investigating because the framework can be applied to any backbone diagnostic model. The LEAR framework also improves the post hoc interpretability of the model as demonstrated using CAM. However, the framework has only been evaluated on the ADNI dataset and it needs to be validated on independent datasets. Moreover, the cross-correlation measure for quantitative evaluation may be a weak measure of interpretability. Figure \ref{fig:counterfactual_usage} shows the explanations based on gradient-based approaches for why the subject was diagnosed as AD by the model and the corresponding counterfactual map based on the proposed generative approach. As shown and assessed with respect to the ground-truth control (CN) map, the counterfactual explanation detects and highlights the ventricle enlargement and cortical atrophies, aligning the result with previous reports.

\subsection{Distillation Methods}

{\bf Local Interpretable Model-agnostic Explanations (LIME)}\\
Magesh et al.~\cite{magesh2020explainable} leveraged the VGG16 network pretrained on \emph{ImageNet} dataset to classify Parkinson's patients from healthy controls. The study also used LIME to explain individual predictions. The underlying reason for choosing this explanation method is unclear. Furthermore, while quantitative validation is an essential measure of the predictability of the heatmaps, the study did not conduct any experiments to validate the generated explanations objectively.

Saboo et al.~\cite{saboo2022deep} developed a deep learning model for cognition prediction over five years after the baseline using clinical and imaging features and leveraged model explainability to identify certain brain structures responsible for cognitive vulnerability or cognitive resilience. Specifically, it identifies those predictive brain structures (medial temporal lobe, fornix, and corpus callosum) that best explain the heterogeneity between cognitively vulnerable and resilient subjects.  To this end, the study used LIME to compute the contributions of each imaging and clinical feature toward the predicted future cognitive score. While LIME generated contributions of features matched with some prior reports, earlier literature blamed LIME for producing unstable contributions over multiple runs. Moreover, like most of the ongoing interpretability practices, this study also did not justify the rationale of using LIME.

Gaur et al.~\cite{gaur2022explanation} proposed an explanation-driven dual-input CNN-based solution to predict the status of brain tumors using brain MRI scans.  The authors adopted both SHAP and LIME to generate explanations because SHAP captures the consistency and accuracy in the explanations and LIME  captures the local behavior of the model around the test example. The study reported a predictive training accuracy of 94.64\% as the performance of the model, which is not the usual practice of reporting model performance. Furthermore, while the study used two post hoc interpretability methods (LIME and SHAP), validation of the generated explanations was not provided to support the study. The explainability was not used as part of the model training, and hence we argue that the model is inappropriately referred to as "explanation-driven." Rather, the concept of XAI was used to analyze the model predictions in a post hoc manner, and not to enhance the model's performance. 

\bigskip
\noindent
{\bf SHapley Additive exPlanations (SHAP)}\\
Ball et al.~\cite{ball2021individual} used three different machine learning approaches to estimate brain age based on cortical development as the variations in brain age and chronological age have been linked to many psychiatric disorders. The authors also used kernel SHAP to identify which features explain errors (brain age delta) in brain age prediction. These explanations are consistent across the models and previously reported brain development regional patterns. However, no generic spatial association among individual feature estimates is found for the brain age prediction error. That is, given the similar demographics and prediction error, feature importance estimates between subjects may vary widely. Moreover, brain age delta estimates for any of the models did not associate noticeably with cognitive performance. Given the limitations of SHAP in estimating feature importance and the lack of any objective validation for these explanations, this study needs further investigation.  

\begin{figure} [!htbp]
 \includegraphics[width=1.0\linewidth]{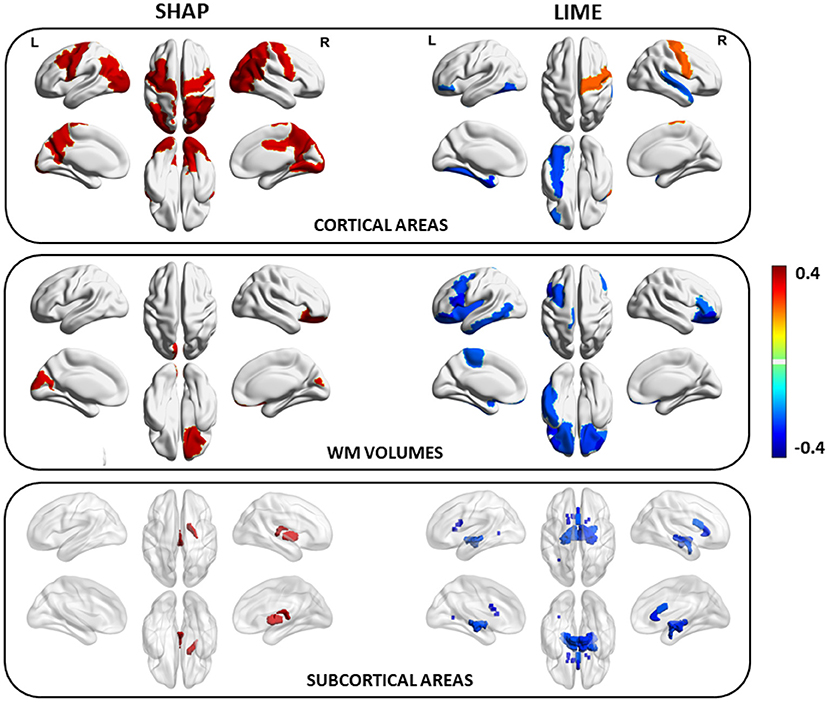}
 
 \centering
 \caption{ Brain regions with the most significant correlations between the XAI scores of the ROI-specific morphological features and the chronological age. For the SHAP method, the study reports that the average thickness, folding, and curvature index statistical attributes calculated based on the precentral gyrus and inferior and lateral occipital cortex were highly correlated with brain age.  Also, the SHAP scores of the cortical thickness features of both hemispheres showed a significant correlation with age. These findings are highly aligned with the previous reports. On the other hand, for LIME, the study found a strong correlation between brain age and the features related to WM volumes of the opercular and triangular parts of the inferior frontal gyrus and inferior temporal gyrus. Overall, the LIME explanations were not consistent and had a very low overlap with SHAP scores. (Image Courtesy: \cite{lombardi2021explainable})}
 \label{fig:lime_shap_usage}
 \end{figure}

Lombardi et al.~\cite{lombardi2021explainable} segmented the T1-weighted brain into 68 cortical regions and 40 sub-cortical regions using atlases to extract morphological features from the scans. The authors developed a fully connected network and analyzed the feature attribution performance of two model-agnostic interpretability methods: LIME and SHAP. To assess the predictive performance and intra-consistency of the interpretability methods, the model was trained multiple times using different subsets of the training fold. SHAP, compared to LIME, provided greater intra-consistency of feature attribution scores implying that the method is comparatively robust to the training set variations. Inter-similarity analysis across the subjects reveals that SHAP values can be better partitioned into different age ranges. Furthermore, compared to LIME, feature attribution scores from SHAP were highly correlated with brain age and the features were more aligned with the previous reports of morphological association with brain age. However, the obvious limitation of this study was that the model was trained  based on a smaller cohort of samples with a limited age range. Furthermore, a quantitative validation approach is required to objectively compare the performance of XAI methods.  Figure \ref{fig:lime_shap_usage} shows how LIME and SHAP were used to identify brain regions associated with the chronological age of the subjects.

\subsection{Intrinsic Methods}
Very few neuroimaging studies so far have considered interpretability as part of the algorithmic aspect of the model from its inception. Such models in the literature are called glass-box or transparent-box models. 
Biffi et al.~\cite{biffi2020explainable} proposed a deep generative model for transparent visualization of the classification space. Some other neuroimaging studies~\cite{eslami2021explainable, mahmood2022deep} considered interpretable models based on their design transparency. We discuss some studies that utilized the concept of intrinsic interpretability as part of the model design or training. 

\bigskip
\noindent
{\bf Attention Mechanism} \\
Lian et al.~\cite{lian2020attention} proposed an attention-guided unified framework that uses a convolutional neural network to work as the task-specific guidance and a multi-branch hybrid network to perform disease diagnosis. The main motivation behind this work is to extract multilevel information at interpersonal, individual, local, and global scales. The first fully convolutional network offers guidance via disease attention maps and thus assists in generating global and individual features. In the second stage, the unified framework leverages disease attention maps (DAMs) calculated using class activation maps. In the second stage, the framework passes the DAMs over different branches to capture patch-level and global-level information for actual classification. The AD classification task was designed from the scratch, whereas the MCI progression prediction model was built based upon the learned parameters of the AD classification model. Compared to the existing competing deep learning approaches, this framework offers little performance improvement. The proposed framework identified different parts of the hippocampus, frontal lobe, fusiform gyrus, amygdala, and ventricle in association with AD or MCI. The discriminative power of these regions has been supported by earlier studies conducted for Alzheimer’s disease. One of the main limitations of this work is that DAMs cannot be considered as explanations for the final predictive model. Because, a smaller part of the predictive model was pre-trained earlier to generate the DAMs and subsequently guide the predictive model for the intended classification task. Moreover, DAMs as generated using class activation maps are high in semantics but low in resolution, which may not reflect the fine details of the discriminative regions in the input space. 

\begin{figure} [!htbp]
 \includegraphics[width=1.0\linewidth]{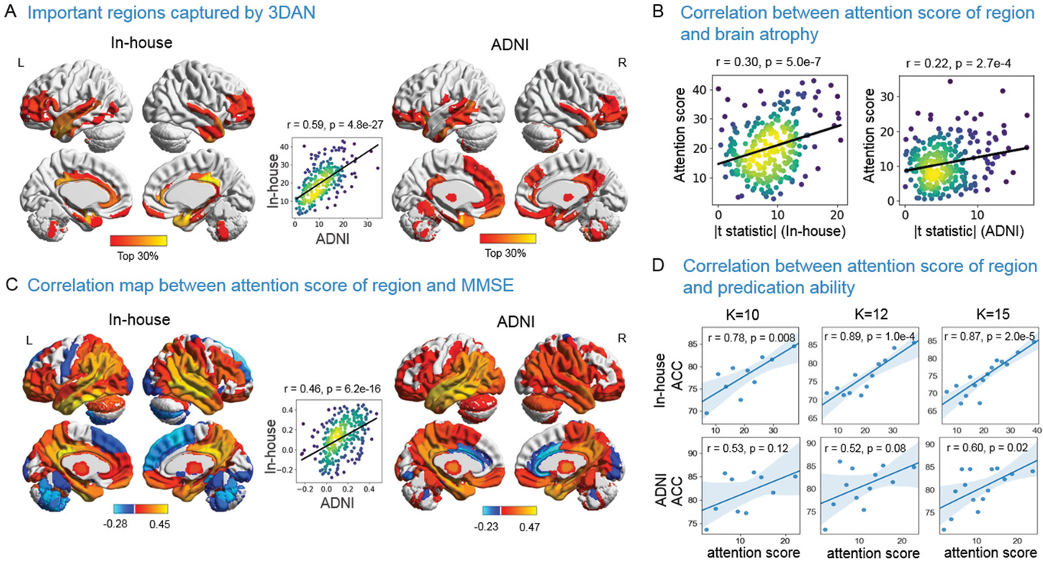}
 
 \centering
 \caption{{\bf Panel A:} Attention maps highlight the temporal lobe, hippocampus, parahippocampal gyrus, cingulate gyrus, thalamus, precuneus, insula, amygdala, fusiform gyrus, and medial frontal cortex as discriminative regions of AD. Noticeably, the attention patterns for both datasets were significantly correlated. {\bf Panel B:} Correlation analysis between T-map and the attention scores of the regions. High correlation values between attention scores and the group difference map for both datasets demonstrate the efficacy of the model's feature extraction. {\bf Panel C:} Correlation analyses between the attention scores and the Mini-Mental State Exam (MMSE) scores. Attention scores of 81\% and 77\% brain regions correlated with MMSE significantly for In-house and ADNI datasets, respectively. {\bf Panel D: } Correlation between mean attention score of top $K$ regions and the classification performance. As can be observed, the attention scores of top $K$ regions highly correlated with the classification performance for both datasets. (Image Courtesy: \cite{jin2020generalizable})}
 \label{fig:att_map_usage}
 \end{figure}

Jin et al.~\cite{jin2020generalizable} proposed a deep learning model combining ResNet and a 3D attention network (3DAN) to integrate the identification of discriminative regions and the classification of AD into a unified framework. The proposed model can diagnose Alzheimer's and simultaneously capture imaging biomarkers responsible for the predictions. As reported, the 3DAN provides a strong association between the model output and the clinical features of AD and MCI. The study claimed the biomarkers (attention maps) to be generalizable, reproducible, and neurobiologically plausible. The study validated these claims by demonstrating strong correlations of the attention maps between datasets, between the mean attention score and the T-map,  between the attention score for the regions and the MMSE scores based on the Brainnetome Atlas, and between classiﬁcation accuracy and the mean attention score of $K$ groups of regions. The attention mechanism was able to produce generalizable and reproducible results as the findings correlated across the datasets. Figure \ref{fig:att_map_usage} shows various aspects of the analyses for the effectiveness of attention maps in capturing significant brain regions associated with the progression of AD. 

\bigskip
\noindent
{\bf Joint Training}\\
Zhu et al.~\cite{zhu2022interpretable} combined interpretable feature learning and dynamic graph learning modules into the Graph Convolutional Network module. These three modules are jointly optimized to provide improved diagnosis and interpretability simultaneously via learning only the essential features. This study used feature summarization and used gray matter volumes from 90 ROIs as features. 
Interpretability of the diagnosis was conducted based on the values of the learned weights of each region. The predictive performance of the top regions was higher compared to other feature selection methods. Moreover, the proposed method consistently identified the middle temporal gyrus right, hippocampal formation left, precuneus left, and uncus left associated with AD, which aligned with the previous reports. While this validation seems plausible, it is unclear if the underlying classification model for this validation across the feature selection methods was the same or different. Furthermore, uninterpretable parameter sensitivity is another concern, which caused a significant drop in predictive performance.  

\begin{figure} [!htbp]
 \includegraphics[width=1.0\linewidth]{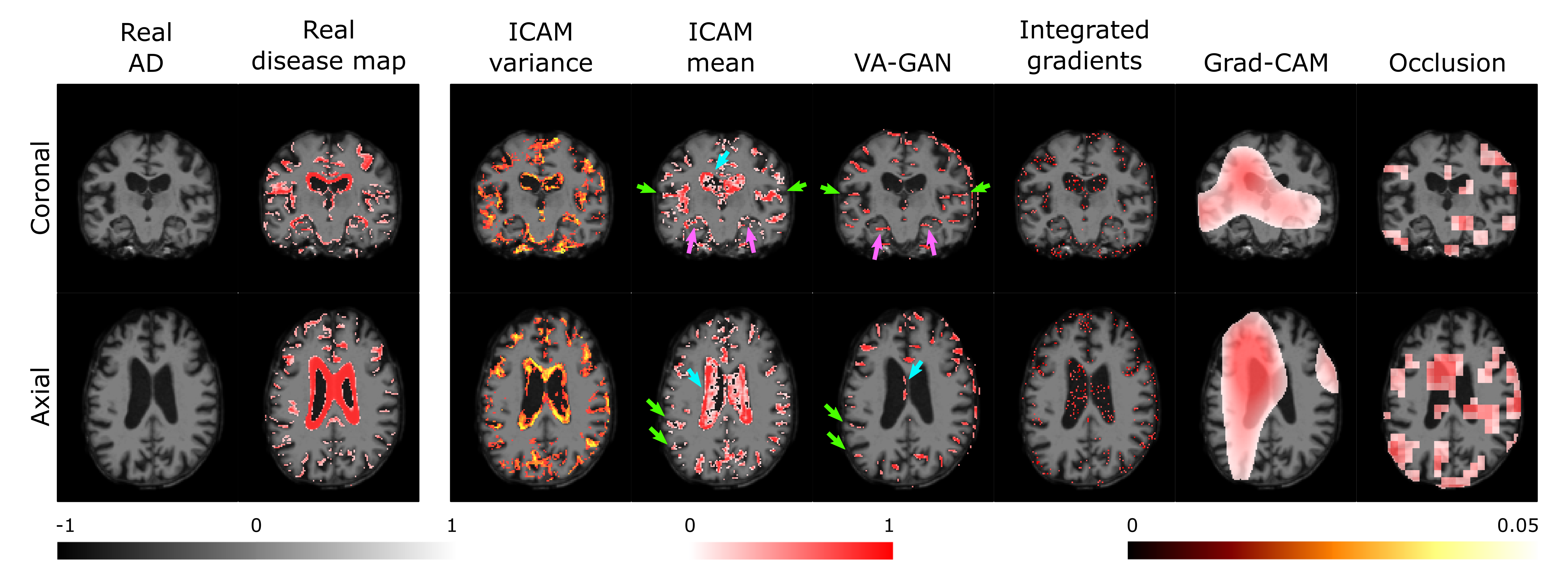}
 
 \centering
 \caption{ Feature attribution (FA) maps using ICAM, VA-GAN and other post-hoc methods for AD classification from ADNI dataset. ICAM better captures the phenotypic variation in brain structure compared to other baseline methods.  In particular, ICAM, when compared with the ground-truth disease map, detects the ventricles (blue arrows), cortex (green arrows), and hippocampus (pink arrows). (Image Courtesy: \cite{bass2020icam})}
 \label{fig:joint_training_usage}
 \end{figure}

Bass et al.~\cite{bass2020icam} proposed a VAE-GAN-based approach, called ICAM (Interpretable Classification via Disentangled Representations and Feature Attribution Mapping),  to learn a shared class-relevant attribute latent space, that is simultaneously suitable for classification, and feature attribution. It can also inform the difference within and between classes.  The authors argued that the post hoc methods are sub-optimal in finding all the discriminative regions of a class and hence not suitable for medical imaging. Instead, this work hypothesized that generative models are more useful to capture class-relevant features. As demonstrated in the results, ICAM's latent attribute space achieved greater discriminative power compared to other approaches. Furthermore, the attribution maps generated using this approach have increased correlations with the ground truth disease maps compared to other popular post hoc methods. Figure \ref{fig:joint_training_usage} shows comparative feature maps generated using ICAM, VA-GAN, and other post-hoc approaches.

\begin{figure} [!htbp]
\centering
 \includegraphics[width=0.8\linewidth]{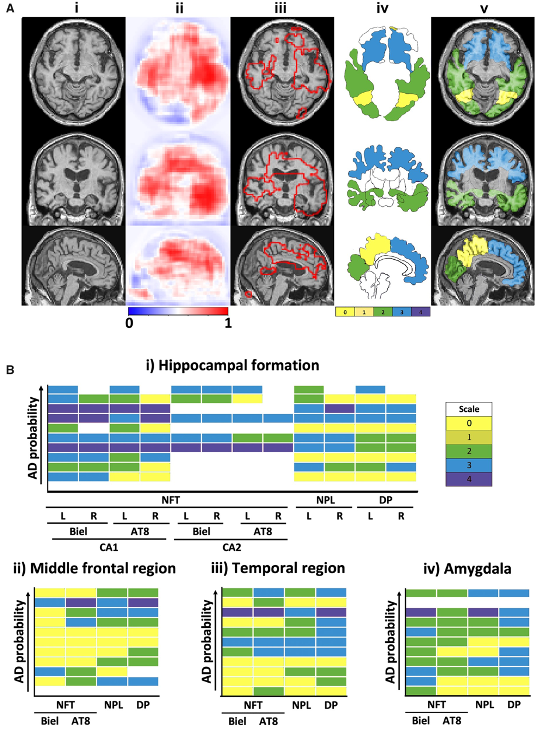}
 \caption{{\bf A:} Overlap of the model's predictions with the neuropathology for a single subject (AD). The column {\bf (i)} shows the MRI scans at different visual planes {\bf (ii)} depicts the predicted disease probability maps {\bf (iii)} A thresholded probability maps overlapped with the MRI scans, {\bf (iv)} shows the segmented brain masks with color-coding indicating different pathology levels, {\bf (v)} shows an overlay of MRI scan, thresholded probability maps, and the color-coded pathology levels.  {\bf B:} For qualitative assessment, the model examined the density of neurofibrillary tangles (NFT), diffuse senile (DP), and neuritic (NPL) or compacted senile plaques in each brain region. As shown, the average probabilities of the model predicted brain regions were consistently correlated with a high grade of amyloid-$\beta$ and tau in different brain regions. Biel = Bielschowsky stain; L = left; R = right. (Image Courtesy: \cite{qiu2020development})}
 \label{fig:model_transp_usage1}
 \end{figure}

\bigskip
\noindent
{\bf Model Transparency}\\
Qiu et al.~\cite{qiu2020development} utilized multimodal inputs such as MRI and other clinical features (age, gender, and Mini-mental State Examination score) to propose an interpretable deep learning framework for Alzheimer's disease. The proposed framework improves predictive performance for disease diagnosis and identifies disease-specific neuroimaging signatures. As part of the architecture, MRI sub-volumes are passed to a fully convolutional neural network, and patient-specific probability maps of the brain are generated. As estimated by the probability maps, the high-risk voxels are passed to a fully connected network for classification. Despite site-specific distinctions among the datasets, the study was able to demonstrate performance consistency across the datasets when training was conducted based on a single cohort. The study further provided neuropathological and neurologist-level validation. For the neuropathological validation, the high-probability brain regions were closely associated with the locations and frequencies of amyloid-$\beta$ and tau pathologies. Figure  \ref{fig:model_transp_usage1} demonstrates how the extracted disease probability maps correspond to the post-mortem findings of neuropathology examinations. For the neurologist-level validation, eleven neurologists conducted diagnoses based on the same multimodal inputs, and the average performance was compared to the model's performance. The model is transparent in the sense that it is able to directly produce disease probability maps from the scans, which is indicative of the disease-specific brain regions. While the preliminary results are promising, the model only considered the subpopulation of binary scenarios, subjects with AD and normal controls, and does not consider progressive stages. Hence, the model is still not directly applicable to the clinical decision-making process. 

Ravi et al.~\cite{ravi2022degenerative} proposed a 4D-Degenerative Adversarial NeuroImage Net (4D-DANINet) for generating realistic 3D brain images over time. 4D-DANINet is a modular framework based on adversarial training and a set of spatiotemporal and biological constraints. It can generate subject-specific longitudinal scans that reflect disease stage and age. The key motivation behind this work is to provide sufficient realistic samples for model validation and efficient model training. The main components of the DANINet are as follows:

\begin{itemize}
\item Conditional deep encoder: This module is a combination of an encoder that embeds each slice to a latent space $z$ and a generator that generates samples from this latent space conditioned on diagnosis and age. This module is trained based on reconstruction loss that minimizes the difference between the actual slice and the projected slice in time. 

\item Discriminator networks: This module uses two discriminators $\mathcal{D}^b$ and $\mathcal{D}^z$.  The $\mathcal{D}^b$ discriminates between real and simulated brain images and the generator (G) generates realistic synthetic images to fool the $\mathcal{D}^b$. The encoder (E) produces embeddings from a uniform distribution to ensure smooth temporal progression. To do this,  the discriminator $\mathcal{D}^z$ is adversarially trained with the encoder (E). 

\item Biological constraints: The 4D-DANNet imposes two separate voxel-level and region-level losses $L^\text{vox}$ and $L^\text{reg}$ to capture smooth intensity changes reflecting disease progression over time.

\item Profile weight functions dynamically determine appropriate weights for the losses as required for efficient training. 
\end{itemize}

\begin{figure} [!htbp]
\centering
 \includegraphics[width=1.0\linewidth]{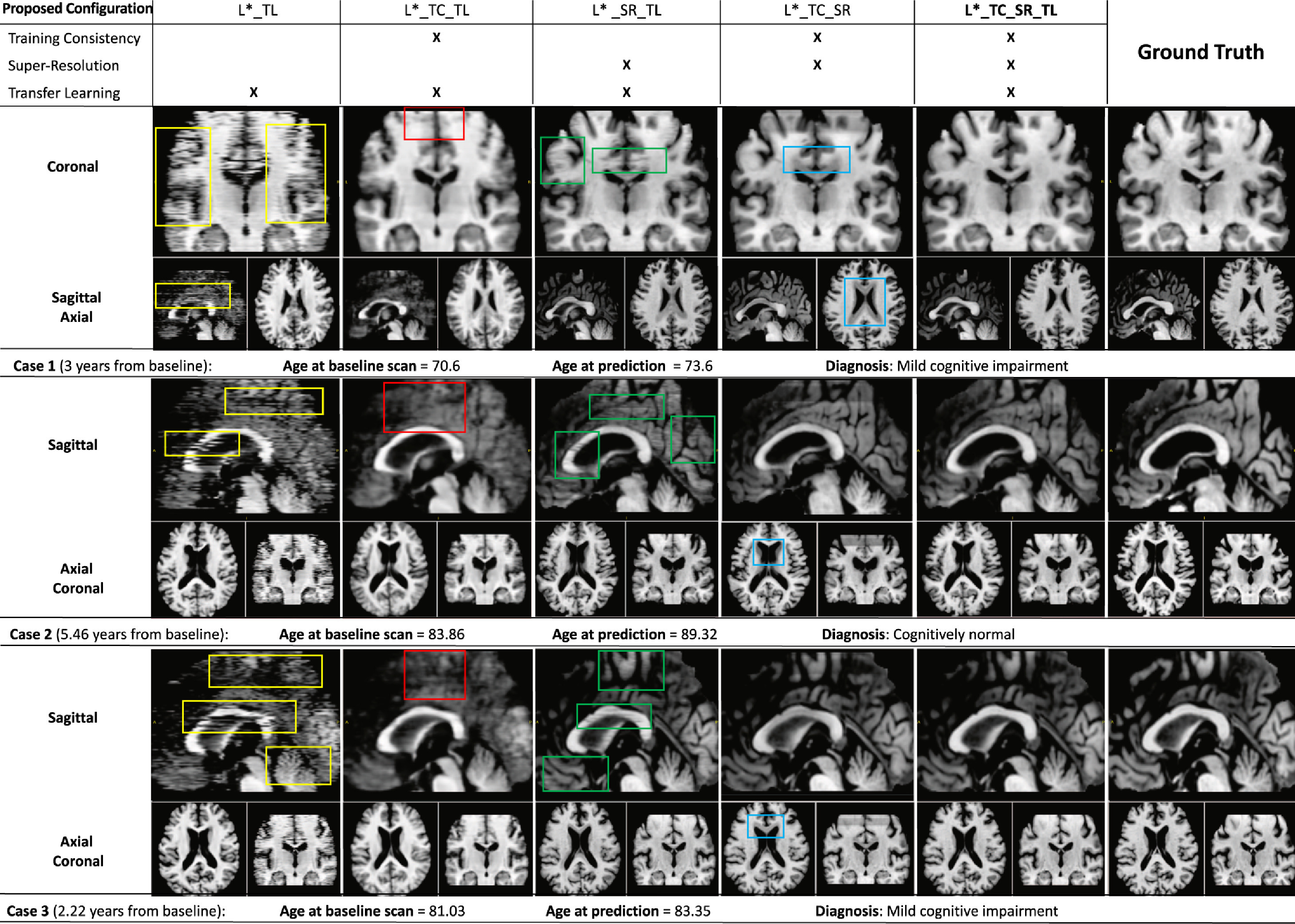}
 \caption{Ablation study demonstrating the performance of the proposed full configuration \texttt{L$^{*}$\_TC\_SR\_TL} in producing superior synthetic MRI. The \texttt{L\_TL} configuration lacks 3D consistency constraints and when training does not converge, it results in artifacts appearing in sagittal and coronal axes (yellow boxes). When configurations include 3D training consistency strategy TC, such artifacts disappear from the resulting images (Refer to \texttt{L\_TC\_TL} and \texttt{L\_TC\_SR\_TL} configurations). Super-resolution does play an important role and if not used, anatomical details are often not visible (red boxes). If SR only model is used without TC, unrealistic images may appear for artifact super-resolution (green boxes). If transfer learning TL procedure (fine-tuning) is omitted during test time, lack of individualization may cause inaccurate morphology (configuration \texttt{L\_TC\_SR}, blue boxes). (Image Courtesy: \cite{ravi2022degenerative})}
 \label{fig:model_transp_usage2}
 \end{figure}

The ablation study demonstrated the significance of training consistency (TC), super-resolution (SR),  and transfer learning (TR) blocks as adopted in the framework to produce realistic synthetic MRI images. Figure \ref{fig:model_transp_usage2} shows how different components of the proposed framework have played their respective roles in producing realistic synthetic MRI images. However, the model may produce less effective brain images due to poorly representative training sets, and cohort differences. 

\subsection{Feature Map Visualization}

Biffi et al.~\cite{biffi2020explainable} proposed a hierarchical deep generative model called ladder variational autoencoder (LVAE). LVAE learns a hierarchy of conditional latent variables to represent the population of anatomical segmentations. The latent space representation in the highest level of the hierarchy can efficiently discriminate clinical conditions. The proposed model performed two classification tasks: 1) Hypertrophic cardiomyopathy (HCM) versus healthy 3D left ventricular (LV) segmentations and 2) AD versus healthy control 3D hippocampal segmentations. The model was predictive of clinical conditions and offered suitable visualization and quantification of the anatomical shape changes associated with those clinical conditions. This study used sampling in the highest latent space to visualize the corresponding regions in the brain space. The authors further claimed that the shape changes, as evident in the visualization, agreed with the clinical literature. 

\begin{figure} [!htbp]
 \includegraphics[width=1.0\linewidth]{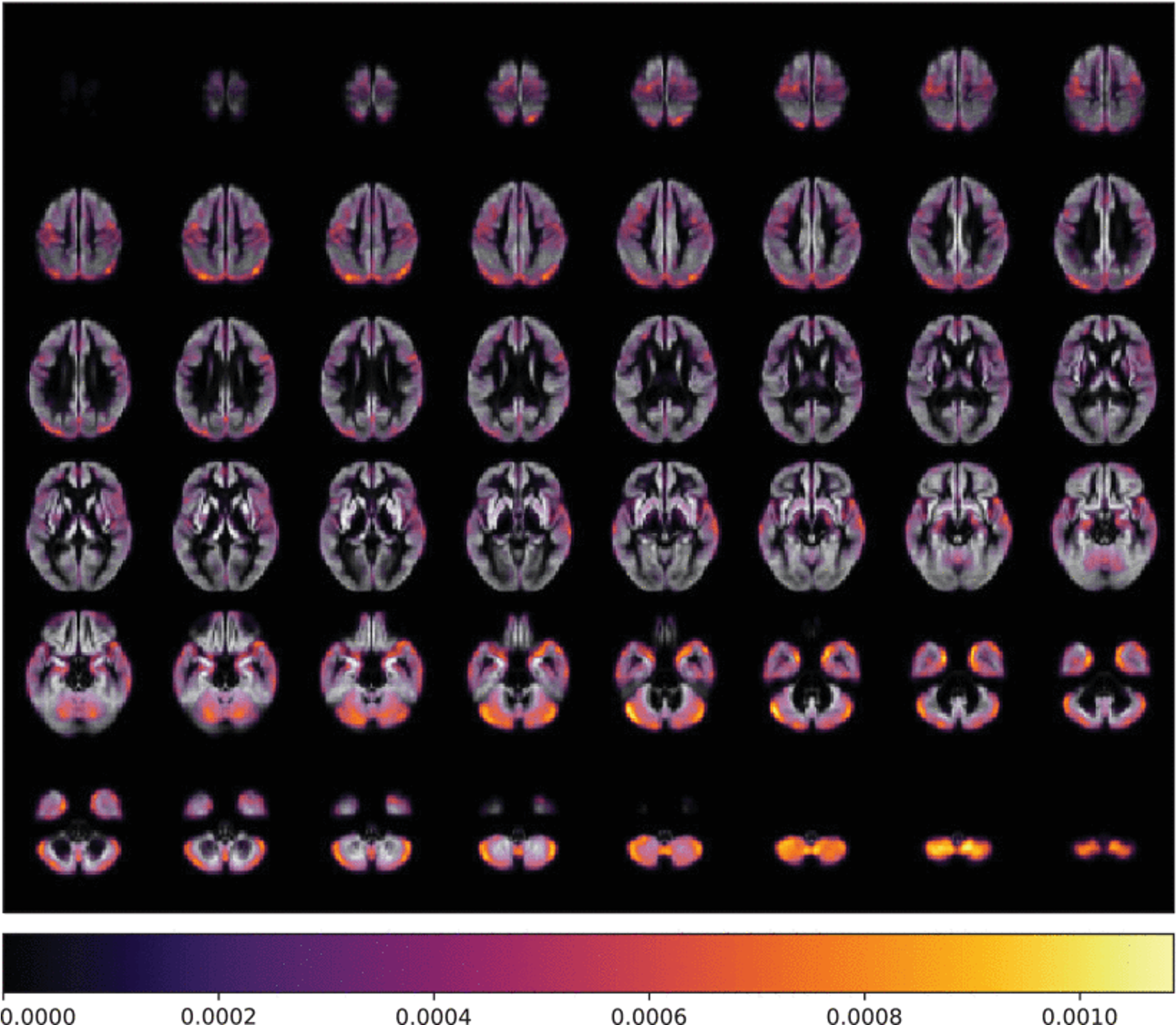}
 
 \centering
 \caption{Area of Influence of the $14^{\text{th}}$ neuron. The area of influence highlighted parts of the temporal and parietal lobes, as well as a shrinkage of the cerebellum. These brain regions are traditionally associated with the progression of AD. Thus, the visualization confirms the underlying association between anatomical changes of neurodegeneration and the magnitude of the neuron in the $z$ space.  (Image Courtesy: \cite{martinez2019studying})}
 \label{fig:feature_map_usage}
 \end{figure}

Martinez-Murcia et al.~\cite{martinez2019studying} used a deep CNN  autoencoder for an exploratory data analysis of AD. The autoencoder demonstrates links between cognitive symptoms and the underlying neurodegenerative process. The autoencoder model uses a data-driven approach to extract imaging characteristics into low-dimensional manifolds. The study further used regression analysis to show that the neurons in the manifold space correlate well with the clinical and neuropsychological test outcomes and diagnoses. Subsequently, the authors used a novel visualization approach using a linear decomposition model to show the brain regions highly influenced by each manifold coordinate, which provides additional information about the association between structural degeneration and the cognitive decline of dementia. Figure \ref{fig:feature_map_usage} shows the brain regions that affect $14^{\text{th}}$ neuron (the most correlated with disease progression) in the $z$-space using a linear decomposition method. The regression model for different test clinical variables, such as ADAS-13, ADAS-11, Age, TAU, etc., was trained with GM maps. 

As brain dynamics change far before the changes happen in the anatomical structure, functional neuroimaging is more powerful to understand brain disorders~\cite{parmar2020spatiotemporal}. Parmar et al., 2020~\cite{parmar2020spatiotemporal} proposed a modified 3D CNN that directly works on the 4D resting-state fMRI data for a multiclass classification task to diagnose different stages of Alzheimer's disease. The network achieved very high accuracy (93\%) with only 30 subjects per class. However, the data augmentation using temporal patches may prevent the spatio-temporal characteristics of the underlying dynamics. The study also showed the temporal features extracted from the first two convolutional layers as network activation maps. As the temporal features move from lower to higher layers, the authors reported that discriminative regions of interest gradually take a definitive structure. The obvious limitation of this study is the lack of interpretability. Moreover, the model was trained and evaluated on only one dataset.


\section{The Usage Trend of Interpretability Methods}
\label{usage_trend}

While the neuroimaging community has used a larger collection of interpretability methods, only a few are popular and considered important for knowledge discovery or potential clinical deployment. Several interpretability methods have often been used as experimental baselines, not for their beneficial effects in this domain. In this section, we conducted an in-depth analysis of the usage of all popular interpretability methods in neuroimaging studies. Indeed, we investigated the usage of these methods in more than 300 neuroimaging papers and observed their usage trend as shown in Table~\ref{usage_trend_table}. As we found in our exploratory analysis, studies have used the methods in the following order of frequency: \begin{inparaenum}[1)]
\item  CAM/Grad-CAM/Grad-CAM++/Guided Grad-CAM~\cite{zhou2016learning, selvaraju2017grad,chattopadhay2018grad}
\item SHapley Additive exPlanations (SHAP)~\cite{lundberg2017unified}
  \item Integrated Gradients~\cite{sundararajan2017axiomatic}
  \item Layer-wise Relevance Propagation~\cite{bach2015pixel}
  \item Occlusion Sensitivity~\cite{zeiler2014visualizing} 
  \item Guided Backpropagation~\cite{springenberg2014striving}
  \item Local Interpretable Model-Agnostic Explanations (LIME)~\cite{ribeiro2016should}
  \item Gradients~\cite{baehrens2010explain, simonyan2013deep}
  \item DeepLIFT~\cite{shrikumar2017learning}
  and 
  \item Smoothgrad~\cite{smilkov2017smoothgrad}
\end{inparaenum} This usage trend also reveals that preference for "gradients" and "guided backpropagation" methods are receiving less attention because of their limitations~\cite{smilkov2017smoothgrad, adebayo2018sanity}, while the steeper rise in the use of integrated gradients and SHAP are potentially due to their strong theoretical foundations. 

\setlength{\tabcolsep}{6pt}

\begin{longtable}{p{9.5cm}R}
 
\caption{The usage trend of popular post hoc interpretability methods in neuroimaging studies.} \label{table:usage_trend} \\ \hline

     \toprule 
    \hline
      \makecell[c]{\bf Interpretabily Method \& Citing Publications}  & {\bf Usage Trend} \endhead \midrule  
      
   {\bf Occlusion Sensitivity~\cite{zeiler2014visualizing}:} \cite{kwak2022differential, kwak2022identifying, polson2022deep, zheng2022transformer, lu2022practical, lee2022deep, bi2022prediction, hahn2022uncertainty, ning2022characterizing, vyas2022deep, wood2022accurate, polson2022identifying, bordin2022explainable, etminani20223d, gao2022deep, lee2022synthesizing, qu2021brain, see2021unraveling, abrol2021deep, zou2021deep, murugan2021demnet, leming2021deep, etminani2021peeking, dyrba2021improving, wood2021automated, turkan2021convolutional, bass2021icam, kwak2021subtyping, wang2021graph, bae2021transfer, thomas2020classifying, mostafa2020parkinson, kwak2020atrophy, oh2020identifying, leming2020stochastic, bass2020icam, abrol2020deep, west2019assessing, bae2019transfer, oh2019classification, islam2019understanding, kam2019deep, liu2019deep, eitel2019testing, esmaeilzadeh2018end, vakli2018transfer, liu2018classification, chambon2018deep, rieke2018visualizing, yang2018visual, gutierrez2018deep} & \raisebox{\dimexpr-\height+\ht\strutbox+9ex}{\includegraphics[width=6cm]{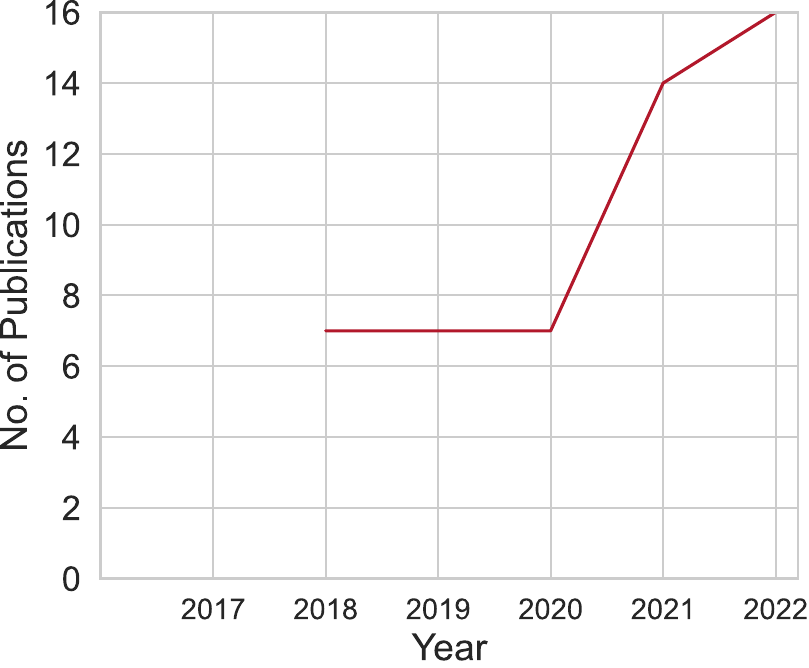}} \\  \hline
   
  {\bf Gradients~\cite{baehrens2010explain, simonyan2013deep}:} \cite{lin2022sspnet, aslan2022deep, zeineldin2022explainability, thomas2022comparing, cui2022towards, sokolova2021investigating, fernandez2021convolutional, hu2021gat, wilms2021towards, liu2021going, pianpanit2021parkinson, mcclure2020improving, lopatina2020investigation, sanchez2020deep, oh2019classification, islam2019understanding, rieke2018visualizing, garg2017using} & \raisebox{\dimexpr-\height+\ht\strutbox+9ex}{\includegraphics[width=6cm]{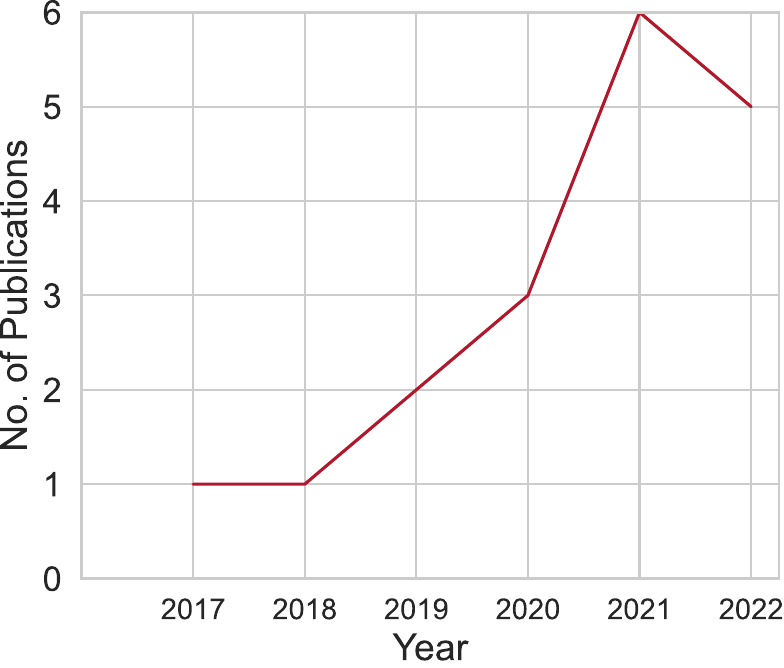}} \\[1cm]  \hline

    {\bf CAM/Grad-CAM/Grad-CAM++/Guided Grad-CAM~\cite{zhou2016learning, selvaraju2017grad,chattopadhay2018grad}:} \cite{fujiwara2022deep, ushizima2022deep, liu2022hierarchical, zhang2022single, huang2022meg, prats2022artificial, inglese2022mri, liu2022cascaded, aslan2022deep, zhang2022detecting, tousignant2022prediction, li20223, ding2022predicting, tasci2022deep, odusami2022intelligent, ban2022diagnosis, yu2022novel, ko2022deep, lu2022practical, jimeno2022artifactid, zihni2022moving, delbarre2022application, dasanayaka2022interpretable, shaban2022resting, altuve2022intracerebral, lu2022two, gao2022classification, yu2022morphological, huang2022transformer, guo2022diagnosing, kumar2022neurophysiologically, jonas2022diagnostic, ahmed2022dad, wang2022deep, fu2022grad, thomas2022comparing, zhou2022interpretable, oh2022learn, chen2022interpretable, giudice2022visual, kurmi2022ensemble, williamson2022improving, polson2022deep, termine2022reproducible, maqsood2022multi, li2022semi, zia2022vant, qin20223d, zeineldin2022explainability, oktavian2022classification, luckett2022deep, watanabe2022multi, polson2022identifying, oruganti2022outlier, garcia2022towards, lin2022sspnet, sokolova2021investigating, bruningk2021back, wang2021interpretable, wang2021clcu, ocasio2021deep, fernandez2021convolutional, ni2021detection, azcona2021analyzing, mishinov2021identification, kuijs2021interpretability, mathur2021deep, jeong2021deep, treacher2021megnet, raju2021deep, patel2021simulating, leming2021single, messina2021voxel, zhang2021intra, mohebbian2021classifying, liu2021going, zihni2021analysis, song2021effective, gao2021multisite, hepp2021uncertainty, ozsahin2021classification, kim2021cerebral, li2021comparison, yasaka2021parkinson, zhang2021explainable, hu2021interpretable, mei2021differentiation, shaji2021classification, pianpanit2021parkinson, turkan2021convolutional, bass2021icam, zhang2021explainable, zhang2021grad, fan2021u, pati2021gandlf, tomaz2021visual, menon2021multimodal, youn2021diagnosis, ullanat2021novel, etminani2021peeking, ko2021engine, wang2021deep, dandil2021detection, leming2021deep, hu2021accurate, wilms2021towards, yee2021construction, esmaeili2021explainable, liu2020pilot, qu2020baenet, kan2020interpretable, leming2020stochastic, li2020multi, pereira2020combination, nguyen2020attend, cote2020interpreting, kim2020understanding, ke2020exploring, akiyama2020deep, nishi2020deep, da2020visual, aslan2020automatic, alegro2020deep, vakli2020predicting, sanchez2020deep, pelka2020sociodemographic, bass2020icam, dyrba2020comparison, azcona2020interpretation, yee2020quantifying, yee2020structural, natekar2020demystifying, feng2020estimating, tang2020deep, chen2020early, wu2020novel, kan2020interpretation, wang2019grey, iizuka2019deep, alegro2019deep, iizuka2019p1, lala2019convolutional, gao2019dense, chen2019use, demyanchuk2019hydrocephalus, feng2019estimating, noguera2019using, wang2019o2, bermudez2019anatomical, hilbert2019data, lee2019explainable, petrov2018deep, pereira2018automatic, baumgartner2018visual, golkov2018q, yang2018visual, andreotti2018visualising, feng2018deep, pominova2018voxelwise, garg2017using, garg2017automatic}	& \raisebox{\dimexpr-\height+\ht\strutbox-0.5ex}{\includegraphics[width=6cm]{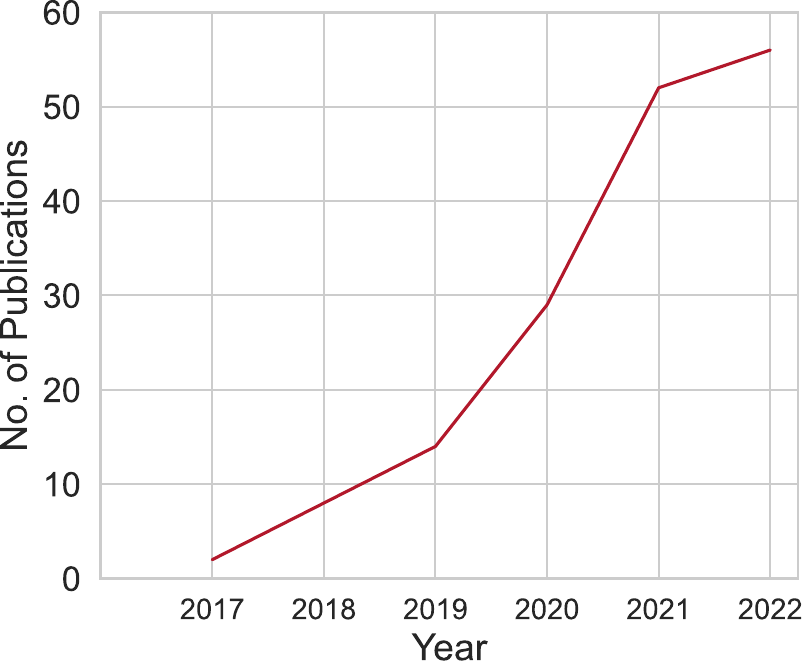}} \\[1cm] \hline
    
{\bf Integrated Gradients~\cite{sundararajan2017axiomatic}:} \cite{wang2022deep, yu2022morphological, rahman2022interpreting, supekar2022robust, wadhera2022computing, lewis2022fine, narazani2022pet, thomas2022comparing, spitzer2022interpretable, sarasua2022hippocampal, richie2022open, kim2022investigation, ferrante2022bayesnetcnn, richie2022analysis, lin2022deep, jiang2022pre, zeineldin2022explainability, prasad2022deep, shen2022contrastive, cui2022towards, supekar2022deep, choi2022subgroups, oh2022learn, polson2022identifying, de2021deep, de2021temporal, manjunatha2021extracting, bass2021icam, xin2021interpretation, lin2021classification, pan2021multi, bass2020icam, varatharajah2019joint, liu2019using, baumgartner2018visual}	& \raisebox{\dimexpr-\height+\ht\strutbox+5ex}{\includegraphics[width=6cm]{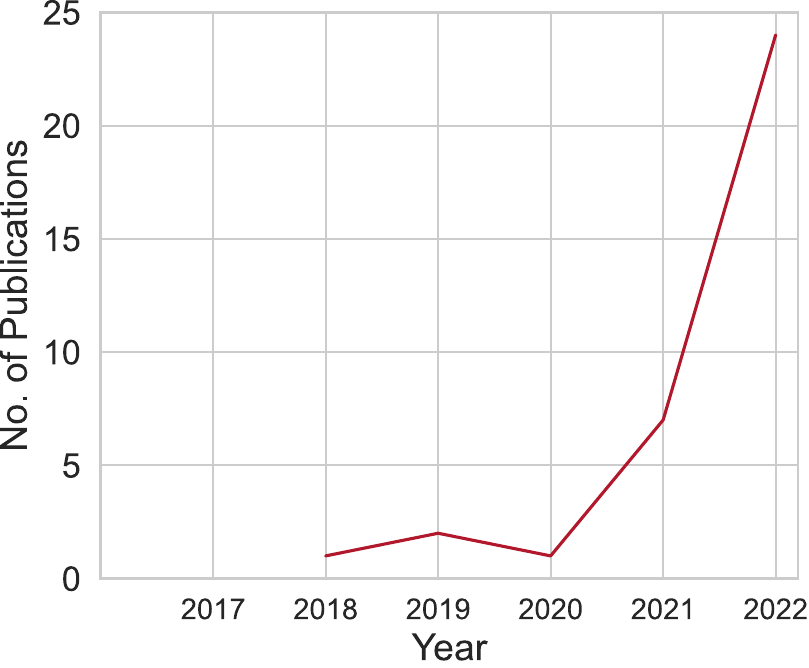}} \\[1cm] \hline

{\bf Smoothgrad~\cite{smilkov2017smoothgrad}:} \cite{zeineldin2022explainability, thomas2022comparing, rahman2022interpreting, wang2022high, wood2022deep, mayor2021evaluation, sokolova2021investigating, mcclure2020improving} & \raisebox{\dimexpr-\height+\ht\strutbox+9ex}{\includegraphics[width=6cm]{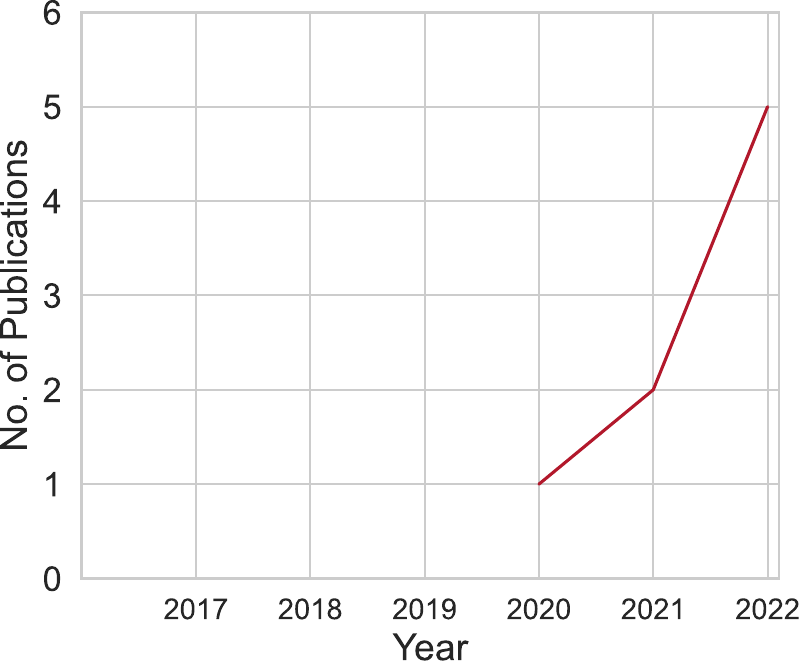}} \\[1cm] \hline

{\bf Guided Backpropagation (GBP)~\cite{springenberg2014striving}:} \cite{cui2022towards, wood2022deep, zeineldin2022explainability, polson2022deep, yu2022morphological, wood2022accurate, thomas2022comparing, polson2022identifying, tinauer2022interpretable, cho2022pre, oh2022learn, yee2021construction, pati2021gandlf, abrol2021deep, hu2021deep, bron2021cross, liu2021going, sokolova2021investigating, pianpanit2021parkinson, cruciani2021explainable, zhang2021functional, bass2021icam, shi2020fetal, kan2020interpretable, li2020multi, wang2020decoding, yee2020quantifying, folego2020alzheimer, yee2020structural, kan2020interpretation, dyrba2020comparison, bass2020icam, islam2019understanding, eitel2019testing, bohle2019layer, rieke2018visualizing, golkov2018q, pereira2018automatic, baumgartner2018visual} & \raisebox{\dimexpr-\height+\ht\strutbox+2ex}{\includegraphics[width=6cm]{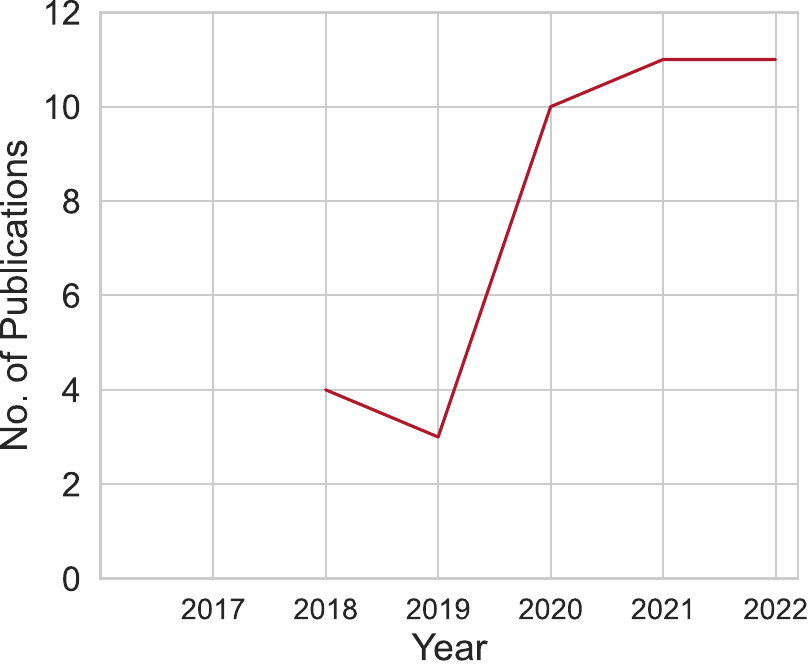}} \\[1cm] \hline

{\bf DeepLIFT~\cite{shrikumar2017learning}:} \cite{prasad2022deep, thomas2022comparing, cui2022towards, oh2022learn, gupta2022decoding, taylor2022brain, gupta2021obtaining, pianpanit2021parkinson, bennett2021universal, hu2021gat, mane2021fbcnet, lopatina2020investigation, gupta2019decoding} & \raisebox{\dimexpr-\height+\ht\strutbox+9ex}{\includegraphics[width=6cm]{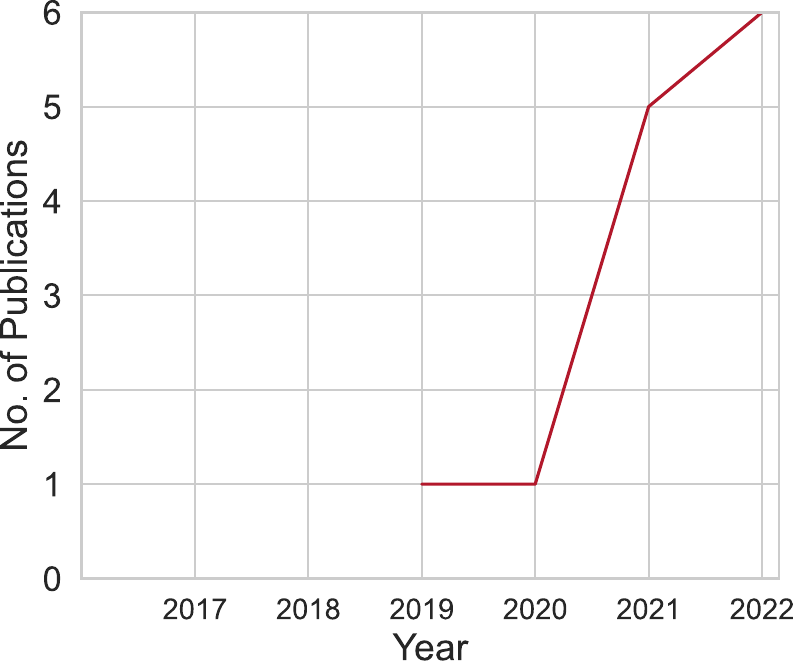}} \\[1cm] \hline

{\bf Layer-wise Relevance Propagation~\cite{bach2015pixel}:} \cite{tinauer2022interpretable, hofmann2022towards, pohl2022interpretability, korda2022convolutional, chen2022deep, torres2022facial, sudar2022alzheimer, kim2022interpretable, cui2022towards, nazari2022explainable, ellis2022approach, lu2022transfer, oh2022learn, ellis2022towards, klingenberg2022higher, hofmann2022towards, zhao2022attention, taylor2022brain, thomas2022comparing, wang2022deep, deatsch2022development, ellis2021gradient, dyrba2021improving, nazari2021data, bang2021spatio, mayor2021evaluation, ellis2021explainable, pena2021toward, raison2021explicability, bennett2021universal, cruciani2021explainable, cruciani2021interpretable, jeon2021mutual, eitel2021patch, dong2021explainable, korda2021identification, wang2020understanding, lopatina2020investigation, dyrba2020comparison, sagar2020learning, douglas2020similarity, masood2020interpretable, jo2020deep, dyrba2020validation, kohoutova2020toward, jo2020deep, bohle2019layer, eitel2019uncovering, eitel2019testing, pang2019clinical, thomas2019analyzing, dang2019novel, xie2019two, pena2019quantifying, islam2019understanding, gotsopoulos2018reproducibility, wang2018temporal, yan2017discriminating, sturm2016interpretable} & \raisebox{\dimexpr-\height+\ht\strutbox+3ex}{\includegraphics[width=6cm]{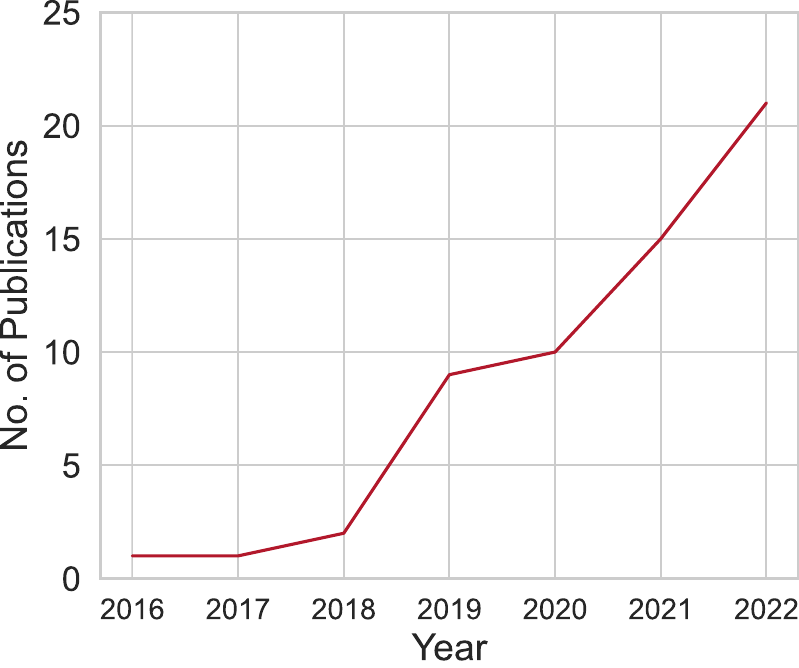}} \\[1cm] \hline

{\bf SHapley Additive exPlanations (SHAP)~\cite{lundberg2017unified}:} \cite{sanchez2022evaluation, lombardi2022robust, lombardi2022embedding, paul2022superior, eder2022interpretable, islam2022explainable, hu2022classification, dai2022alterations, brink2022age, gaur2022explanation, allwright2022machine, ran2022brain, shaji2022explainable, allwright2022machine, xu2022combined, kim2022predicting, bloch2022machine, bang2022interpretable, kannampallil2022cross, pang2022multimodal, ghosal2022biologically, gomez2022prediction, qiu2022multimodal, ben2022neural, xie2022morphometric, scheda2022explanations, ho2022predicting, treaba2021cortical, bang2021interpretable, bozek2021classification, kim2021diagnosis, huang2021multi, breitenbach2021automatic, bloch2021data, danso2021developing, lombardi2021explainable, bloch2021comparison, wickramaratne2021deep, turkan2021convolutional, salih2021new, pianpanit2021parkinson, el2021multilayer, sommer2021classification, ball2021individual, pan2021multi, chun2020visualizing, li2020efficient, sendi2020visualizing, suh2020development, ahmedt2020identification, saueressig2020exploring, al2020predicting, azevedo2019machine} & \raisebox{\dimexpr-\height+\ht\strutbox+6ex}{\includegraphics[width=6cm]{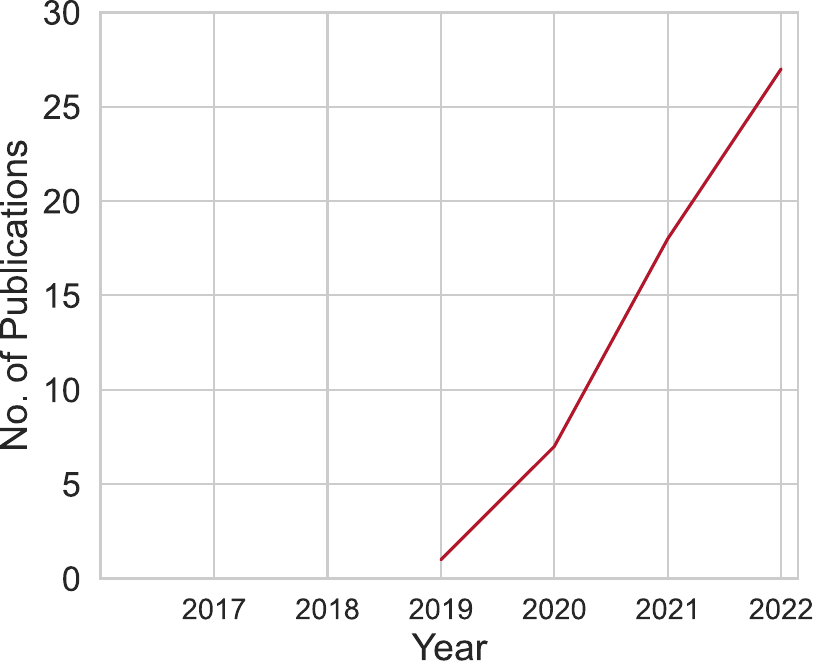}} \\[1cm]  \hline

 {\bf Local Interpretable Model-Agnostic Explanations (LIME)~\cite{ribeiro2016should}:} \cite{lombardi2022embedding, tasci2022deep, saboo2022deep, magboo2022explainable, gaur2022explanation, magboo2022important, sanchez2022evaluation, jahan2022explainable, lampe2022comparative, giudice2022visual, salih2021new, lombardi2021explainable, shad2021exploring, breitenbach2021automatic, sidulova2021towards, magesh2020explainable, hu2020interpretable, douglas2020similarity, ramos2019machine, santana2019using, pereira2018enhancing} & \raisebox{\dimexpr-\height+\ht\strutbox+5ex}{\includegraphics[width=6cm]{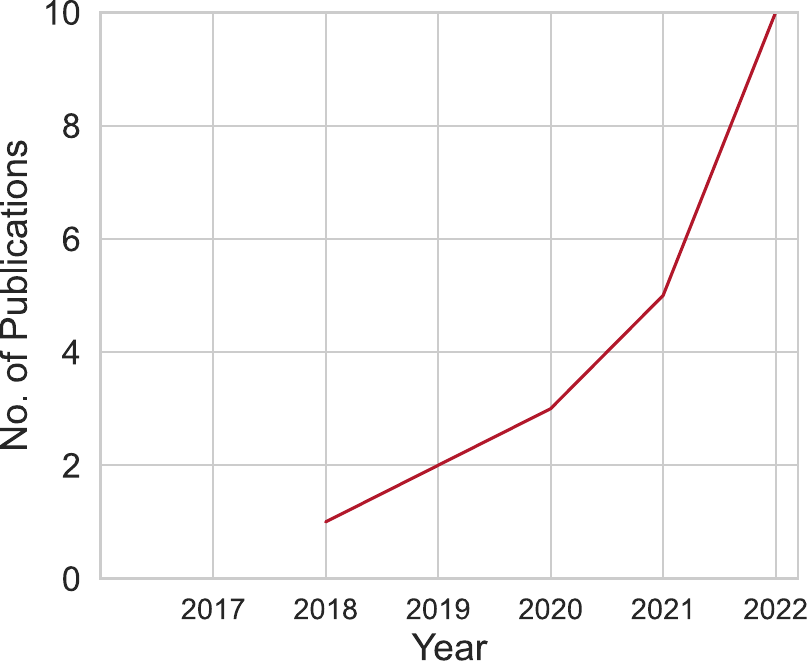}} \\[1cm]  \hline
     
     \bottomrule
     \label{usage_trend_table}
    \end{longtable}

\setlength{\tabcolsep}{6pt}

\section{Suggestions for Interpretable Models in Neuroimaging}
\label{suggestions}

In this section, we discuss the significant pitfalls of interpretability research in neuroimaging. One of the obvious concerns in interpretable deep learning models is that DL models are capable of learning in numerous ways for the same input-output relationship~\cite{hinton2018deep} and in most cases the hidden factors are not interpretable. Hence, the real challenge is to verify if the model learned from the true evidence or relied on some unintended spurious correlations~\cite{adebayo2022post,geirhos2020shortcut,lapuschkin2019unmasking} that could be entirely unknown to the humans. As such, explanations vary widely among architectures, model initializations, and interpretability methods. 

Model architecture is also a major consideration. As different neural networks may assign different regions as important for predictions, most of them would tell about different aspects of the disorder because of their very nature of different computations performed during training. Combining explanations from different models and further analysis of these explanations in association with medical experts may be useful in revealing undiscovered aspects of the disease. To this end, a unified framework~\cite{kohoutova2020toward} in interpretable neuroimaging research may be useful so that the findings across the studies can be directly compared to share advancement benchmarks. Based on our analysis and review, we recommend that we may focus on the following directions for useful DL-based understanding of the disorders: 1) Objective quantification of the explanation method's performance, 2) Investigation of the explanation sensitivity to interpretability parameters, 3)  Understanding if the generated explanations point to any causality or underlying mechanism of the disorder, 4) Investigating the reliability of the underlying model via model debugging 5) Combining various aspects revealed using multiple approaches, multiple model initializations and model ensembles. This is important because even when models use "true" evidence, explanations and their relative importance may be different. So combining them in a faithful manner can reveal useful unknown insights for the disorders. As Rieke et al.~\cite{rieke2018visualizing} pointed out that different visualization methods (gradients or non-gradient approaches) vary widely, so in line with other earlier studies, we suggest investigating multiple methods instead of blindly relying on one interpretability method. 

We also should be careful while choosing a particular interpretability method. For example, while many earlier studies used the \emph{occlusion sensitivity} method to generate explanations, Yang et al., 2018~\cite{yang2018visual} pointed out several limitations of the approach. For example, this approach uses semantically meaningless neighborhoods and an unspecified way of choosing the grid size. Moreover,  the method is computationally very intensive. As no backpropagation from the target score is involved during heatmap generation, this explanation is considered to be limited~\cite{oh2019classification}. Also, 3D-Grad-CAM can be useful if we need to track the attention of the convolution layers, but Grad-CAM or CAM is not useful for generating explanations in the input space and hence not suitable for data interpretation. While LRP has been used extensively, it has inherent limitations. LRP cannot maintain implementation invariance as it uses modified back-propagation rules. For future interpretability practices, we leave the following suggestions for the neuroimaging community: 

\begin{enumerate}[font=\bfseries]

\item {\bf Be aware of shortcut learning:} Deep learning models are very prone to fall into the trap of shortcut learning because it always looks for the easiest possible solution for the problem~\cite{geirhos2020shortcut}. The understanding of the influence of model architecture, training data, loss function, optimization parameters may reveal the nature of shortcuts the model may learn and thus preclude the possibility for model deployment in clinical practices or for guiding further discovery.  Moreover, expert knowledge in the neuroimaging domain may help identify these undesirable behaviors. 

\item {\bf High accuracy does not necessarily indicate higher human comprehensibility:} A highly accurate model does not necessarily mean the model is more interpretable or relies on correct hidden features. While it is always preferable for clinicians, doctors, and scientists to find the true underlying factors for a model's prediction, it is very unlikely that model would rely on the same set of features every time it starts with new initialization weights. As indicated by Hinton in \cite{hinton2018deep}, with the deep learning models, it is not trying to identify the correct ``hidden factors'' responsible for a particular diagnosis. Instead, DL models could rely on a different set of hidden factors in the data to model the relationship between input and output variables. Hence, a highly accurate model may not have intelligible interpretations for humans. Refer to Figure \ref{fig:all_cams} and Figure~\ref{fig:all_cams_issue} to get ideas that explanations can immensely vary for different models even when the predictions remain the same. 

\begin{figure} [!htbp]
 \includegraphics[width=1\linewidth]{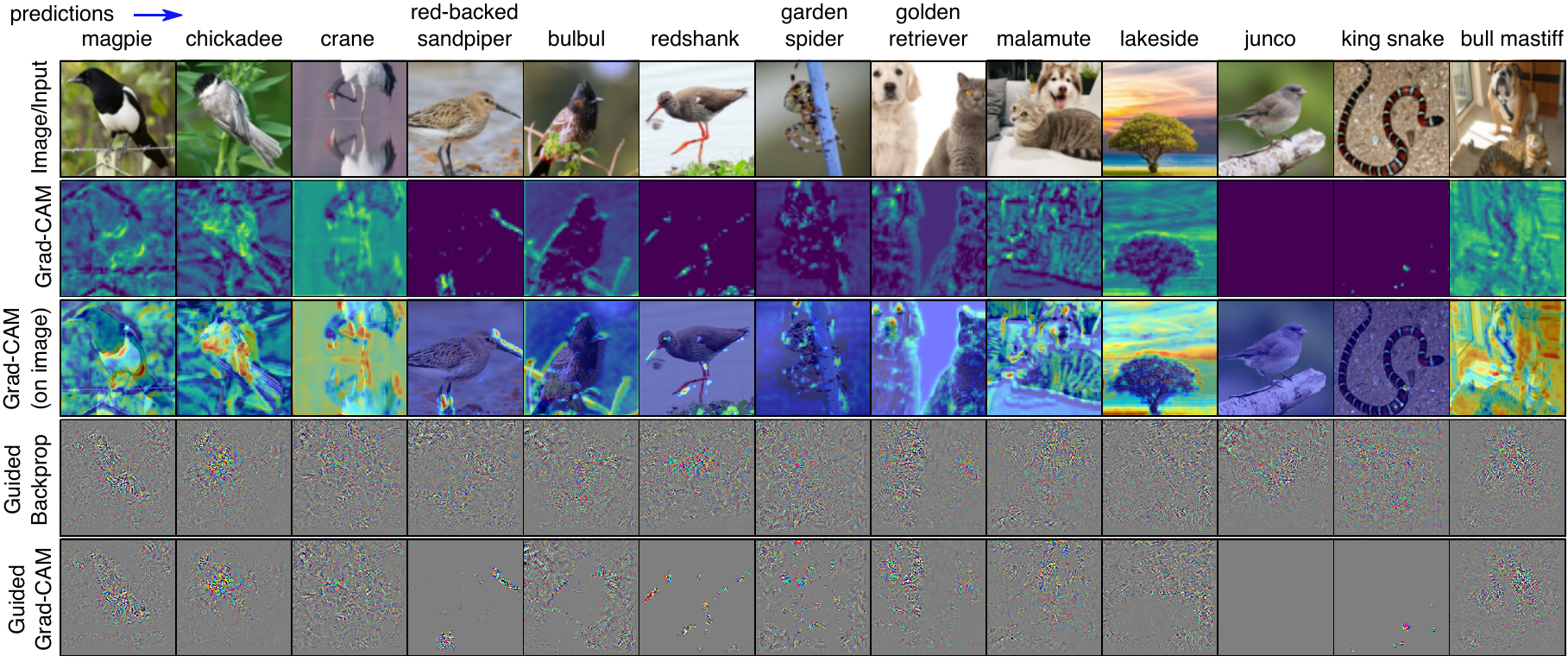}
 \centering
 \caption{Explanations generated using Grad-CAM, Guided Backpropagation, and Guided Grad-CAM methods for \emph{Inception v3} (trained on ImageNet) model predictions. Compared to Figure \ref{fig:all_cams} explanations for \emph{Resnet 50} model, the explanations for \emph{Inception v3} model are very different even for the same predictions. }
 \label{fig:all_cams_issue}
 \end{figure}
 
\item {\bf Design strong out-of-distribution (o.o.d) tests:} While the model may perform well on the i.i.d test samples, it may not generalize well for the real-world datasets because of associated distribution shifts~\cite{geirhos2020shortcut}. Hence, for successful deployment, we must design suitable o.o.d tests as suggested by Geirhos et al.~\cite{geirhos2020shortcut}. While it may be less hard for designing those tests for some problem domains, building strong o.o.d test cases may not be so easy for neuroimaging. However, multi-cohort datasets may help mitigate this problem. 

\item {\bf Check for stability in generated explanations:} Even if the model architecture, training data, and other hyperparameters remain same, each time the model is refitted starting from different initialization, the model may learn a very different set of features~\cite{hinton2018deep} which may or may not be desirable. This is indeed a very similar effect which occurs for the o.o.d cases, where a distribution-wise different test sample may end up with an untended behavior. This instability issue for the same model architecture, on the other hand, may happen during training time for different initializations. 

\item {\bf We need to be aware of the fragility of neural network interpretations:} 
The fundamental problem with the popular interpretability methods is their robustness~\cite{ghorbani2019interpretation}. Ghorbani et al.~\cite{ghorbani2019interpretation} showed that interpretations based on feature importance maps such as DeepLIFT, integrated gradients, and influence functions are susceptible to adversarial attacks. Put another way, a systematic perturbation of the input can lead to a very different interpretation (heatmap) because of the complexity of input feature space in deep neural networks. In neuroimaging, earlier studies, so far we are aware, usually overlooked this fragility of the interpretations, which may lead to misleading interpretations. Moreover, there is an inherent human bias to trust the model as correct and look for interpretations only based on predictive performance. While model inspection or debugging can be a hard problem in neuroimaging, it should be an essential consideration for this safety-critical domain. 

\item {\bf Lack of any guiding principle to select explanation methods:} 
While studies have leveraged different explanation methods for deep learning models, there is little theoretical evidence or guiding principle to choose a method for a particular study. Recently, Han et al.~\cite{han2022explanation} demonstrated how different explanation methods describe different neighborhoods and thus produce different explanations. Some disagreement scenarios are common because there could be differences in the underlying aspects the methods are investigating. For example, permutation importance~\cite{fisher2019all} and SHAP~\cite{lundberg2017unified, castro2009polynomial} in case of model overfitting may produce very different explanations. However, some disagreement scenarios are not expected. For example, gradients and LIME should produce similar interpretations because they both focus on local neighborhoods. However, in practice, they produce very different explanations. The authors in~\cite{han2022explanation} also showed how some methods cannot recover the underlying model and are entirely independent. The authors also provided valuable suggestions on choosing interpretability methods based on the nature of the data. They further suggested building an explanation method for the data for which no explanation method from the literature is considered beneficial. 

\item {\bf Post hoc methods are blamed for being insufficient:} As post hoc methods heavily rely on the models they are applied to, the methods can only discover the minimal discriminative parts sufficient for the prediction. For example, while LRP and GBP have been shown to be able to identify homogeneous brain regions, e.g., the hippocampus, they cannot identify heterogeneous regions, e.g., cortical folds~\cite{eitel2019testing, bohle2019layer}. 

\item {\bf Attribution normalization and polarity considerations varied widely:} 
For the post-processing of different explanations, studies use an ad-hoc approach. There has yet to be an agreement on how to post-process the heatmaps. This agreement must correspond to the underlying model and the interpretability method used. This necessity of the agreement is especially applicable to gradient-based attribution methods. Studies used the sign information differently to finalize the heatmaps. As the distribution of the values in the explanation maps generated using different methods varies widely, the need for agreed upon normalization has been a open research question~\cite{dyrba2020comparison}.

\item {\bf Studies generally use an ad-hoc approach to validate explanations:} 
For the validation of results, studies generally use informal and unreliable ways. Sometimes they used intuitions, hypotheses, and earlier results to justify the current attributions. These validation techniques are very susceptible and may end up with misleading conclusions. As Levakov et al.~\cite{levakov2020deep} indicated, any reasonable conclusions regarding the contributions should be made based on common parts of the maps from multiple models. Furthermore, deep learning models usually capture complex hierarchical and multivariate interactions. Localizing the brain regions should only be considered as an approximation of the significance. Even a small architectural modification can be a significant determinant of model performance and feature attribution maps, as indicated by Lin et al.~\cite{lin2022sspnet}. 

\item {\bf Validate explanations based on their predictability and expert evaluation:} 
While RAR~\cite{rahman2022interpreting} and ROAR~\cite{hooker2019benchmark} evaluations of the salient regions is promising and may further enhance the trust in the significance of what the model has learned, it may still need to be guaranteed that the model did not rely on spurious correlations. The domain experts should confirm the validation of the interpretations. Equivalently the explanations must match a significant proportion of the expert-extracted knowledge. We suggest complementing quantitative validation with neuro-scientifically valid explanations.

\item {\bf Use structure-function fusion model for model diagnosis:} 
Earlier studies, in general, independently focused on the anatomical or functional aspects of the dynamics. However, using both modalities simultaneously and corresponding existing knowledge in each modality during explanation generation may provide rigorous validation and bring trust in the explanations.

\item {\bf Counterfactuals may reveal the underlying biological mechanism:} 
Wachter et al.~\cite{wachter2017counterfactual} first introduced \emph{counterfactual} explanations to know about the hypothetical reality that could alter the model's decision. Dandi et al.~\cite{dandl2020multi} refined the formulation to satisfy the different practical desiderata of counterfactual explanations to make them useful in real-world applications. In the context of neuroimaging, we believe \emph{countefactual} explanations may help understand the underlying biological mechanism that potentially caused the specific disorder in the first place. To our knowledge, no neuroimaging study has ever used counterfactuals to understand the model's decision-making process.

\item {\bf Layer-wise Relevance Propagation (LRP) needs further investigation:} 
As seen from the interpretability in neuroimaging literature, LRP has been widely used, and its popularity is on an upward trend. However, the explanations produced by LRP are not reliable. Indeed, Shrikumar et al.~\cite{shrikumar2017learning} showed a strong connection between LRP and $gradient \odot input$, especially when all the activations are piecewise linear as in ReLU or Leaky ReLU. Ancona et al.~\cite{ancona2017towards} also showed that $\epsilon$-LRP is equivalent to the feature-wise product of the input and the modified partial derivative. Kindermans et al.~\cite{kindermans2017learning} showed that DeConvNet, Guided BackProp, and LRP cannot produce the theoretically correct explanation even for a linear model---the most straightforward neural network.

\item {\bf SHAP is popular, but it should not be trusted blindly:} 
SHAP, though very popular in the XAI community, has some issues. For example, SHAP assumes that the features are independent, while they are very unlikely. While features may be correlated, the algorithm may generate unrealistic observations (instances) with permutations. Moreover, no explanation method produces explanations that imply causality. SHAP indicates the importance of a feature based on the model prediction, not the importance in the real world. Humans are very prone to confirmation bias. It is not very uncommon that humans tend to create narratives as a result of confirmation bias. The most important question is: Did the model learn to predict for the right reasons? This question is vital because machine learning models do not know about truths, and it only cares about correlations, and proxy or secondary or less important variables may be loosely or tightly correlated with the actual cause. They can be revealed as very important features.
Moreover, Kwon and Zou~\cite{kwon2022weightedshap} recently showed that SHAP is suboptimal in that it gives the same weight to all marginal contributions for a feature $\vx_i$, which may potentially lead to attribution mistakes if different marginal contributions have different signal and noise. The authors further proposed a simple modification of the original SHAP, called WeightedSHAP, that estimates the weights automatically from the data.

\item {\bf Studies generally focused only on classification and regression tasks:}
While many studies in interpretable deep learning models for general classification tasks exist, further subgrouping into patient subtypes or clustering is still a novel area. This lack of interpretability literature for clustering tasks is equally true for neuroimaging and other domains. Very few studies did projection transformation from the latent space to observe the area of influence~\cite{martinez2019studying, biffi2020explainable}.

\item {\bf Effectiveness of transfer learning in neuroimaging needs justification:} what causes the increased accuracy? What knowledge does it transfer? Raghu et al.~\cite{raghu2019transfusion} showed that transfer learning from natural images to medical images did help little with performance. Instead, as the authors surmised, the slight improvement may come from the over-parameterization of the standard models trained on natural images. Moreover,  studies are not certain about the aspects of knowledge they are transferring from the natural image domain to the medical image domain or from one disorder area to another.
\end{enumerate}

\section{Conclusion}
\label{conclusion}

This article comprehensively introduces the problem of \emph{interpretability} for AI models and thus offers a field guide for future AI practitioners in the neuroimaging domain. In the earlier sections, we discussed the philosophical ground, dimensions, methods, and desirable axiomatic properties of model interpretability for reliable knowledge discovery. We also provide a useful taxonomy that directly points to all the major interpretability approaches and their use cases in many neuroimaging studies. We further discuss different sanity tests and evaluation metrics required to justify the validity of the explanations generated by any post hoc method. In the later sections, we discussed how deep learning approaches have been used widely in recent neuroimaging studies. Indeed, we performed an in-depth analysis of usage trends of the most prevailing interpretability methods. We reckon that these analyses will be helpful for future neuroimaging practitioners looking for ideas of how scientists are using these approaches for novel discoveries and how model interpretability is changing the course of neuroimaging studies in recent years. Lastly, we discuss different caveats of interpretability practices and provide insights on how this specialized sub-field of AI can be used wisely and meaningfully for better diagnosis, prognosis, and treatment of brain disorders.

\bibliographystyle{unsrt} 
\phantomsection
\addcontentsline{toc}{section}{References}
\bibliography{combined_bibliography}

\end{document}